%% file: aaai-got.tex
\documentclass[letterpaper]{article} % DO NOT CHANGE THIS
\usepackage{aaai24}  % DO NOT CHANGE THIS

\nocopyright
 
\newcommand{\name}{Graph of Thoughts}
\newcommand{\nameS}{Graph of Thoughts\ }
\newcommand{\nameR}{graph-of-thoughts}
\newcommand{\nameA}{GoT}
\newcommand{\nameAS}{GoT\ }

\usepackage{times}  % DO NOT CHANGE THIS
\usepackage{helvet}  % DO NOT CHANGE THIS
\usepackage{courier}  % DO NOT CHANGE THIS
\usepackage[hyphens]{url}  % DO NOT CHANGE THIS
\usepackage{graphicx} % DO NOT CHANGE THIS
\usepackage[square,numbers,sort&compress]{natbib}

\usepackage{caption} % DO NOT CHANGE THIS AND DO NOT ADD ANY OPTIONS TO IT
\usepackage{algorithm}
\usepackage{algorithmic}
\usepackage{tabularx}

\usepackage{newfloat}
\usepackage{listings}
\DeclareCaptionStyle{ruled}{labelfont=normalfont,labelsep=colon,strut=off} % DO NOT CHANGE THIS
\floatstyle{ruled}
\newfloat{listing}{tb}{lst}{}
\floatname{listing}{Listing}
\input{mac-includes}

\input{mac-math}
\newcommand{\faY}[0]{\faBatteryFull}
\newcommand{\faH}[0]{\faBatteryHalf}
\newcommand{\faN}[0]{\faTimes}

\title{\name: Solving Elaborate Problems with Large Language Models}

\newcommand*\samethanks[1][\value{footnote}]{\footnotemark[#1]}

\author{
    Maciej Besta\textsuperscript{\rm 1}\thanks{Equal contribution},
    Nils Blach\textsuperscript{\rm 1}\samethanks,
    Ales Kubicek\textsuperscript{\rm 1},
    Robert Gerstenberger\textsuperscript{\rm 1},\\
    Michał Podstawski\textsuperscript{\rm 2},
    Lukas Gianinazzi\textsuperscript{\rm 1},
    Joanna Gajda\textsuperscript{\rm 3},
    Tomasz Lehmann\textsuperscript{\rm 3},\\
    Hubert Niewiadomski\textsuperscript{\rm 3},
    Piotr Nyczyk\textsuperscript{\rm 3},
    Torsten Hoefler\textsuperscript{\rm 1}
}
\affiliations{
    \textsuperscript{\rm 1}ETH Zurich, \textsuperscript{\rm 2}Warsaw University of Technology, \textsuperscript{\rm 3}Cledar\\
    bestam@inf.ethz.ch, nils.blach@inf.ethz.ch, htor@inf.ethz.ch \vspace{1em}
}

\begin{document}

\maketitle

\begin{abstract}
We introduce \nameS (\nameA): a framework that advances prompting capabilities in large language models (LLMs) beyond those offered by paradigms such as Chain-of-Thought or Tree of Thoughts (ToT). The key idea and primary advantage of \nameAS is the ability to model the information generated by an LLM as an \emph{arbitrary graph}, where units of information (``LLM thoughts'') are vertices, and edges correspond to dependencies between these vertices. This approach enables combining arbitrary LLM thoughts into synergistic outcomes, distilling the essence of whole networks of thoughts, or enhancing thoughts using feedback loops. We illustrate that \nameAS offers advantages over state of the art on different tasks, for example increasing the quality of sorting by 62\% over ToT, while simultaneously reducing costs by $>$31\%. We ensure that \nameAS is extensible with new thought transformations and thus can be used to spearhead new prompting schemes.
This work brings the LLM reasoning closer to human thinking or brain mechanisms such as recurrence, both of which form complex networks.
\end{abstract}

{\small\textbf{Website \& code:} {\url{https://github.com/spcl/}\nameR}}

% \maciej{Other name ideas: ReasonGraph, ThinkGraph, Thoughts-As-Graph, ThoughtGraph, Network of Thoughts, Net of Thoughts}

\input{introduction}
\input{background}
\input{scheme}
\input{arch}
\input{cases}
\input{theory}
\input{eval}
\input{rw}
\input{conc}

\input{acks}

\bibliography{references.complete}

\newpage\newpage
\input{appendix}

%\maciej{Double check how this backtracing in ToTs really work - can't they just abandon a given path and do not backtrack? Or, by saying backtracking, they mean some conceptual concept that does not entail performance penalty?}

\end{document}

%% file: mac-includes.tex
\newif\ifsq     % Squeeze space?
\newif\ifsqCAP
\newif\ifsqVS
\newif\ifsqEN
\newif\ifsqTIT
\newif\ifsqTAB

\newcommand{\ignore}[1]{}

\sqtrue
\sqCAPfalse
\sqENtrue
\sqVStrue
\sqTITtrue
\sqTABtrue

\sqfalse
\sqCAPfalse
\sqENfalse
\sqVSfalse
\sqTITfalse
\sqTABtrue

\usepackage{etex}
\usepackage{epstopdf}
\usepackage{placeins}

\usepackage{graphicx}
\usepackage{float}
\usepackage{dblfloatfix}
\usepackage{multirow}
\usepackage{rotating}
\usepackage{makecell}
\usepackage{tabulary}
\usepackage{parcolumns}
\usepackage{tikz}
\usetikzlibrary{tikzmark}

\usepackage{xpatch}
\expandafter\xpatchcmd
\csname pgfk@/tikz/every picture/.@cmd\endcsname
{\thepage}{\arabic{page}}{}{}

\tikzstyle{comment} = [draw, fill=blue!70, text=white, text width=3cm, minimum height=1cm, rounded corners, align=left, font=\scriptsize]
\tikzstyle{background_alg} = [draw, fill=blue!20, opacity=0.4, inner sep=4pt, rounded corners=2pt]

\usetikzlibrary{shapes}
\usetikzlibrary{plotmarks}
\usetikzlibrary{calc, fit}

\usepackage{enumitem}

\usepackage{amsthm}
\usepackage{amsmath,amssymb,amsfonts}
\usepackage{mathtools,mathrsfs}

\usepackage{soul}
\usepackage{fontawesome}
\usepackage{pifont}
\usepackage{textcomp}
\usepackage{booktabs}
\usepackage{url}
\usepackage{pbox}
\usepackage[10pt]{moresize}
\ifsqCAP
\usepackage[font={sf, scriptsize}]{caption}
\usepackage[font={sf, scriptsize}]{subcaption}
\else
\fi

\newcommand{\vspaceSQ}[1]{\ifsqVS\vspace{#1}\fi}
\newcommand{\enlargeSQ}[1]{\ifsqEN\enlargethispage{\baselineskip}\fi}

\ifsqTIT
\usepackage[compact]{titlesec}
\titlespacing*{\section}{0pt}{3pt}{-1pt}
\titlespacing*{\subsection}{0pt}{0pt}{-3pt}
\titlespacing*{\subsubsection}{0pt}{2pt}{1pt}
\fi

\usepackage{xcolor}
\definecolor{darkgrey}{RGB}{70,70,70}
\definecolor{lightgrey}{RGB}{200,200,200}
\definecolor{lyellow}{RGB}{255,255,100}
\definecolor{llyellow}{RGB}{250,250,180}
\definecolor{lgreen}{RGB}{144,238,144}
\definecolor{raphael_comments}{RGB}{13, 145, 24}

\usepackage[customcolors]{hf-tikz}
\hfsetbordercolor{white}
\hfsetfillcolor{vlgray}

\definecolor{vlgray}{rgb}{0.77 0.77 0.77}
\definecolor{ablack}{rgb}{0.2 0.2 0.2}
\definecolor{vllgray}{rgb}{0.9 0.9 0.9}
\definecolor{bblue}{rgb}{0.7 0.7 0.99}

\usepackage{colortbl}

\usepackage{inconsolata}
\usepackage{listings}

\ifsq
\else
\fi

\newcommand{\maciej}[1]{\textcolor{blue}{[Maciej: #1]}}

\definecolor{hlL}{rgb}{0.8 0.8 0.99}

\newcounter{highlight}

\newcounter{hlLR}

\newcounter{hlLIR}

\newcounter{hlLIIR}

\newcounter{Ahighlight}

\usepackage{scalerel,stackengine}
\stackMath
\newcommand\rwh[1]{%
\savestack{\tmpbox}{\stretchto{%
  \scaleto{%
        \scalerel*[\widthof{\ensuremath{#1}}]{\kern-.6pt\bigwedge\kern-.6pt}%
                  {\rule[-\textheight/2]{1ex}{\textheight}}%WIDTH-LIMITED BIG WEDGE
                              }{\textheight}% 
}{0.5ex}}%
\stackon[1pt]{#1}{\tmpbox}%
}

\usepackage[hang,flushmargin]{footmisc}

\renewcommand{\epsilon}{\ensuremath\varepsilon}

\renewcommand{\phi}{\ensuremath{\varphi}}

\if 0

\usepackage[linesnumbered,ruled]{algorithm2e}
\usepackage{multicol}
\SetKwComment{Comm}{$\triangleright$\ }{}
\SetAlFnt{\scriptsize}
\SetAlCapFnt{\scriptsize}
\SetAlCapNameFnt{\scriptsize}
\SetKwInOut{Input}{Input}
\SetKwInOut{Output}{Output}

\makeatletter
\NewDocumentCommand{\LeftComment}{s m}{%
\Statex \IfBooleanF{#1}{\hspace*{\ALG@thistlm}}\(\triangleright\) #2}
\makeatother

\fi

%% file: introduction.tex
\section{Introduction}
\label{sec:intro}

Large language models (LLMs) are taking over the world of AI. Recent years saw a rapid development of models primarily based on the decoder-only Transformer variant~\cite{vaswani2017attention}, such as GPT~\cite{radford2019language, radford2018improving, bubeck2023sparks, brown2020language}, PaLM~\cite{chowdhery2022palm}, or LLaMA~\cite{touvron2023llama}.

\emph{Prompt engineering} is a resource-efficient approach for solving different LLM tasks. In brief, one includes the task description  within the input sent to an LLM. If this description is appropriately formulated, the LLM solves the task using its autoregressive token-based mechanism for generating text. Such prompts may contain example tasks with solutions (few-shot prompting, also referred to as in-context learning (ICL)), or even no example tasks at all (zero-shot prompting). In recent years it was shown that this mechanism can be used to solve a broad set of tasks that involve mathematical, commonsense, or symbolic reasoning.

Chain-of-Thought (CoT)~\cite{wei2022chain} is an approach for prompting, in which one includes the intermediate steps of reasoning within the prompt (intermediate ``thoughts''), besides the task input/output. CoT was shown to significantly improve the capability of LLMs to solve problems without resorting to any model updates. One major improvement over CoT, Self-Consistency with CoT (CoT-SC)~\cite{wang2022self}, is a scheme where multiple CoTs are generated, and then the best one is selected as the outcome. More recently, CoT and CoT-SC were extended with Tree of Thoughts (ToT)~\cite{long2023large, yao2023tree, xie2023decomposition}, which models the LLM reasoning process with a tree. This facilitates using different paths of thoughts, and offers novel capabilities such as backtracking from non-promising outcomes. Unfortunately, the ToT approaches still fundamentally limit the reasoning abilities within a prompt by imposing the rigid tree structure on the thought process.

In this work, we argue that fundamentally more powerful prompting can be achieved by enabling LLM thoughts to form an arbitrary graph structure. This is motivated by numerous phenomena such as human reasoning, brain structure, or algorithmic execution. When working on a novel idea, a human would not only follow a chain of thoughts (as in CoT) or try different separate ones (as in ToT), but would actually form a more complex network of thoughts. For example, one could explore a certain chain of reasoning, backtrack and start a new one, then realize that a certain idea from the previous chain could be combined with the currently explored one, and merge them both into a new solution, taking advantage of their strengths and eliminating their weaknesses.
Similarly, brains form complex networks, with graph-like patterns such as recurrence~\cite{friston2008hierarchical}.
Executing algorithms also expose networked patterns, often represented by Directed Acyclic Graphs.
The corresponding \emph{graph-enabled transformations} bring a promise of more powerful prompting when applied to LLM thoughts, but they are not naturally expressible with CoT or ToT.

We observe that these (and many other) thought transformations can be naturally enabled when \emph{modeling the reasoning process of an LLM as a graph}. For this, we propose \nameS (\nameA), an approach that \emph{enhances LLMs' capabilities through networked reasoning} (\textbf{contribution~\#1}). In \nameA, an LLM thought is modeled as a vertex, while an edge is a dependency between such thoughts. Using \nameA, one can aggregate arbitrary thoughts by constructing vertices that have more than one incoming edge. Overall, the graph abstraction harnessed by \nameAS seamlessly generalizes CoT and ToT to more complex thought patterns, \emph{without resorting to any model updates}.

Yet, putting \nameAS to practice requires solving several design challenges. For example, what is the best graph structure for different tasks? How to best aggregate thoughts to maximize accuracy and minimize cost? To answer these and many other questions, we carefully design a modular architecture for implementing \nameAS (\textbf{contribution~\#2}), coming with two design highlights.
%
\if 0
Similarly to the ToT design by Long~\cite{long2023large}, our architecture harnesses a prompter agent (to construct prompts), a evaluator module (to verify or score the LLM thoughts), a memory module (to maintain the state of the conversation), and a controller (to manage the conversation). There are two significant enhancements that we employ. 
\fi
%
First, we enable a \emph{fine-grained control over individual thoughts}. This enables us to fully control the ongoing conversation with the LLM, and apply advanced thought transformations, such as combining most promising thoughts from the ongoing reasoning into a new one. Second, we ensure that our architecture can be seamlessly extended with novel thought transformations, patterns of reasoning (i.e., graphs of thoughts), and LLM models. This enables rapid prototyping of novel prompting ideas using \nameA, while experimenting with different models such as GPT-3.5, GPT-4, or Llama-2~\cite{touvron2023llama2}.

% We further enhance \nameAS by harnessing \emph{different classes of thoughts}. For example, it was discovered that asking the LLM to first plan the next steps, and only then execute them, can improve the quality of ultimate solutions~\cite{??}. We embrace this in \nameAS      the \emph{heterogeneous graph model} to harness

% Two modes for taking decisions (fast, unconscious; slow, conscious)~\cite{stanovich1999rational, sloman1996empirical, kahneman2002representativeness, daniel2000thinking}

% \maciej{Find use cases for the following} Next, GoT embraces the heterogeneous and higher-order graph models, which enables different classes of thoughts as well as hierarchical thoughts. This further advances the reasoning capability by harnessing hierarchies of thoughts, thoughts of different classes. 

% \maciej{multi-modal? Vertex as text and vertex as picture?}

We illustrate several use cases for \nameAS (sorting, keyword counting for summaries, set operations, document merging) and we detail how to implement them using the graph-based paradigm (\textbf{contribution~\#3}).
We evaluate \nameAS and show its advantages over the state of the art (\textbf{contribution~\#4}). Overall, we observe that \nameAS is particularly well-suited for tasks that can be naturally decomposed into smaller subtasks that are solved individually and then merged for a final solution. Here, \nameAS outperforms other schemes, for example improving upon CoT and ToT by, respectively, $\approx$70\% and $\approx$62\%, in terms of the quality of sorting, while \emph{simultaneously} reducing costs by $>$31\% over ToT.

We qualitatively compare \nameAS to other prompting schemes\footnote{Note that we do not include a recent scheme called Graph-of-Thought~\cite{yao2023beyond} because it is not a prompting scheme. While its name suggests close connections to ToT and CoT, as a fine-tuning scheme, it resorts to model updates, and is thus outside the focus of this work. Similarly, the graph-of-thoughts repository~\cite{qrdlgit@got} does not enable general graph-based reasoning and harnesses instead ToT with BFS.} in Table~\ref{tab:schemes}. \nameAS is the only one to enable arbitrary graph-based thought transformations within a prompt, such as aggregation, embracing all previously proposed schemes.
%
% Note that, while we also list a recent Graph-of-Thought scheme~\cite{yao2023beyond} due to its name suggesting graph-like capabilities, it is actually a fine-tuning approach. Hence, it is outside the focus of this work as it does not enable any of these capabilities within a prompt, resorting to model updates.

\input{table-schemes}

Finally, we propose a new metric for evaluating a prompting strategy, the \emph{volume of a thought} (\textbf{contribution~\#5}). With this metric, we aim to understand better the differences between prompting schemes. For a given thought $v$, the volume of $v$ is \emph{the number of LLM thoughts, from which one can reach $v$ using directed edges}. Intuitively, these are all the LLM thoughts that have had the potential to contribute to $v$. We show that \nameA, by incorporating thought transformations such as aggregation, enables thoughts to have fundamentally larger volumes than other schemes.

\begin{figure*}[t]
\centering
\vspaceSQ{-0.5em}
\includegraphics[width=1.0\textwidth]{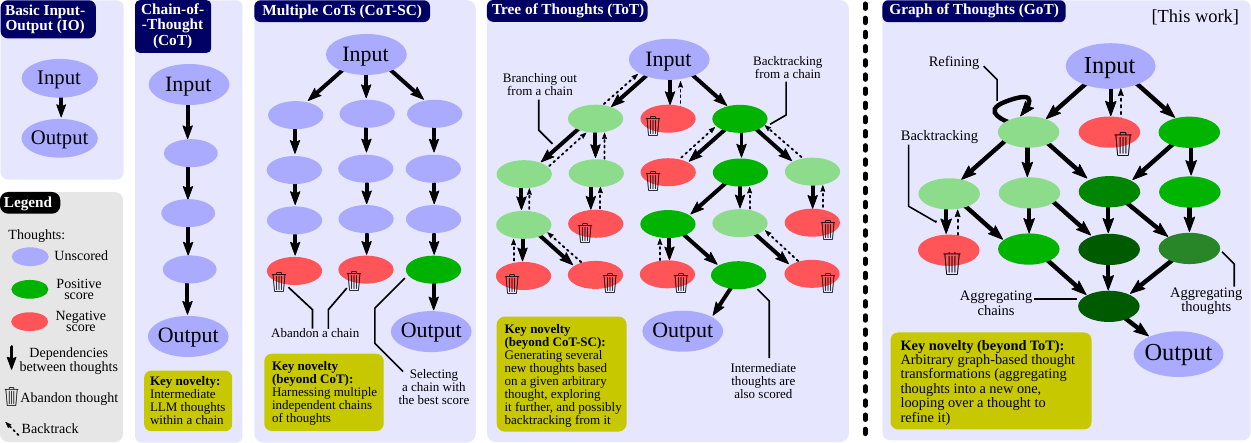}
\vspace{-1em}
\caption{Comparison of \nameS (\nameA) to other prompting strategies.}
\vspaceSQ{-0.5em}
\label{fig:schemes}
\end{figure*}

%% file: table-schemes.tex
\if 0

\begin{table*}[t]
\vspaceSQ{-1em}
\centering
\setlength{\tabcolsep}{1.5pt}
\footnotesize
\ifsqTAB\renewcommand{\arraystretch}{0.7}\fi
\ifsqTAB\footnotesize\fi
%\ssmall
%\sf
\begin{tabular}{llllll}
\toprule
%
\textbf{Scheme} & \textbf{Prompting?} & \textbf{Single chain?} & \textbf{Multiple chains?} & \textbf{Tree?} & \textbf{Arbitrary graph?} \\
%
\midrule
%
Chain-of-Thought (CoT)~\cite{wei2022chain} & \faY & \faY & \faN & \faN & \faN  \\
Self-Consistency with CoT~\cite{wang2022self} & \faY & \faY & \faY & \faN & \faN \\
Thought decomposition~\cite{xie2023decomposition} & \faY & \faY & \faY & \faH & \faN  \\
Tree-of-Thought (ToT)~\cite{long2023large} & \faY & \faY & \faY & \faY & \faN \\
Tree of Thoughts (ToT)~\cite{yao2023tree} & \faY & \faY & \faY & \faY & \faN \\
%
% \midrule
%
Graph-of-Thought (GoT)$^\dagger$~\cite{yao2023beyond} & \faN\ (fine-tuning) & \faN & \faN & \faN & \faN \\
%
\midrule
%
\textbf{\nameS (\nameA)} & \faY & \faY & \faY & \faY & \faY \\
%
\bottomrule
\end{tabular}
\vspace{-1em}
\caption{Comparison of prompting schemes, with respect to the supported transformations of thoughts \emph{within a prompt}. $^\dagger$While GoT is not a prompting scheme, we still include it as it name suggests close connections to the ToT and CoT schemes. However, as a fine-tuning scheme, it does not support structured prompts.
%
``\faY'': full support,
``\faH'': partial support,
``\faN'': no support.}
\label{tab:schemes}
\vspaceSQ{-0.5em}
\end{table*}

\fi

\begin{table}[h]
\vspaceSQ{-0.5em}
\centering
\setlength{\tabcolsep}{1.5pt}
\ifsqTAB\renewcommand{\arraystretch}{0.9}\fi
\ifsqTAB\footnotesize\fi
%\ssmall
%\sf
\begin{tabular}{llllll}
\toprule
\textbf{Scheme} & \textbf{Sc?} & \textbf{Mc?} & \textbf{Tr?} & \textbf{Ag?} \\
\midrule
Chain-of-Thought (CoT)~\cite{wei2022chain} &  \faY & \faN & \faN & \faN  \\
Self-Consistency with CoT~\cite{wang2022self} &  \faY & \faY & \faN & \faN \\
Thought decomposition~\cite{xie2023decomposition} &  \faY & \faY & \faH & \faN  \\
Tree-of-Thought (ToT)~\cite{long2023large} &  \faY & \faY & \faY & \faN \\
Tree of Thoughts (ToT)~\cite{yao2023tree} &  \faY & \faY & \faY & \faN \\
%
% \midrule
%
% Graph-of-Thought (GoT)$^\dagger$~\cite{yao2023beyond} & \faN & \faN & \faN & \faN & \faN \\
%
\midrule
\textbf{\nameS (\nameA)} & \faY & \faY & \faY & \faY \\
\bottomrule
\end{tabular}
\vspace{-1em}
\caption{Comparison of prompting schemes, with respect to the supported transformations of thoughts.
%
%\textbf{``P?''}: is a given scheme purely prompting based?
%
\textbf{``Sc?''}: single chain of thoughts?
\textbf{``Mc?''}: multiple chains of thoughts?
\textbf{``Tr?''}: tree of thoughts?
\textbf{``Ag?''}: arbitrary graph of thoughts?
``\faY'': full support,
``\faH'': partial support,
``\faN'': no support.}
\label{tab:schemes}
\vspaceSQ{-0.5em}
\end{table}

%% file: background.tex
\section{Background \& Notation}

We first outline background concepts and notation.

\subsection{Language Models \& In-Context Learning}

The \textbf{conversation with the LLM} consists of user messages (\emph{prompts}) and LLM replies (\emph{thoughts}).
We follow the established notation~\cite{yao2023tree} and we denote a pre-trained language model (LM) with parameters $\theta$ as $p_\theta$. Lowercase letters such as $x, y, z, ...$ indicate LLM thoughts. 
We purposefully do not prescribe what is a single ``thought'', and instead make it use-case specific. Hence, a single thought can be a paragraph (e.g., in article summary), a document (e.g., in document generation), a block of code (e.g., in code debugging or optimization), and so on.

We next describe specific \textbf{prompting approaches}.
%
\if 0
Following the notation by~\cite{yao2023tree}, we use a simplified notation, in which the output $y$ of a model $p_\theta$, given a prompt $X$ and an input $x$, is referred to as $y \sim p_\theta^{X} (y | x)$. 
%
\fi

\subsubsection{Input-Output (IO)}

The Input-Output (IO) prompting is a straightforward approach, in which we use an LLM to turn an input sequence $x$ into the output $y$ \emph{directly}, without any intermediate thoughts.
%
\if 0
We have $y \sim p^{IO}_\theta (y|x)$.
\fi

\subsubsection{Chain-of-Thought (CoT)}

Second, in Chain-of-Thought (CoT), one introduces intermediate thoughts $a_1, a_2, ...$ between $x$ and $y$. 
%
\if0
To obtain the next thought $i$, we have $a_i \sim p^{CoT}_\theta (a_i | x, a_1, a_2, ..., a_{i-1})$ and $y \sim p^{CoT}_\theta (y | x, a_1, a_2, ..., a_N)$.
\fi
%
This strategy was shown to significantly enhance various LM tasks over the plain IO baseline, such as mathematical puzzles~\cite{wei2022chain} or general mathematical reasoning~\cite{drori2022neural}.

\subsubsection{Multiple CoTs}

Third, one can generalize CoT into \emph{multiple CoTs} by generating \emph{several} (independent) $k$ CoTs, and returning the one with the best output (according to some prescribed scoring metric). 
%
\if0
We have $[y^{(1)}, ..., y^{(k)}] \sim p_\theta^{CoT} (y^{(1)}, ..., y^{(k)} | x, a^{(1)}_1, a^{(1)}_2, ..., a^{(1)}_N, a^{(2)}_1, a^{(2)}_2, ..., a^{(2)}_N,$ $..., a^{(k)}_1, a^{(k)}_2, ..., a^{(k)}_N)$. 
\fi
%
It was introduced by Wang et al.~in the scheme called Self-Consistency with CoT (CoT-SC)~\cite{wang2022self}. This approach enhances CoT because it offers an opportunity to explore different reasoning paths. However, it does not offer ``local exploration'' within a path, such as backtracking.

\subsubsection{Tree of Thoughts (ToT)}

Finally, the Tree of Thoughts (ToT) scheme was introduced independently by Yao~\cite{yao2023tree} and Long~\cite{long2023large} (where it is referred to as Tree-of-Thought); it was used implicitly to a certain degree by other schemes such as thought decomposition~\cite{xie2023decomposition}. It enhances CoT-SC by modeling the process or reasoning as a \emph{tree} of thoughts.
A single tree node represents a partial solution. Based on a given node, the \emph{thought generator} constructs a given number $k$ of new nodes. Then, the \emph{state evaluator} generates scores for each such new node. Depending on the use case, the evaluation could be conducted using an LLM itself, or it can harness human scores. Finally, the schedule of extending the tree is dictated by the utilized search algorithm (for example BFS or DFS).

\if0
Formally, a single tree node is modeled as a state $s = [x, a_{1, ..., i}]$, which represents a partial solution. Then, based on $s$, a \emph{thought generator} $G(p_\theta, s, k)$ constructs $k$ new candidate states $y^{(i)} \sim $

\maciej{Finish}
\fi

\if0
\subsection{Decomposition of Thoughts}

% The above definitions leave a lot of room for flexibility in terms of how prompts, and specific thoughts, are constructed, sampled, and ordered.

Especially in the context of CoT, one does not have to provide intermediate thoughts explicitly, but rather attempt to generate them based solely on the input. In that case, for a single CoT, we would have $[y, a_1, a_2, ..., a_N] \sim p^{CoT}_\theta (y, a_1, a_2, ..., a_N | x)$ instead of $y \sim p^{CoT}_\theta (y | x, a_1, a_2, ..., a_N)$.

\fi

\if 0
\subsection{Graph Theory}

We use \textbf{graph theory} to model the LLM though process. Here, a graph is a tuple $G=(V,E)$; $V$ is a set of vertices (nodes); $|V|=n$ and $|E|=m$. $G$ is directed and thus the edges are a subset of ordered vertex pairs $E\subseteq V\times V$. 
%
% $N_{in}(v)$ and $N_{out}(v)$ are in- and out-neighborhoods of a vertex~$v$, i.e., $N_{in}(v) = \{w \in V\ |\ (w,v) \in E \}$ and $N_{out}(v) = \{w \in V\ |\ (v,w) \in E \}$. Then, $d_{in}(v)$ and $d_{out}(v)$) are the corresponding in- and out-degrees. 
%
A vertex models an LLM thought while and edge $(u,v)$ models the fact that $u$ is used to obtain $v$. 

\fi

%% file: scheme.tex
\section{The \nameAS Framework}

We now detail the \nameAS framework. We present it in Figure~\ref{fig:schemes}, and compare it to other prompting strategies.

Formally, \nameAS can be modeled as a tuple $(G, \mathcal{T}, \mathcal{E}, \mathcal{R})$, where $G$ is the ``LLM reasoning process'' (i.e., all the LLM thoughts within the context, with their relationships), $\mathcal{T}$ are the potential thought transformations, $\mathcal{E}$ is an evaluator function used to obtain scores of thoughts, and $\mathcal{R}$ is a ranking function used to select most relevant thoughts. 

\subsection{Reasoning Process}

We model the reasoning process as a directed \textbf{graph} $G = (V, E)$; $V$ is a set of vertices and $E\subseteq V\times V$ is a set of edges. 
$G$ is directed and thus the edges are a subset of ordered vertex pairs $E\subseteq V\times V$.
A vertex contains a \emph{solution} to a problem at hand (be it an initial, intermediate, or a final one). The concrete form of such a thought depends on the use case; it could be a paragraph (in writing tasks) or a sequence of numbers (in sorting). A directed edge $(t_1, t_2)$ indicates that thought $t_2$ has been constructed using $t_1$ as ``direct input'', i.e., by explicitly instructing the LLM to use $t_1$ for generating $t_2$.

In certain use cases, graph nodes belong to different \emph{classes}. For example, in writing tasks, some vertices model \emph{plans of writing a paragraph}, while other vertices model \emph{the actual paragraphs of text}. In such cases, \nameAS embraces a \textbf{heterogeneous graph} $G = (V, E, c)$ to model the LLM reasoning, where $c$ maps vertices $V$ into their respective classes $C$ (in the above case, it would be $C = \{ plan, par \}$). Hence, any vertex $v$ can model different aspects of reasoning.
%
\if 0
%
Most often, it forms a \emph{solution} to a problem at-hand (be it an initial, intermediate, or a final one), i.e., $c(v) = s$. The concrete form of such a thought depends on a specific use case. For example, it could be a paragraph (in writing tasks) or a sequence of numbers (in sorting tasks). On top of that, we also use vertices to model \emph{plans} of future actions; in which case $c(v) = p$. We found out that explicitly asking the LLM to plan future thoughts by generating a thought that contains this plan, enables more powerful reasoning. Overall, unless stated otherwise, we assume $C = \{s, p\}$. 
%
\fi

We associate $G$ with the LLM reasoning process. To advance this process, one applies \textbf{thought transformations} to $G$. An example of such a transformation is to merge best-scoring (so far) thoughts into a new one. Another example is to loop over a thought, in order to enhance it. Note that these transformations strictly extend the set of transformations available in the CoT, CoT-SC, or ToT. 
%
% For example, a potential transformation for a ToT would be to backtrack from a given path of thought. We detail all thought transformations in Section~\ref{sec:transf}.

\subsection{Transformations of Thoughts}
\label{sec:transf}

\begin{figure}[t]
\centering
\vspaceSQ{-1em}
\includegraphics[width=1.0\columnwidth]{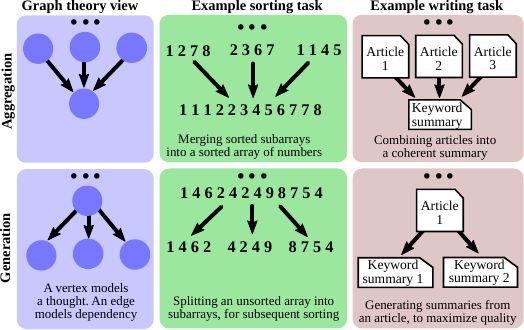}
\vspace{-0.5em}
\caption{Examples of aggregation and generation thought transformations.}
\vspaceSQ{-1em}
\label{fig:examples}
\end{figure}

\nameAS enables novel transformations of thoughts thanks to the graph-based model for reasoning. We refer to them as \textbf{graph-enabled transformations}.
For example, in writing, one could combine several input articles into one coherent summary. In sorting, one could merge several sorted subarrays of numbers into a final sorted array. 
We illustrate examples of aggregation and generation in Figure~\ref{fig:examples}.

Formally, each such transformation can be modeled as $\mathcal{T}(G, p_\theta)$ where $G = (V,E)$ is the graph reflecting the current state of the reasoning, and $p_\theta$ is the used LLM. $\mathcal{T}$ modifies $G$ usually by adding new vertices and their incoming edges.
We have $G' = \mathcal{T}(G, p_\theta) = (V', E')$, where $V' = ( V \cup V^+ ) \setminus V^-$ and $E' = ( E \cup E^+ ) \setminus E^-$. $V^+$ and $E^+$ are new vertices and edges inserted into $G$ to model the new thoughts and their dependencies, respectively. To maximize the expressiveness of \nameAS -- we also enable the user to explicitly \emph{remove} thoughts, by specifying the corresponding vertices and edges to be removed ($V^-$ and $E^-$, respectively).
Here, it is the user's responsibility to ensure that the sets $V^+, E^+, V^-,$ and $E^-$ come with consistent transformations (i.e., for example, that the user does not attempt to remove a vertex that does not exist). 
This enables seamless incorporation of schemes where, in order to save space within the context, one can remove parts of reasoning that do not promise improvements.

The specific form of $\mathcal{T}$ and how it impacts $G$ depends on a specific transformation. We first detail the primary graph-enabled thought transformations, and then proceed to describe how \nameAS embraces the transformations from the earlier schemes. Unless stated otherwise, $V^- = E^- = \emptyset$.

\begin{figure*}[hbtp]
\centering
\vspaceSQ{-1em}
\includegraphics[width=1.0\textwidth]{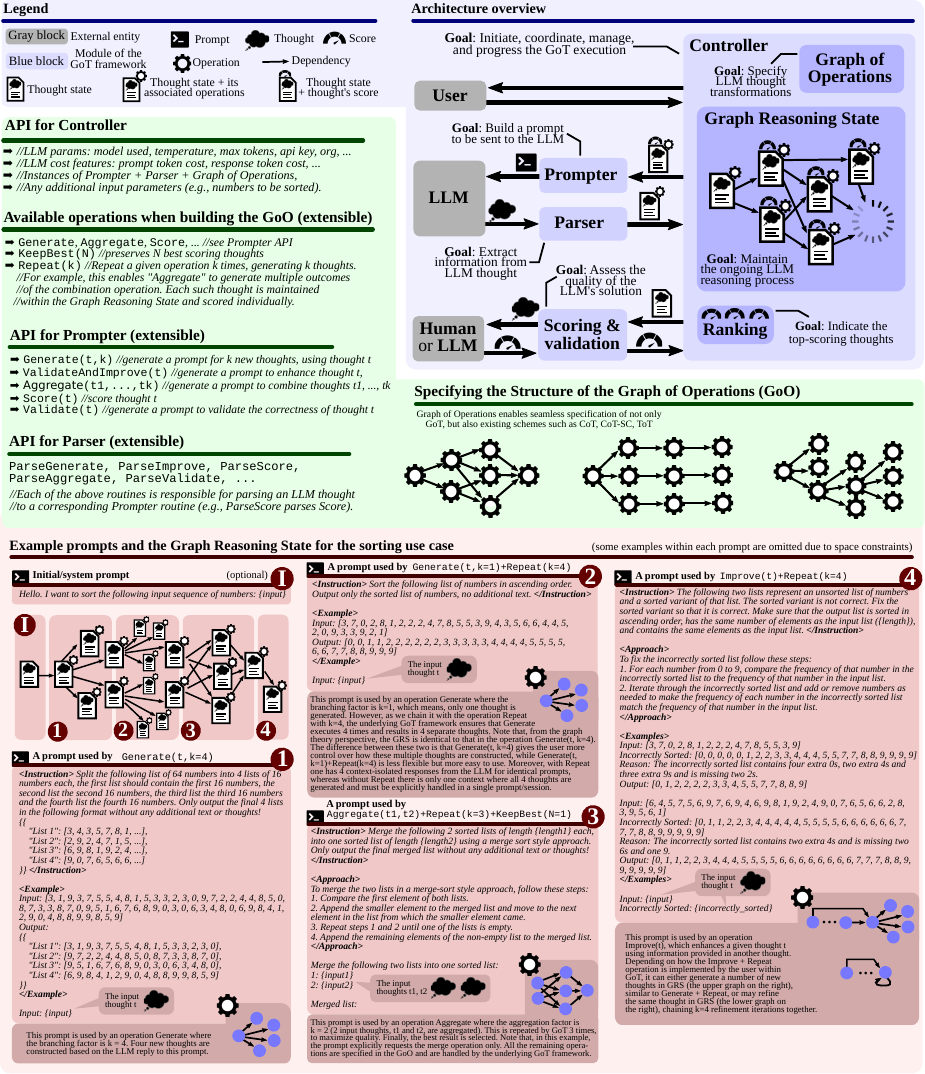}
\vspaceSQ{-1em}
\caption{The system architecture of \nameA, and the APIs of respective modules. The user can straightforwardly extend the design towards new prompting schemes, experiment with novel thought transformations, and plug in different LLMs. The \textcolor{blue}{blue part of the figure contains the architecture overview}, \textcolor{green}{the green part lists the API}, and \textcolor{red}{the red part contains example prompts together with a GRS and operations involved}.}
\vspaceSQ{-1em}
\label{fig:arch}
\end{figure*}

\subsubsection{Aggregation Transformations}

First, with \nameA, one can \textbf{aggregate arbitrary thoughts} into new ones, to combine and reinforce the advantages of these thoughts, while eliminating their disadvantages. 
%
% For example, in a writnig task, this could correspond to 
%
In the basic form, in which only one new vertex is created, $V^+ = \{v^+\}$ and $E^+ = \{ (v_1, v^+), ..., (v_k, v^+) \}$, where $v_1, ..., v_k$ are the merged $k$ thoughts.
% 
%\subsubsection{Convergence of Paths}
%
More generally, this enables \textbf{aggregating reasoning paths}, i.e., longer chains of thoughts, beyond just individual thoughts. With the graph model, it is simply achieved by adding outgoing edges from the vertices $v_1, ..., v_k$, modeling final thoughts in several chains, into a single thought $v^+$ combining these chains.

\subsubsection{Refining Transformations}

Another thought transformation is the \textbf{refining} of a current thought $v$ by modifying its content: $V^+ = \{\}$ and $E^+ = \{ (v, v) \}$. %This could be done after deciding that the current chain (with $v$ in it) is not going to improve with more steps, but using $v$'s contents could enhance $u$.
This loop in the graph indicates an iterated thought with the same connections as the original thought. % can also iterate over the current thought. %Note that, formally, this class is a special case of aggregation, in which there is only one input vertex.

\subsubsection{Generation Transformations}

Finally, one can \textbf{generate one or more new thoughts based on an existing single thought} $v$. This class embraces analogous reasoning steps from earlier schemes, such as ToT or CoT-SC. Formally, we have $V^+ = \{ v^+_1, ..., v^+_k \}$ and $E^+ = \{ (v, v^+_1), ... ,(v, v^+_k) \}$.

\subsection{Scoring \& Ranking Thoughts}

Thoughts are scored to understand whether the current solution is good enough. A score is modeled as a general function $\mathcal{E}(v, G, p_\theta)$, where $v$ is a thought to be evaluated. We use the state of the whole reasoning process ($G$) in $\mathcal{E}$ for maximum generality, because -- for example -- in some evaluation scenarios, scores may be relative to other thoughts.

% \subsection{Ranking Thoughts}

\nameAS can also rank thoughts. We model this with a function $\mathcal{R}(G, p_\theta, h)$ where $h$ specifies the number of highest-ranking thoughts in $G$ to be returned by $\mathcal{R}$. While the specific form of $\mathcal{R}$ depends on the use case, we most often use a simple yet effective strategy where $h$ thoughts with the highest scores are returned, i.e., $v_1, ..., v_h = \mathcal{R}(G, p_\theta, h)$.

Specific forms of $\mathcal{E}$ and $\mathcal{R}$ depend on the use case. We discuss the details in Section~\ref{sec:cases}. For example, the score (or rank) for sorting corresponds to the count of elements correctly sorted (or incorrectly, when using the error as a score).

%% file: arch.tex
\section{System Architecture \& Extensibility}
\label{sec:arch}

% The architecture extends that of ToT by Long~\cite{long2023large}. 
%
The \nameAS architecture consists of a set of interacting modules, see Figure~\ref{fig:arch} \textcolor{blue}{(the blue part)}. These modules are the {Prompter} (prepares the messages for the LLM), the Parser (extracts information from LLM thoughts), the {Scoring module} (verifies and scores the LLM thoughts), and the {Controller} (coordinates the entire reasoning process, and decides on how to progress it).
The Controller contains two further important elements: the Graph of Operations (GoO) and the Graph Reasoning State (GRS). GoO is a static structure that specifies the \emph{graph decomposition of a given task}, i.e., it prescribes transformations to be applied to LLM thoughts, together with their order \& dependencies.
GRS is a dynamic structure that maintains the state of the ongoing LLM reasoning process (the history of its thoughts and their states).

%
\if 0
Note that -- while we borrow this structure from the ToT work -- we extend it in two ways. First, we accommodate the general graph-based reasoning. Second, we ensure straightforward extendibility with new prompting schemes and use cases.
\fi

\subsection{Prompter}

The {Prompter} prepares the prompts to be sent to the LLM. This module is responsible for the specifics of encoding the graph structure within the prompt.
The \nameAS architecture enables the user to implement use case specific graph encodings by providing full access to the graph structure.
%
% \maciej{Can you outline/summarize the way we do it, with concrete examples of the encoding, notation, etc.?}
%
% For example, in order to refer to specific graph vertices, in most cases, we use a simple monotonically increasing numbering of thoughts.

\subsection{Parser}

The {Parser} extracts information from LLM thoughts. For each such thought, the Parser constructs the \emph{thought state}, which contains this extracted information. The thought state is then used to update the GRS accordingly.

\subsection{Scoring \& Validation}

Here, we verify whether a given LLM thought satisfies potential correctness conditions, and then we assign it a score. Depending on how the score is derived, the module may consult the LLM. Moreover, depending on the use case, the score may also be assigned by a human. Finally, use cases such as sorting use simple local scoring functions.

\subsection{Controller}

The {Controller} implements a specific strategy for selecting thoughts from its GRS structure. It also selects what transformations should be applied to which thoughts, and then passes this information to the Prompter. It also decides whether the whole process should be finalized, or whether the next round of interaction with the LLM should be initiated. In our current design, this is dictated by the execution plan specified in the GoO.

\subsection{GoO \& GRS}

The user constructs a GoO instance, which prescribes the execution plan of thought operations. The GoO is a static structure that is constructed once, before the execution starts. Each operation object knows its predecessor and successor operations.
Then, during the execution, an instance of the GRS maintains the continually updated information about the LLM reasoning process. This includes which operation has been executed so far, the states of all the generated LLM thoughts, their validity and scores, and any other relevant information.

The above elements offer extensible \textbf{APIs}, enabling straightforward implementations of different prompting schemes. The APIs are outlines \textcolor{green}{in the green part} of Figure~\ref{fig:arch}, and detailed in the documentation. We also provide \textbf{examples of prompts} used by these operations and a corresponding GRS in the \textcolor{red}{red part} of Figure~\ref{fig:arch}.

\if 0
The operations available to the user when constructing GoO are \texttt{Generate(t,k)} (generate k new thoughts based on a thought t), \texttt{Improve(t)} (enhance a thought t), \texttt{Aggregate(t1,...,tk)} (combine thoughts t1, ..., tk), \texttt{Score(t)} (score a thought t), \texttt{Validate(t)} (validate the correctness of a thought t), and \texttt{KeepBest(N)} (preserve N best scoring thoughts so far; the user can also specify the scope of thoughts to select from). Moreover, each operation can be instructed to generate multiple thoughts for the same input, for the purpose of obtaining more samples. For example, this enables \texttt{Aggregate} to generate multiple combination outcomes. Each such outcome is recorded within GRS and scored individually.
%
Finally, to facilitate experiments, the user can provide LLM's input parameters (model used, temperature, max tokens, api keys, organization, etc.) and the LLM cost characteristics (prompt token cost, response token cost). The latter automatizes cost estimation when running experiments within \nameA.
\fi

%% file: cases.tex
\section{Example Use Cases}
\label{sec:cases}

We now describe several use cases of \nameA. We detail one use case (sorting)
and summarize the others.

\subsection{Sorting}
\label{sec:sorting}

We focus on the decomposition of the sorting use case and Graph of Operations, which are central for implementing and executing any workload within \nameA.

We consider sorting numbers 0--9 with duplicates.
The considered LLMs are unable to sort a sequence of such numbers correctly beyond a certain length consistently because duplicate counts do not match.

\begin{figure}[hbtp]
\centering
\vspaceSQ{-1em}
\includegraphics[width=1.0\columnwidth]{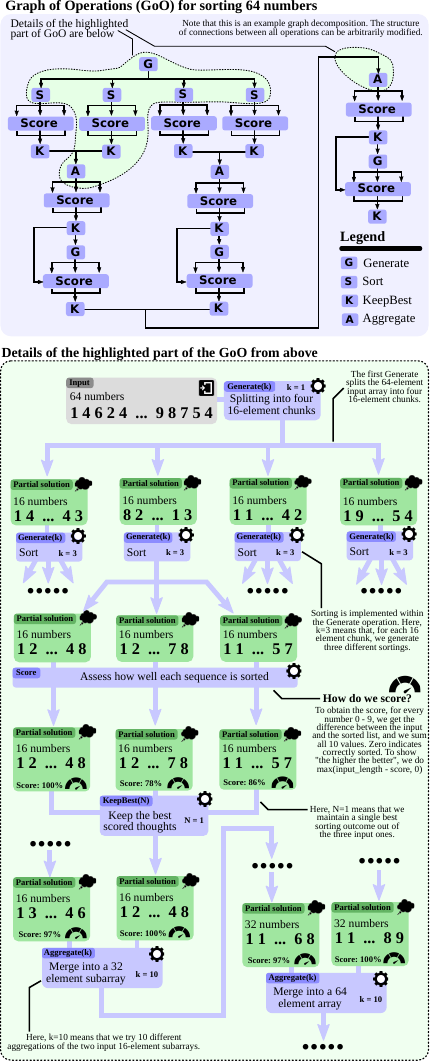}
\vspace{-1em}
\caption{An example graph decomposition of the sorting use case in \nameA. All used operations (Generate, Aggregate, Score, KeepBest) are described in Figure~\ref{fig:arch}.}
\vspaceSQ{-1em}
\label{fig:sorting}
\end{figure}

In \nameA, we employ merge-based sorting: First,  one decomposes the input sequence of numbers into subarrays. Then, one sorts these subarrays  individually, and then respectively merges them into a final solution.
Figure~\ref{fig:sorting} illustrates this use case together with its graph decomposition. Here, an LLM thought is a sequence of sorted numbers.
%
% Computing the score of such a thought is explained in Figure~\ref{fig:sorting} (the bottom-right part ``How do we score?'').

To score an outcome, denote an input sequence with $[a_1, a_2, ..., a_n]$ and an output one with $[b_1, b_2, ..., b_m]$.
We use the following score that determines ``the scope'' of errors:

$$
\text{error-scope} = X + Y
$$
where $p \in \{1,...,m\}$, $q \in \{1,...,n\}$, and
\begin{align}
X &= \sum_{i=1}^{m-1} \text{sgn}(\max(b_i - b_{i+1}, 0)) , \nonumber \\
Y &= \sum_{i=0}^{9} \left|\ |\{b_p : b_p = i\}| - |\{a_q : a_q = i\}|\ \right| \nonumber
\end{align}

Here, $X$ indicates how many consecutive pairs of numbers are incorrectly sorted. If two numbers $i$ and $i+1$ are incorrectly sorted (i.e., $b_i > b_{i+1}$), then the expression within the summation returns 1, increasing the error score by one. For two numbers correctly sorted, this expression amounts to 0.
Then, $Y$ determines how well a given output sequence preserves the frequency of output numbers. Specifically, for each considered number $x$ ($x \in \{0, ...,9\}$), we obtain the difference between the count of input elements being equal to $x$, vs.~the count of output elements equal to $x$. For an output sequence perfectly preserving the frequency of $x$, this would amount to 0. Any single ``deviation'' in this count, increases the ``error scope'' by 1. We then sum this over all considered values of $x$.
When plotting this score, to improve the clarity of plots, we additionally apply clipping $\min(\text{error-scope}, n)$, as some baselines (IO, CoT) result in large numbers of outliers with high error scope.
Finally, to use a ``positive score'' describing ``the scope of correctly sorted'' elements, one can use the value $\max(n - \text{error-scope}, 0)$.

\subsection{Set Operations}
\label{sec:set_ops}

Moreover, we also consider set operations, focusing on set intersection. They have numerous applications (particularly set intersection) in problems ranging from genome or document comparisons
to pattern matching~\cite{cook2006mining, schaeffer2007graph, jiang2013survey, besta2021graphminesuite, friggeri2011triangles, prat2012shaping, besta2022probgraph, besta2019communication}.
Set intersection of two sets is implemented similarly as the sorting. The second input set is split into subsets and the intersection of those subsets with the first input set is determined with the help of the LLM. Afterwards the resulting intersection sets are aggregated for the final results. For the evaluation we use different set sizes of 32, 64 and 128 elements and we vary the number of elements found in both sets to be between 25\% and 75\%. 

Our score indicates the total number of missing or incorrectly included elements in the final intersection. Specifically, denote two input sets with $A = [a_1, a_2, ..., a_n]$ and $B = [b_1, b_2, ..., b_n]$, and the output set with $C = [c_1, c_2, ..., c_m]$. Then,

$$
\text{error-scope} = X_1 + X_2 + X_d
$$
where $X_1 = |C \setminus (A \cap B)|$ are the number of elements in $C$ that are not supposed to be there, $X_2 = |(A \cap B) \setminus C|$ are the number of elements missing from $C$, and $X_d$ is the number of duplicates in $C$ (because the LLM expresses the set as a list in natural language).
%
% Similarly to the sorting case, when plotting this score, to improve the clarity of plots, we clip it with $\min(\text{error-scope}, n)$, excluding tails of outliers with high error scope in IO and CoT baselines.
%
%
Finally, to use a ``positive score'' describing ``the scope of correctly computed'' elements, one can use the value $\max(n - \text{error-scope}, 0)$.

\subsection{Keyword Counting}

Keyword counting finds the frequency of keywords in a given category (countries in our example implementation) within the input text. \nameAS splits the input text into multiple passages, counts the keywords in each passage and aggregates the subresults. The number of passages is configurable and can also be left to the LLM, making it possible to treat each sentence as a separate passage.
Here, to score a thought, we first -- for each keyword -- derive the absolute difference between the computed count and the correct one. We then sum all these differences to get the final score.

\subsection{Document Merging}
Finally, we also provide document merging.
Here, the goal is to generate a new Non-Disclosure Agreement (NDA) document based on several input ones that partially overlap in terms of their contents. The goal is to ensure minimal amount of duplication, while maximizing information retention. Document merging is broadly applicable in, e.g., legal procedures, where multiple sources of information have to be combined into a single document or article.
To score a solution, we query the LLM for two values (3 times for each value, and take the average). The first value corresponds to the solution redundancy (10 indicates no redundancy, 0 implies at least half the information is redundant), the second value stands for information retention (10 indicates all information is retained, 0 says that none is retained). We compute the harmonic mean of these values. 

\if0
$mean_redundancy = fmean(redundancy_scores)$
$mean_retain = fmean(retain_scores)$

$f1 = 2 * mean_redundancy * mean_retain / (mean_redundancy + mean_retain)$

% \maciej{Mention once we finally know what we have. Article filtering? Article summaries? Game24? Set operations? Generating reports? Code debugging? Creative writing?}
\fi

%% file: theory.tex
\section{The Latency-Volume Tradeoff}
\label{sec:theory}

We now show that \nameAS improves upon previous prompting schemes in terms of the tradeoff between latency (number of hops in the graph of thoughts to reach a given final thought) and \emph{volume}. We define volume -- for a given thought $t$ -- as \emph{the number of preceding LLM thoughts that could have impacted $t$}. Formally, the volume of $t$ is the number of thoughts from which there exists a path to $t$ in the graph of thoughts.
%
% In our analysis, we assume that (1) prompting an LLM within a session is sequential, but separate sessions can execute in parallel,
%
We assume that outputting a single thought costs $O(1)$ time and fix the total cost to $\Theta(n)$ for each prompting scheme.

%, and a unit of memory (within the context) is a single thought. 
%
%(which corresponds to a step in CoT or a vertex in ToT and \nameA). 
%
%We fix the size of each thought to be the same, and the capacity of the context to be $\Theta(N)$.
%
The structure of the schemes is as follows. CoT-SC consists of $k$ independent chains originating from a single starting thought. ToT is a complete $k$-ary tree. %Finally, in \nameA, the initial thought branches into $k$ children, and this continues for each child, for a given number of levels (as in the tree). Then, the ``leaf vertices'' are aggregated, also $k$ at a time (forming a ``mirrored'' version of the tree), until they converge within a single final thought that has no outgoing edges.
Finally, in \nameA , a complete $k$-ary tree is joined at its leaves with a ``mirrored'' $k$-ary tree of the same size but with its edges reversed.

The analysis is detailed in Table~\ref{tab:tradeoffs}. 
CoT offers a large volume of up to $N$, but at the cost of a high latency of $N$. CoT-SC reduces the latency by a factor of $k$ (which corresponds to its branching factor), but it simultaneously decreases the volume by $k$ as well. ToT offers a latency of $\log_k N$ but also has low volume. \nameAS is the only scheme to come with both a low latency of $\log_k N$ and a high volume $N$. This is enabled by the fact that \nameAS harnesses aggregations of thoughts, making it possible to reach the final thought from any other intermediate thought in the graph decomposition.

\input{table-tradeoffs}

\if 0
\subsection{Fixed Cost Per Session}

Assume a fixed context length (per session) that can contain up to $N$ thoughts (modeled as vertices of a graph $G$).
%
A CoT scheme is inherently sequential, and thus it takes up to $O(N)$ time to arrive at the output. Simultaneously, as it uses a single session, its total cost is $O(N)$.
%
While CoT-SC introduces multiple thought chains, each such chain is also sequential. Assuming a single chain runs within its own separate session, it similarly takes $O(N)$ time to arrive at the output. A total cost across all used $k$ sessions is then $O(k N)$.
%
Next, ToT schemes 
\fi

%% file: table-tradeoffs.tex
\begin{table}[h]
%\vspaceSQ{-1em}
\centering
\setlength{\tabcolsep}{3pt}
\footnotesize
\ifsq\renewcommand{\arraystretch}{0.7}\fi
\ifsq
%\small
\else
\footnotesize
\fi
%\sf 
\begin{tabular}{lll}
\toprule
\textbf{Scheme} & \textbf{Latency} & \textbf{Volume} \\
%
% \midrule
%
% \multicolumn{4}{c}{\textbf{\ul{Latency} \& \ul{volume} for a fixed (maximum) \ul{memory} $O(N)$, i.e., at most $N$ thoughts:}} \\
%
\midrule
Chain-of-Thought (CoT) & $N$  & $N$ \\
Self-Consistency with CoT (CoT-SC) & $N/k$  & $N/k$ \\
Tree of Thoughts (ToT) & $\log_k N$  & $O(\log_k N)$ \\
%Tree of Thoughts (ToT), variant~A & $\log_k N$ & $O(N)$ & $O(\log_k N)$ \\
%Tree of Thoughts (ToT), variant~B & $k \log_k N$ & $O(N)$ & $k \cdot O(\log_k N)$ \\
%
\midrule
\textbf{\nameS (\nameA)} &  $\log_k N$ & $N$ \\
\bottomrule
\end{tabular}
\vspace{-1em}
\caption{Comparison of prompting schemes, with respect to their fundamental tradeoff between latency and volume.
\textbf{\nameAS offers the best tradeoff.}}
\label{tab:tradeoffs}
\vspaceSQ{-0.5em}
\end{table}

%% file: eval.tex
\begin{figure*}[t]
\centering
%
%\vspaceSQ{-1em}
\includegraphics[width=0.7\textwidth]{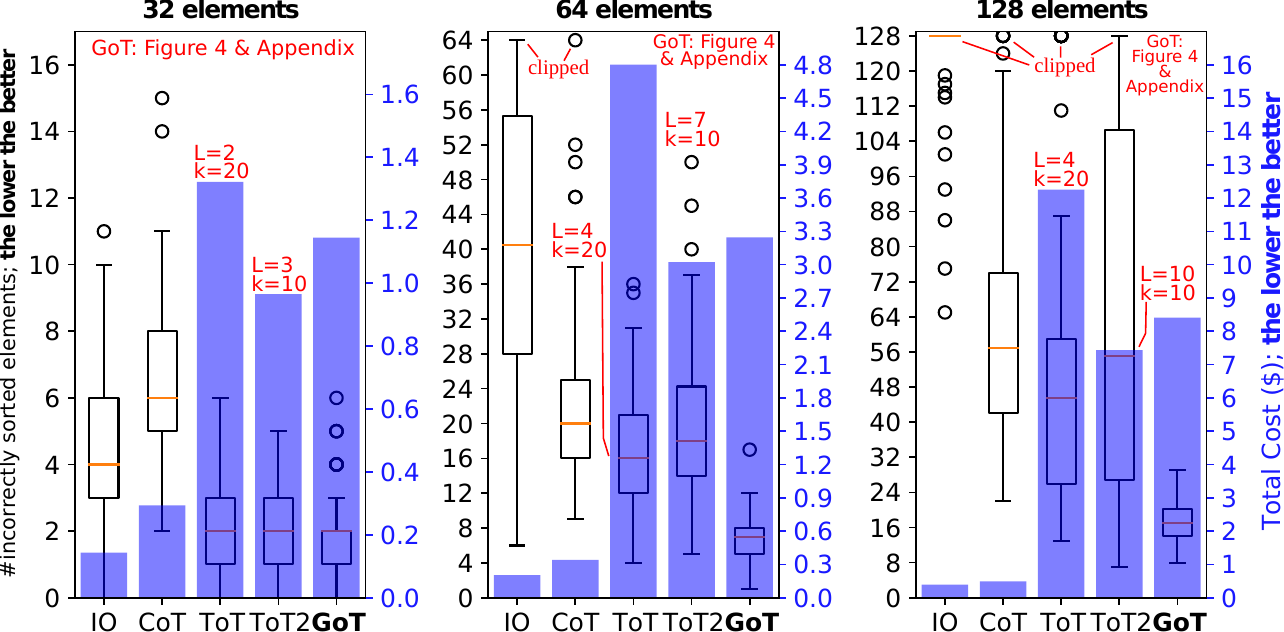}
\vspaceSQ{-0.5em}
\caption{Number of errors and cost in sorting tasks with ChatGPT-3.5. $L$ and $k$ indicate the structure of ToT (see Sections~\ref{sec:transf} and \ref{sec:theory}).}
%
%\vspaceSQ{-1em}
\label{fig:eval-sorting}
\vspaceSQ{-0.5em}
\end{figure*}

\section{Evaluation}

We show the advantages of \nameAS over the state of the art. We focus on comparing \nameAS to ToT, as it was shown to consistently outperform other schemes. Still, for a broad comparison, we also experiment with IO, CoT, and CoT-SC. As our analysis results in a large evaluation space, we present representative results and omit data that does not bring relevant insights (e.g., CoT-SC).

\subsection{Evaluation Methodology}

We use 100 input samples for each task and comparison baseline. We set the temperature to 1.0 and use a 4k context size unless stated otherwise. For each experiment, we fix the numbers of thoughts in respective schemes to achieve similar costs in each experiment.

\textbf{Parameters}
We experiment extensively with the branching factor $k$ and the number of levels $L$ to ensure that we compare \nameAS to cost-effective and advantageous configurations. We plot two variants of ToT: one with higher $k$ and lower depth (ToT), the other with lower $k$ but higher $L$ (ToT2).
We usually aim to achieve a sweet spot in the tradeoff between sparser generation rounds (lower $k$) vs.~more rounds (larger $L$). Usually more responses per round is more expensive (e.g.,~80 vs.~60 total responses for Figure~\ref{fig:eval-keyword-tr} but \$6 vs.~\$3 costs).
We also try different problem sizes $P$ (e.g., in sorting, $P$ states how many numbers are to be sorted).

\textbf{Used LLMs}
Due to budget restrictions, we focus on GPT-3.5.
%
%However, for a limited number of tasks, we tried GPT-4, which followed similar outcome patterns. 
%
We also experimented with Llama-2, but it was usually worse than GPT-3.5 and also much slower to run, making it infeasible to obtain enough samples.

\subsection{Analysis of \nameA's Advantages}

\begin{figure*}[t]
\centering
\vspaceSQ{-1em}
\includegraphics[width=0.7\textwidth]{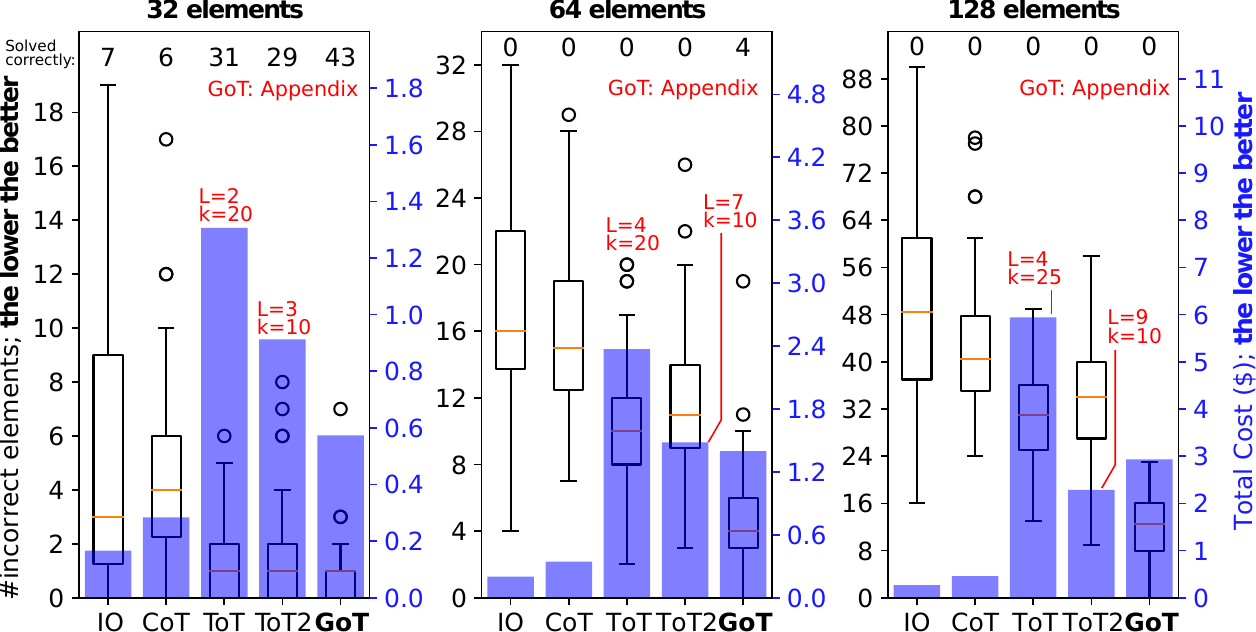}
\vspaceSQ{-0.5em}
\caption{Number of errors and cost in set intersection tasks with ChatGPT-3.5. $L$ and $k$ indicate the structure of ToT (see Sections~\ref{sec:transf} and \ref{sec:theory}).}
%
%\vspaceSQ{-1em}
\label{fig:eval-set-intersection}
\end{figure*}

\begin{figure}[h]
\centering
%
%\vspaceSQ{-1em}
\includegraphics[width=0.7\columnwidth]{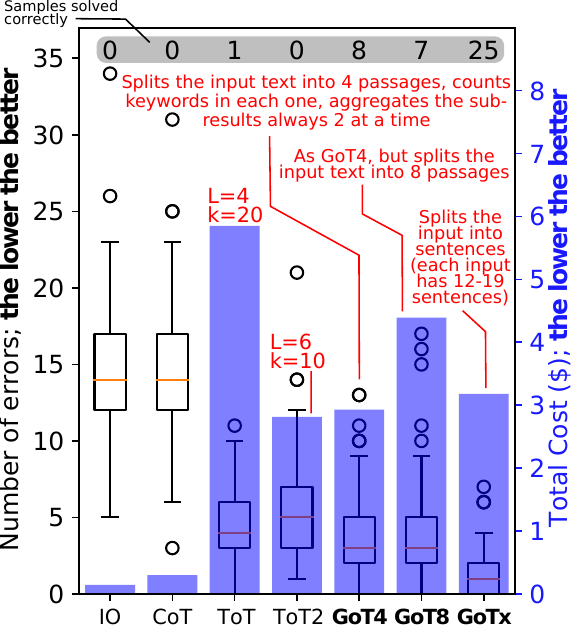}
\caption{Number of errors and cost in keyword counting with ChatGPT-3.5. $L$ and $k$ indicate the structure of ToT (see Sections~\ref{sec:transf} and \ref{sec:theory}).}
\label{fig:eval-keyword-tr}
\end{figure}

\begin{figure}[h]
\centering
\includegraphics[width=0.7\columnwidth]{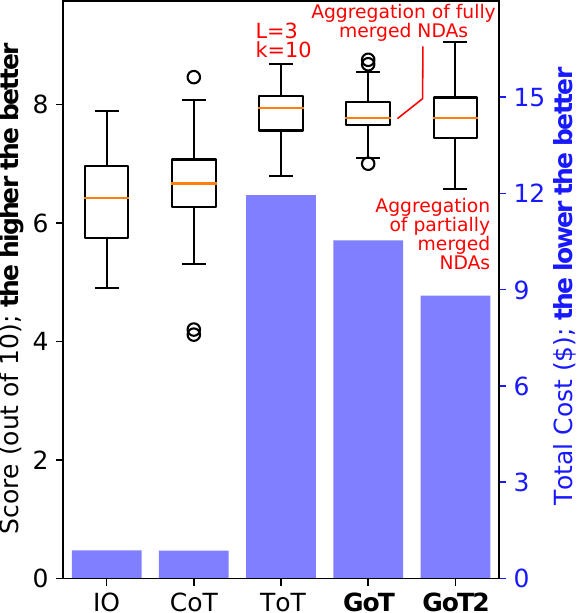}
\caption{Score and cost in document merging with ChatGPT-3.5. $L$ and $k$ indicate the structure of ToT (see Sections~\ref{sec:transf} and \ref{sec:theory}). Number of samples: 50; context size: 16k tokens.}
\label{fig:eval-nda-tr}
\end{figure}

The results of the analysis are in Figure~\ref{fig:eval-sorting} (sorting), \ref{fig:eval-set-intersection} (set intersection), \ref{fig:eval-keyword-tr} (keyword counting), and \ref{fig:eval-nda-tr} (document merging);
see Section~\ref{sec:cases} for the description of specific use cases.
\emph{Overall, \nameAS improves the quality of outcomes over all the considered baselines and it reduces inference costs compared to ToT}.

\textbf{\nameAS vs.~ToT}
\nameAS improves upon ToT and ToT2 by a large margin over all the considered problem instances. ToT usually comes with somewhat higher quality than ToT2, but simultaneously much higher costs. \nameA's costs are always lower than ToT, and comparable (in some cases lower, in others higher) to ToT2. For example, it reduces median error by $\approx$62\%, thereby achieving a higher quality of sorting, for $P=128$ in comparison to ToT while ensuring $>$31\% cost reductions. These advantages are due to \nameA's ability to decompose complex tasks into simpler subtasks, solve these subtasks independently, and then incrementally merge these outcomes into the final result.

\textbf{\nameAS vs.~IO and CoT}
\nameAS consistently delivers much higher quality of outcomes than IO/CoT. For example, for sorting ($P=64$), \nameA's median error is $\approx$65\% and $\approx$83\% lower than, respectively, CoT and IO. Yet, the costs of \nameAS -- and ToT -- are much higher than in IO and CoT. This is mostly due to our configuration of CoT, where we do not artificially inflate the lengths of the chains of reasoning if this does not improve the outcomes. The higher costs of \nameAS and ToT are driven by $k$ new thoughts built for each \texttt{Generate} operation; these multiple thoughts are one of the reasons for \nameA's superiority in quality.

\textbf{Increasing Complexity of Tackled Problems}
Most importantly, the advantages of \nameAS in the quality \emph{increase for all the baselines with the growing size of the problem $P$}. For example, in sorting, while for $P=32$ \nameAS only negligibly improves upon ToT2, its median error count becomes lower by $\approx$61\% for $P=64$ and $\approx$69\% for $P=128$. The quartiles also become respectively better. The results for other schemes also follow the intuition; for example, IO becomes consistently worse with the increasing $P$, which is expected as a single thought is unlikely to solve a large problem instance. \emph{Overall, this analysis illustrates that \nameAS is indeed well-suited for elaborate problem cases}, as the execution schedules usually become more complex with the growing problem sizes.

% We plot the results for different input set sizes (32, 64 and 128 elements) in Figure~\ref{fig:eval-set-intersection}. The number of elements found in both sets varies between 25\% and 75\%. The methods are implemented similarly as in the sorting use case.

\subsection{Discussion on Task Decomposition}

When splitting a task into subtasks and then solving these subtasks, the size of responses and the input (in tokens) are reduced proportionally to the degree of the task decomposition. However, the ``static'' part of the prompt (i.e., few-shot examples) may become a significant overhead (see \nameA4 to \nameA8 in Figure~\ref{fig:eval-keyword-tr}).
Here, we observe that these few-shot examples can usually also be reduced in size (e.g., the passages used to demonstrate keyword counting can also be made smaller and still be indicative of the actual input size), thus actively working towards decreasing the cost (e.g., see the difference between \nameA8 and {\nameA}x in Figure~\ref{fig:eval-keyword-tr}).

The overall goal when conducting graph decomposition is to break down a task to the point, where the LLM can solve it correctly for the majority of time using a single prompt (or with a few additional improvement steps). This significantly lowers the number of improvement/refinement steps needed during the later stages of the graph exploration. Furthermore, as indicated by our results, combining or concatenating subresults is usually an easier task than solving large task instances from scratch. Hence, the LLM is often successful when aggregating the final solution.

% Using fewer but ``wider'' \texttt{Generate} operations (e.g., $k$ = 20 as for ToT instead of $k$ = 10 as in ToT2) results in fewer but larger steps towards the final solution. This should benefit problems where the exploration space is large and should also ensure a larger verity in relation to the completion token count.
%
% can be significantly longer maybe => higher costs

\if 0
\subsubsection{Degree-Diameter Tradeoff}

How does the accuracy change when the graph structure changes? Most importantly, what is the impact of graph diameter and maximum degree?
\maciej{Intuitively, for growing of both, it should decrease - some "balance" should be achieved for both being not too small nor too large? Very low diameter: large max degree (for fixed N); very low degree: large diam (for fixed N)}

tot generates 20 instead of 10 answers for each generation prompt, but in turn has fewer "layers"/generate operations (doesn't try to improve sorting as often). We tried to have tot2 roughly at the same price point as tag and similarly tot (but there it was usually a bit more expensive as you know)
\fi

%% file: rw.tex
\section{Related Work}

We summarize relations between \nameAS and related work. %We now summarize those not mentioned in Section~\ref{sec:intro} nor in Table~\ref{tab:schemes}, and how they relate to \nameA.

\subsection{Prompting Paradigms \& Approaches}

We detail different prompting paradigms in Section~\ref{sec:intro} and Table~\ref{tab:schemes}. 
There are numerous other works related to prompting. We now briefly summarize selected most related ones; more extensive descriptions can be found in dedicated surveys~\cite{wang2023interactive, lertvittayakumjorn2021explanation, wang2021putting, hartmann2022survey}. Wang et al.~proposed Plan-and-Solve, an approach to enhance CoT with an explicit planning stage~\cite{wang2023plan}. Using complexity-based criteria to enhance prompting within a CoT was designed by Fu et al.~\cite{wang2022self, fu2022complexity}. The self-taught reasoner (STaR)~\cite{zelikman2022star} generates several chain of thoughts, and selects the ones that are valid. Similarly, a scheme by Shum et al.~\cite{shum2023automatic} generates a pool of CoT candidates, and selects the best candidate based on whether the candidates match the ground truth and on a policy gradient-based method. Automatic prompt generation overcomes the issues of scaling in CoT~\cite{shin2020autoprompt, li2021prefix, lester2021power}. Zhou et al.~propose to harness selecting the best prompt out of a candidate set~\cite{zhou2022large}.
Skeleon-of-Thought~\cite{ning2023skeleton} generates at first a number of skeleton answers (brief bullet points of 3 to 5 words) and expands on these points in parallel in a second step.

Finally, in prompt chaining, one cascades different LLMs. This enables prompting different LLMs via different contexts, enabling more powerful reasoning~\cite{creswell2022selection, nye2021show, wu2022ai, dohan2022language, qiao2022reasoning, wu2022promptchainer, wu2022ai}. \nameAS is orthogonal to this class of schemes, as it focuses on a single context capabilities.

\subsection{Self-Reflection \& Self-Evaluation}

Self-reflection and self-evaluation were introduced recently~\cite{shinn2023reflexion, paul2023refiner, madaan2023self, xie2023decomposition, zhu2023solving}. They are used to enhance different tasks, for example for code generation~\cite{chen2023teaching} or computer operation tasks~\cite{kim2023language}. In \nameA, we partially rely on self-evaluation when taking decisions on how to expand the graph of thoughts within a prompt.

\subsection{LLMs \& Planning}

There are many works recently on how to plan complex tasks with LLMs~\cite{huang2022language, zhang2023planning, yao2022react, yang2023foundation, wang2023describe, huang2022inner}. \nameAS could be seen as a generic framework that could potentially be used to enhance such schemes, by offering a paradigm for generating complex graph-based plans.

\if 0
\subsection{Combining LLMs with External Units}

Another part of the LLM landscape is combining LLMs with different additional tools/agents. Examples include AutoGPT~\cite{richards2023auto} (augmenting GPT-4 with additional modules including an execution agent and a memory unit), HuggingGPT~\cite{shen2023hugginggpt} (making the LLM responsible for selecting various tools), training external entities to check the LLM outcomes~\cite{cobbe2021training}, using search trees augmented with external text~\cite{liu2023llm+}, or relying on a classical external planner~\cite{schlag2023large}.
%
\nameAS could potentially be extended as a framework for harnessing the interactions between such interacting agents.
\fi

\subsection{Graphs and Graph Computing}
Graphs have become an immensely popular and important part of the general computing landscape~\cite{lumsdaine2007challenges, malewicz2010pregel, gregor2005parallel, gregor2005lifting, sakr2020future}. Recently, there has been a growing interest in domains such as graph databases~\cite{robinson2015graph, besta2019demystifying, besta2023high, besta2023gdi, besta2022neural}, graph pattern matching~\cite{fan2010graph, cheng2008fast, teixeira2015arabesque, besta2022probgraph, besta2021sisa, besta2021graphminesuite}, graph streaming~\cite{feng2015distinger, dhulipala2019low, besta2019practice}, and graph machine learning as well as graph neural networks~\cite{hamilton2017representation, wu2020comprehensive, zhou2020graph, zhang2020deep, chami2020machine, hamilton2017representation, bronstein2017geometric, besta2021motif, gianinazzi2021learning, scarselli2008graph, besta2022parallel}. The graph abstraction has been fruitful for many modern research domains, such as social sciences (e.g., studying human interactions), bioinformatics (e.g., analyzing protein structures), chemistry (e.g., designing chemical compounds), medicine (e.g., drug discovery), cybersecurity (e.g., identifying intruder machines), healthcare (e.g., exposing groups of people who submit fraudulent claims), web graph analysis (e.g., providing accurate search services), entertainment services (e.g., predicting movie popularity), linguistics (e.g., modeling relationships between words), transportation (e.g., finding efficient routes), physics (e.g., understanding phase transitions and critical phenomena), and many others~\cite{lumsdaine2007challenges, cook2006mining, jiang2013survey, horvath2004cyclic, chakrabarti2006graph}.
In this work, we harness the graph abstraction as a key mechanism that enhances prompting capabilities in LLMs.

%% file: conc.tex
\section{Conclusion}

Prompt engineering is one of the central new domains of the large language model (LLM) research. It enables using LLMs efficiently, without any model updates. However, designing effective prompts is a challenging task.

In this work, we propose \nameS (\nameA), a new paradigm that enables the LLM to solve different tasks effectively without any model updates. The key idea is to model the LLM reasoning as an arbitrary graph, where thoughts are vertices and dependencies between thoughts are edges. This enables novel transformations of thoughts, such as aggregation. Human's task solving is often non-linear, and it involves combining intermediate solutions into final ones, or changing the flow of reasoning upon discovering new insights. \nameAS reflects this with its graph structure.
%
% It also brings the LLM reasoning closer to that of a human being. 

\nameAS outperforms other prompting schemes, for example ensuring 62\% increase in the quality of sorting over ToT, while simultaneously reducing costs by $>$31\%. 
%
% Even more importantly, \nameA's generic formulation based on arbitrary graphs opens a way to future novel powerful reasoning approaches that will benefit from the richness of graph structures.
%
We also propose a novel metric for a prompting scheme, the volume of a thought, to indicate the scope of information that a given LLM output could carry with it, where \nameAS also excels. This provides a step towards more principled prompt engineering.

The graph abstraction has been the foundation of several successful designs in computing and AI over last decades, for example AlphaFold for protein predictions. Our work harnesses it within the realm of prompt engineering.

%% file: acks.tex
\small

\section*{Acknowledgements}

We thank Hussein Harake, Colin McMurtrie, Mark Klein, Angelo Mangili, and the whole CSCS team granting access to the Ault and Daint machines, and for their excellent technical support. 
We thank Timo Schneider for help with infrastructure at SPCL.
This project received funding from the European Research Council (Project PSAP, No.~101002047), and the European High-Performance Computing Joint Undertaking (JU) under grant agreement No.~955513 (MAELSTROM). This project was supported by the ETH Future Computing Laboratory (EFCL), financed by a donation from Huawei Technologies. This project received funding from the European Union’s HE research and innovation programme under the grant agreement No. 101070141 (Project GLACIATION).

\normalsize

%% file: appendix.tex
\appendix

\section{Positive Score Evaluation}

The following figures plot the same data as Figures~\ref{fig:eval-sorting} and \ref{fig:eval-set-intersection} respectively, however use the "positive score" described in Sections~\ref{sec:sorting} and \ref{sec:set_ops}.

\begin{figure}[h]
\centering
\includegraphics[width=0.975\columnwidth]{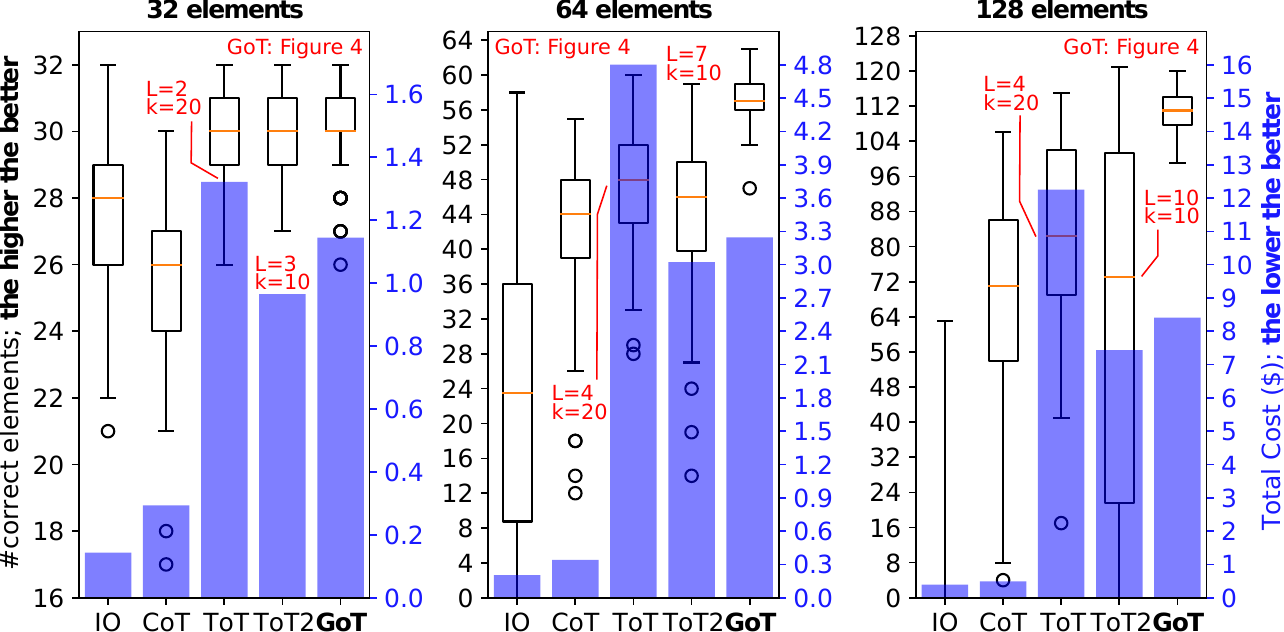}
\caption{Accuracy and cost in sorting tasks with ChatGPT-3.5. $L$ and $k$ indicate the structure of ToT (see Sections~\ref{sec:transf} and \ref{sec:theory}).}
\label{fig:eval-sorting-positive}
\end{figure}

\begin{figure}[h]
\centering
\includegraphics[width=0.975\columnwidth]{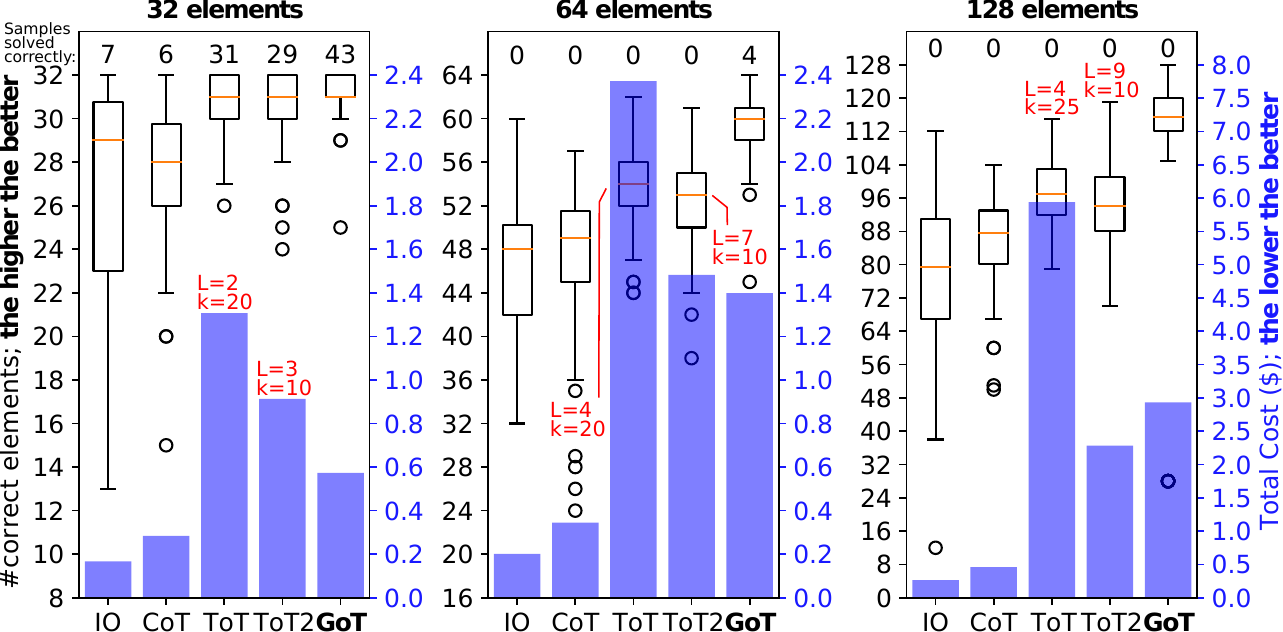}
\caption{Accuracy and cost in set intersection with ChatGPT-3.5. $L$ and $k$ indicate the structure of ToT (see Sections~\ref{sec:transf} and \ref{sec:theory}).}
\label{fig:eval-set-intersection-positive}
\end{figure}

\section{Example Prompts - Sorting}
\definecolor{mygreen}{RGB}{49,127,56}
\definecolor{myblue}{RGB}{40,0,249}
We present the prompts only for the sorting of 32-element lists, as those for 64-element and 128-element lists are identical, except for the \textit{split\_prompt} where the number of elements in the one-shot example matches the problem size.

For sorting, we employ three distinct types of operations that interact with the LLM, each with its corresponding prompts. 
First, there is the \textbf{Generate} operation, utilizing the \textit{sort\_prompt} to guide the LLM in sorting a provided list of values, and the \textit{split\_prompt} to direct the LLM to split a specified list into a designated number of sublists. 
Next, the \textbf{Improve} operation employs the \textit{improve\_prompt} to instruct the LLM to refine a sorted list if it detects mistakes. 
Finally, the \textbf{Aggregate} operation leverages the \textit{merge\_prompt} to guide the LLM in merging two pre-sorted lists into a single sorted list.

First, we present the prompt stubs (Table \ref{tab:sorting_prompt_stubs}), serving as templates to dynamically generate appropriate prompts at runtime. For clarity, we display their corresponding few-shot examples separately in Table \ref{tab:sorting_few_shots}. Following this, we outline the LLM interactions throughout the process of solving the sorting use case (Table \ref{tab:sorting_example_goo} - Table \ref{tab:sorting_example_part4}).

%Some notable observations are that most of the results for the initial sort\_prompt are identical, despite being wrong, and the reasoning step for the improve\_prompt was either missing or wrong and only rarely right, while still overall helping to actually improve the  result. 
%\subsection{Prompt Stubs}

\begin{table*}
    \centering
    \renewcommand{\arraystretch}{1.5}
    \caption{Prompt stubs for the sorting tasks; parameters in single curly brackets will be substituted at runtime.}
    \label{tab:sorting_prompt_stubs}
    \begin{tabularx}{\textwidth}{X}
    \hline
    \textbf{sort\_prompt:} \textless{}Instruction\textgreater{} Sort the following list of numbers in ascending order. Output only the sorted list of numbers, no additional text. \textless{}/Instruction\textgreater{}\\
    \textless{}Examples\textgreater{} See Table~\ref{tab:sorting_few_shots} \textless{}/Examples\textgreater{}\\
    Input: \{input\_list\}\\
    \hline
    \textbf{split\_prompt (32 elements):} \textless{}Instruction\textgreater{} Split the following list of 32 numbers into 2 lists of 16 numbers each, the first list should contain the first 16 numbers and the second list the second 16 numbers.\\
    Only output the final 2 lists in the following format without any additional text or thoughts!:\\
    \{\{ \\
    \quad \texttt{"List 1": [3, 4, 3, 5, 7, 8, 1, ...],}\\
    \quad \texttt{"List 2": [2, 9, 2, 4, 7, 1, 5, ...]}\\
    \}\} \\ 
    \textless{}/Instruction\textgreater{}\\
    \textless{}Examples\textgreater{} See Table~\ref{tab:sorting_few_shots} \textless{}/Examples\textgreater{}\\
    Input: \{input\_list\}\\
    \hline
    \textbf{improve\_prompt:} \textless{}Instruction\textgreater{} The following two lists represent an unsorted list of numbers and a sorted variant of that list. 
    The sorted variant is not correct. Fix the sorted variant so that it is correct.
    Make sure that the output list is sorted in ascending order, has the same number of elements as the input list (\{length\}), and contains the same elements as the input list.\textless{}/Instruction\textgreater{}\\
    \textless{}Approach\textgreater{}\\
    To fix the incorrectly sorted list follow these steps:
    \begin{enumerate}
        \item For each number from 0 to 9, compare the frequency of that number in the incorrectly sorted list to the frequency of that number in the input list.
        \item Iterate through the incorrectly sorted list and add or remove numbers as needed to make the frequency of each number in the incorrectly sorted list match the frequency of that number in the input list.
    \end{enumerate}
    \textless{}/Approach\textgreater{}\\
    \textless{}Examples\textgreater{} See Table~\ref{tab:sorting_few_shots} \textless{}/Examples\textgreater{}\\
    Input: \{input\_list\}\\
    Incorrectly Sorted: \{sorted\_list\}\\
    \hline
    \textbf{merge\_prompt:} \textless{}Instruction\textgreater{} Merge the following 2 sorted lists of length \{length\} each, into one sorted list of length \{length\_combined\} using a merge sort style approach.
    Only output the final merged list without any additional text or thoughts!: \textless{}/Instruction\textgreater{}\\
    \textless{}Approach\textgreater{}\\
    To merge the two lists in a merge-sort style approach, follow these steps:
    \begin{enumerate}
        \item Compare the first element of both lists.
        \item Append the smaller element to the merged list and move to the next element in the list from which the smaller element came.
        \item Repeat steps 1 and 2 until one of the lists is empty.
        \item Append the remaining elements of the non-empty list to the merged list.
    \end{enumerate}
    \textless{}/Approach\textgreater{}\\
    Merge the following two lists into one sorted list:
    \begin{enumerate}
        \item \{input\_list1\}
        \item \{input\_list2\}
    \end{enumerate}
    Merged list:\\
    \hline
    \end{tabularx}
\end{table*}
%\subsection{Few-Shot Examples}
\definecolor{grey}{rgb}{0.52, 0.52, 0.51}
\begin{table*}
    \centering
    \renewcommand{\arraystretch}{1.5}
    \caption{Few-shot examples for each prompt used for the sorting tasks; some lists are truncated for brevity.}
    \label{tab:sorting_few_shots}
    \begin{tabularx}{\textwidth}{X}
    \hline
    \textbf{sort\_prompt:} \\
    \textless{}Examples\textgreater{} \\
    \texttt{Input: [5, 1, 0, 1, 2, 0, 4, 8, 1, 9, 5, 1, 3, 3, 9, 7]}\\
    \texttt{Output: [0, 0, 1, 1, 1, 1, 2, 3, 3, 4, 5, 5, 7, 8, 9, 9]}\\\\
    \texttt{Input: [3, 7, 0, 2, 8, 1, 2, 2, 2, 4, 7, 8, 5, 5, 3, 9, 4, 3,} \textcolor{gray}{$\dots$ (Omitted 14/32 numbers)}\texttt{]}\\
    \texttt{Output: [0, 0, 1, 1, 2, 2, 2, 2, 2, 2, 3, 3, 3, 3, 3, 4, 4, 4,} \textcolor{gray}{ $\dots$ (Omitted 14/32 numbers)}\texttt{]}\\\\
    \texttt{Input: [4, 4, 9, 7, 9, 7, 0, 0, 4, 9, 1, 7, 9, 5, 8, 7, 5, 6,} \textcolor{gray}{ $\dots$ (Omitted 46/64 numbers)}\texttt{]}\\
    \texttt{Output: [0, 0, 0, 0, 0, 0, 0, 1, 1, 2, 2, 2, 2, 2, 2, 3, 3, 3,} \textcolor{gray}{ $\dots$ (Omitted 46/64 numbers)}\texttt{]}\\
    \textless{}/Examples\textgreater{}\\
    \hline
    \textbf{split\_prompt (32 elements):} \\
    \textless{}Examples\textgreater{} \\
    \texttt{Input: [9, 6, 7, 7, 2, 0, 2, 2, 3, 5, 0, 9, 2, 2, 4, 4, 5, 2,} \textcolor{gray}{ $\dots$ (Omitted 14/32 numbers)}\texttt{]}\\
    \texttt{Output:} \\
    \{\{\\
        \quad \texttt{"List 1": [9, 6, 7, 7, 2, 0, 2, 2, 3, 5, 0, 9, 2, 2, 4, 4],}\\
        \quad \texttt{"List 2": [5, 2, 5, 1, 2, 8, 3, 8, 3, 9, 6, 0, 4, 2, 2, 3]}\\
    \}\}\\
    \textless{}/Examples\textgreater{}\\
    \hline
    \textbf{improve\_prompt:} \\
    \textless{}Examples\textgreater{} \\
    \texttt{Input: [3, 7, 0, 2, 8, 1, 2, 2, 2, 4, 7, 8, 5, 5, 3, 9]}\\
    \texttt{Incorrectly Sorted: [0, 0, 0, 0, 0, 1, 2, 2, 3, 3, 4, 4, 4, 5, 5, 7, 7, 8, 8, 9, 9, 9, 9]}\\
    \texttt{Reason: The incorrectly sorted list contains four extra 0s, two extra 4s and three extra 9s and is missing two 2s.}\\
    \texttt{Output: [0, 1, 2, 2, 2, 2, 3, 3, 4, 5, 5, 7, 7, 8, 8, 9]}\\\\
    \texttt{Input: [6, 4, 5, 7, 5, 6, 9, 7, 6, 9, 4, 6, 9, 8, 1, 9, 2, 4,} \textcolor{gray}{ $\dots$ (Omitted 14/32 numbers)}\texttt{]}\\
    \texttt{Incorrectly Sorted: [0, 1, 1, 2, 2, 3, 4, 4, 4, 4, 4, 5, 5, 5, 5, 6, 6, 6,} \textcolor{gray}{ $\dots$ (Omitted 14/32 numbers)}\texttt{]}\\
    \texttt{Reason: The incorrectly sorted list contains two extra 4s and is missing two 6s and one 9.}\\
    \texttt{Output: [0, 1, 1, 2, 2, 3, 4, 4, 4, 5, 5, 5, 5, 6, 6, 6, 6, 6,} \textcolor{gray}{ $\dots$ (Omitted 14/32 numbers)}\texttt{]}\\\\
    \texttt{Input: [4, 4, 9, 7, 9, 7, 0, 0, 4, 9, 1, 7, 9, 5, 8, 7, 5, 6,} \textcolor{gray}{ $\dots$ (Omitted 46/64 numbers)}\texttt{]}\\
    \texttt{Incorrectly Sorted: [0, 0, 0, 0, 0, 0, 0, 1, 1, 2, 2, 2, 2, 3, 3, 3, 4, 4,} \textcolor{gray}{ $\dots$ (Omitted 46/64 numbers)}\texttt{]}\\
    \texttt{Reason: The incorrectly sorted list contains one extra 8 and is missing two 2s, one 3, three 4s, two 5s, one 6, six 7s and one 9.}\\
    \texttt{Output: [0, 0, 0, 0, 0, 0, 0, 1, 1, 2, 2, 2, 2, 2, 2, 3, 3, 3,} \textcolor{gray}{ $\dots$ (Omitted 46/64 numbers)}\texttt{]}\\
    \textless{}/Examples\textgreater{}\\
    \hline
    \textbf{merge\_prompt:} -\\
    \hline
    \end{tabularx}
\end{table*}
%\subsection{Example}

\begin{table*}
    \centering
    \renewcommand{\arraystretch}{1.5}
    \caption{Sorting of a 32 element list: Execution plan (\textbf{GoO})}
    \label{tab:sorting_example_goo}
    \begin{tabularx}{\textwidth}{X}
    \hline
    \textbf{GoO:} 
    \begin{enumerate}
        \item Split the input list into two sub-lists of equal size (split\_prompt)
        \item For each sub-list: Sort the sub-list (sort\_prompt) five times; score each sort attempt; keep the best
        \item Merge the sorted sub-lists into one fully sorted list (merge\_prompt) 10 times; score each merge attempt; keep the best
        \item Fix any potential mistakes in the sorted list (improve\_prompt) 10 times; score each improvement attempt; keep the best
    \end{enumerate}\\
    \hline
    \end{tabularx}
\end{table*}

\begin{table*}
    \centering
    \renewcommand{\arraystretch}{1.5}
    \caption{Sorting of a 32 element list: \textbf{Step 1 (Prompt/Response)}}
    \label{tab:sorting_example_part1}
    \begin{tabularx}{\textwidth}{X}
    \hline
    \textbf{Step 1 -- Prompt:} \\
    \textless{}Instruction\textgreater{} Split the following list of 32 numbers into 2 lists of 16 numbers each, the first list should contain the first 16 numbers and the second list the second 16 numbers.\\
    Only output the final 2 lists in the following format without any additional text or thoughts!:\\
    \{\{ \\
    \quad \texttt{"List 1": [3, 4, 3, 5, 7, 8, 1, ...],}\\
    \quad \texttt{"List 2": [2, 9, 2, 4, 7, 1, 5, ...]}\\
    \}\} \\ 
    \textless{}/Instruction\textgreater{}\\
    \textless{}Examples\textgreater{} See Table~\ref{tab:sorting_few_shots} \textless{}/Examples\textgreater{}\\
    Input: \texttt{[8, 7, 1, 1, 1, 1, 3, 3, 0, 9, 4, 1, 0, 2, 5, 1, 0, 5, 6, 7, 1, 4, 5, 9, 4, 6, 2, 5, 8, 6, 2, 6]}\\
    \textbf{Step 1 -- Response \textcolor{mygreen}{(Fully Correct)}:} \\
    %"{\n    \"List 1\": [8, 7, 1, 1, 1, 1, 3, 3, 0, 9, 4, 1, 0, 2, 5, 1],\n    \"List 2\": [0, 5, 6, 7, 1, 4, 5, 9, 4, 6, 2, 5, 8, 6, 2, 6]\n}"
    \{\{\\
        \quad \texttt{"List 1": [8, 7, 1, 1, 1, 1, 3, 3, 0, 9, 4, 1, 0, 2, 5, 1],}\\
        \quad \texttt{"List 1": [0, 5, 6, 7, 1, 4, 5, 9, 4, 6, 2, 5, 8, 6, 2, 6]}\\
    \}\}\\
    \hline
    \end{tabularx}
\end{table*}

\begin{table*}
    \centering
    \renewcommand{\arraystretch}{1.5}
    \caption{Sorting of a 32 element list: \textbf{Step 2 (Prompts/Responses)}}
    \label{tab:sorting_example_part2}
    \begin{tabularx}{\textwidth}{X}
    \hline
    \textbf{Step 2a -- Prompt:} \\
    \textless{}Instruction\textgreater{} Sort the following list of numbers in ascending order. Output only the sorted list of numbers, no additional text. \textless{}/Instruction\textgreater{}\\
    \textless{}Examples\textgreater{} See Table~\ref{tab:sorting_few_shots} \textless{}/Examples\textgreater{}\\
    Input: \texttt{[8, 7, 1, 1, 1, 1, 3, 3, 0, 9, 4, 1, 0, 2, 5, 1]}\\
    \textbf{Step 2a -- 5 Responses:}
    % '[0, 0, 1, 1, 1, 1, 1, 1, 2, 3, 3, 4, 5, 7, 8, 9]', '[0, 0, 1, 1, 1, 1, 1, 2, 3, 3, 4, 5, 7, 8, 9]', '[0, 0, 1, 1, 1, 1, 1, 2, 3, 3, 4, 5, 7, 8, 9]', '[0, 0, 1, 1, 1, 1, 1, 2, 3, 3, 4, 5, 7, 8, 9]', '[0, 0, 1, 1, 1, 1, 1, 2, 3, 3, 4, 5, 7, 8, 9]'
    \begin{enumerate}
        \item \textbf{\texttt{[0, 0, 1, 1, 1, 1, 1, 1, 2, 3, 3, 4, 5, 7, 8, 9]} \textcolor{mygreen}{(Fully Correct)}}
        \item \texttt{[0, 0, 1, 1, 1, 1, 1, 2, 3, 3, 4, 5, 7, 8, 9]} \textcolor{myblue}{(1 Error - Missing one 1)}
        \item \texttt{[0, 0, 1, 1, 1, 1, 1, 2, 3, 3, 4, 5, 7, 8, 9]} \textcolor{myblue}{(1 Error - Missing one 1)}
        \item \texttt{[0, 0, 1, 1, 1, 1, 1, 2, 3, 3, 4, 5, 7, 8, 9]} \textcolor{myblue}{(1 Error - Missing one 1)}
        \item \texttt{[0, 0, 1, 1, 1, 1, 1, 2, 3, 3, 4, 5, 7, 8, 9]} \textcolor{myblue}{(1 Error - Missing one 1)}
    \end{enumerate}\\
    \textbf{Step 2b -- Prompt:} \\
    \textless{}Instruction\textgreater{} Sort the following list of numbers in ascending order. Output only the sorted list of numbers, no additional text. \textless{}/Instruction\textgreater{}\\
    \textless{}Examples\textgreater{} See Table~\ref{tab:sorting_few_shots} \textless{}/Examples\textgreater{}\\
    Input: \texttt{[0, 5, 6, 7, 1, 4, 5, 9, 4, 6, 2, 5, 8, 6, 2, 6]}\\
    \textbf{Step 2b -- 5 Responses:}
    % '[0, 1, 2, 2, 4, 4, 5, 5, 5, 6, 6, 6, 7, 8, 9]', '[0, 1, 2, 2, 4, 4, 5, 5, 5, 6, 6, 6, 7, 8, 9]', '[0, 1, 2, 2, 4, 4, 5, 5, 5, 6, 6, 6, 7, 8, 9]', '[0, 1, 2, 2, 4, 4, 5, 5, 5, 6, 6, 6, 6, 7, 8, 9]', '[0, 1, 2, 2, 4, 4, 5, 5, 5, 6, 6, 6, 7, 8, 9]'
    \begin{enumerate}
        \item \texttt{[0, 1, 2, 2, 4, 4, 5, 5, 5, 6, 6, 6, 7, 8, 9]} \textcolor{myblue}{(1 Error - Missing one 6)}
        \item \texttt{[0, 1, 2, 2, 4, 4, 5, 5, 5, 6, 6, 6, 7, 8, 9]} \textcolor{myblue}{(1 Error - Missing one 6)}
        \item \texttt{[0, 1, 2, 2, 4, 4, 5, 5, 5, 6, 6, 6, 7, 8, 9]} \textcolor{myblue}{(1 Error - Missing one 6)}
        \item \textbf{\texttt{[0, 1, 2, 2, 4, 4, 5, 5, 5, 6, 6, 6, 6, 7, 8, 9]} \textcolor{mygreen}{(Fully Correct)}}
        \item \texttt{[0, 1, 2, 2, 4, 4, 5, 5, 5, 6, 6, 6, 7, 8, 9]} \textcolor{myblue}{(1 Error - Missing one 6)}
    \end{enumerate}\\
    \hline
    \end{tabularx}
\end{table*}

\begin{table*}
    \centering
    \renewcommand{\arraystretch}{1.5}
    \caption{Sorting of a 32 element list: \textbf{Step 3 (Prompt/Responses)}}
    \label{tab:sorting_example_part3}
    \begin{tabularx}{\textwidth}{X}
    \hline
    \textbf{Step 3 -- Prompt:} \\
    \textless{}Instruction\textgreater{} Merge the following 2 sorted lists of length 16 each, into one sorted list of length 32 using a merge sort style approach.
    Only output the final merged list without any additional text or thoughts!: \textless{}/Instruction\textgreater{}\\
    \textless{}Approach\textgreater{}\\
    To merge the two lists in a merge-sort style approach, follow these steps:
    \begin{enumerate}
        \item Compare the first element of both lists.
        \item Append the smaller element to the merged list and move to the next element in the list from which the smaller element came.
        \item Repeat steps 1 and 2 until one of the lists is empty.
        \item Append the remaining elements of the non-empty list to the merged list.
    \end{enumerate}
    \textless{}/Approach\textgreater{}\\
    Merge the following two lists into one sorted list:
    \begin{enumerate}
        \item \texttt{[0, 0, 1, 1, 1, 1, 1, 1, 2, 3, 3, 4, 5, 7, 8, 9]}
        \item \texttt{[0, 1, 2, 2, 4, 4, 5, 5, 5, 6, 6, 6, 6, 7, 8, 9]}
    \end{enumerate}
    Merged list:\\
    \textbf{Step 3 -- 10 Responses:}
    % ['[0, 0, 0, 1, 1, 1, 1, 1, 1, 2, 2, 2, 3, 3, 4, 4, 4, 5, 5, 5, 6, 6, 6, 6, 7, 7, 8, 8, 9, 9]', '[0, 0, 0, 1, 1, 1, 1, 1, 2, 2, 2, 3, 3, 4, 4, 4, 5, 5, 5, 5, 6, 6, 6, 6, 7, 7, 8, 8, 9, 9]', '[0, 0, 0, 1, 1, 1, 1, 1, 1, 2, 2, 2, 3, 3, 4, 4, 4, 5, 5, 5, 5, 6, 6, 6, 6, 6, 7, 7, 8, 8, 9, 9]', '[0, 0, 0, 1, 1, 1, 1, 1, 1, 2, 2, 2, 3, 3, 4, 4, 4, 5, 5, 5, 5, 6, 6, 6, 6, 6, 7, 7, 8, 8, 9]', '[0, 0, 0, 1, 1, 1, 1, 1, 1, 2, 2, 2, 3, 3, 4, 4, 4, 5, 5, 5, 5, 6, 6, 6, 6, 6, 7, 7, 8, 8, 9, 9]', '[0, 0, 0, 1, 1, 1, 1, 1, 1, 2, 2, 2, 3, 3, 4, 4, 4, 5, 5, 5, 6, 6, 6, 6, 7, 7, 8, 8, 9, 9]', '[0, 0, 0, 1, 1, 1, 1, 1, 1, 2, 2, 2, 3, 3, 4, 4, 4, 5, 5, 5, 5, 6, 6, 6, 6, 6, 7, 7, 8, 8, 9, 9]', '[0, 0, 0, 1, 1, 1, 1, 1, 1, 2, 2, 2, 3, 3, 4, 4, 4, 5, 5, 5, 5, 6, 6, 6, 6, 7, 7, 8, 8, 9, 9]', '[0, 0, 0, 1, 1, 1, 1, 1, 1, 2, 2, 2, 3, 3, 4, 4, 4, 5, 5, 5, 5, 6, 6, 6, 6, 7, 7, 8, 8, 9, 9]', '[0, 0, 0, 1, 1, 1, 1, 1, 1, 2, 2, 2, 3, 3, 4, 4, 4, 5, 5, 5, 6, 6, 6, 6, 7, 7, 8, 8, 9, 9]'
    \begin{enumerate}
        \item \texttt{[0, 0, 0, 1, 1, 1, 1, 1, 1, 2, 2, 2, 3, 3, 4, 4, 4, 5, 5, 5, 6, 6, 6, 6, 7, 7, 8, 8, 9, 9]}
        \linebreak \textcolor{myblue}{(2 Errors - Missing one 1 and one 5)}
        \item \texttt{[0, 0, 0, 1, 1, 1, 1, 1, 2, 2, 2, 3, 3, 4, 4, 4, 5, 5, 5, 5, 6, 6, 6, 6, 7, 7, 8, 8, 9, 9]}
        \linebreak \textcolor{myblue}{(2 Errors - Missing two 1s)}
        \item \texttt{[0, 0, 0, 1, 1, 1, 1, 1, 1, 2, 2, 2, 3, 3, 4, 4, 4, 5, 5, 5, 5, 6, 6, 6, 6, 6, 7, 7, 8, 8, 9, 9]}
        \linebreak \textcolor{myblue}{(2 Errors - Missing one 1 and got one extra 6)}
        \item \texttt{[0, 0, 0, 1, 1, 1, 1, 1, 1, 2, 2, 2, 3, 3, 4, 4, 4, 5, 5, 5, 5, 6, 6, 6, 6, 6, 7, 7, 8, 8, 9]}
        \linebreak \textcolor{myblue}{(3 Errors - Missing one 1, one 9 and got one extra 6)}
        \item \texttt{[0, 0, 0, 1, 1, 1, 1, 1, 1, 2, 2, 2, 3, 3, 4, 4, 4, 5, 5, 5, 5, 6, 6, 6, 6, 6, 7, 7, 8, 8, 9, 9]}
        \linebreak \textcolor{myblue}{(2 Errors - Missing one 1 and got one extra 6)}
        \item \texttt{[0, 0, 0, 1, 1, 1, 1, 1, 1, 2, 2, 2, 3, 3, 4, 4, 4, 5, 5, 5, 6, 6, 6, 6, 7, 7, 8, 8, 9, 9]}
        \linebreak \textcolor{myblue}{(2 Errors - Missing one 1 and one 5)}
        \item \texttt{[0, 0, 0, 1, 1, 1, 1, 1, 1, 2, 2, 2, 3, 3, 4, 4, 4, 5, 5, 5, 5, 6, 6, 6, 6, 6, 7, 7, 8, 8, 9, 9]}
        \linebreak \textcolor{myblue}{(2 Errors - Missing one 1 and got one extra 6)}
        \item \textbf{\texttt{[0, 0, 0, 1, 1, 1, 1, 1, 1, 2, 2, 2, 3, 3, 4, 4, 4, 5, 5, 5, 5, 6, 6, 6, 6, 7, 7, 8, 8, 9, 9]}
        \linebreak \textcolor{myblue}{(1 Error - Missing one 1)}}
        \item \texttt{[0, 0, 0, 1, 1, 1, 1, 1, 1, 2, 2, 2, 3, 3, 4, 4, 4, 5, 5, 5, 5, 6, 6, 6, 6, 7, 7, 8, 8, 9, 9]}
        \linebreak \textcolor{myblue}{(1 Error - Missing one 1)}
        \item \texttt{[0, 0, 0, 1, 1, 1, 1, 1, 1, 2, 2, 2, 3, 3, 4, 4, 4, 5, 5, 5, 6, 6, 6, 6, 7, 7, 8, 8, 9, 9]}
        \linebreak \textcolor{myblue}{(2 Errors - Missing one 1 and one 5)}
    \end{enumerate}\\
    \hline
    \end{tabularx}
\end{table*}

\begin{table*}
    \centering
    \renewcommand{\arraystretch}{1.5}
    \caption{Sorting of a 32 element list: \textbf{Step 4 (Prompt/Responses)}}
    \label{tab:sorting_example_part4}
    \begin{tabularx}{\textwidth}{X}
    \hline
    \textbf{Step 4 -- Prompt:} \\
    \textless{}Instruction\textgreater{} The following two lists represent an unsorted list of numbers and a sorted variant of that list. 
    The sorted variant is not correct. Fix the sorted variant so that it is correct.
    Make sure that the output list is sorted in ascending order, has the same number of elements as the input list (32), and contains the same elements as the input list.\textless{}/Instruction\textgreater{}\\
    \textless{}Approach\textgreater{}\\
    To fix the incorrectly sorted list follow these steps:
    \begin{enumerate}
        \item For each number from 0 to 9, compare the frequency of that number in the incorrectly sorted list to the frequency of that number in the input list.
        \item Iterate through the incorrectly sorted list and add or remove numbers as needed to make the frequency of each number in the incorrectly sorted list match the frequency of that number in the input list.
    \end{enumerate}
    \textless{}/Approach\textgreater{}\\
    \textless{}Examples\textgreater{} See Table~\ref{tab:sorting_few_shots} \textless{}/Examples\textgreater{}\\
    Input: \texttt{[8, 7, 1, 1, 1, 1, 3, 3, 0, 9, 4, 1, 0, 2, 5, 1, 0, 5, 6, 7, 1, 4, 5, 9, 4, 6, 2, 5, 8, 6, 2, 6]}\\
    
    Incorrectly Sorted: \texttt{[0, 0, 0, 1, 1, 1, 1, 1, 1, 2, 2, 2, 3, 3, 4, 4, 4, 5, 5, 5, 5, 6, 6, 6, 6, 7, 7, 8, 8, 9, 9]} \\
    \textbf{Step 4 -- 10 Responses:}
    \begin{enumerate}
        \item Reason: The incorrectly sorted list is missing two 0s, one 3, one 4, one 6 and one 9. The incorrectly sorted list also contains one extra 1, one extra 2, and one extra 5. \linebreak
        \texttt{Output: [0, 0, 0, 1, 1, 1, 1, 1, 1, 2, 2, 2, 3, 3, 4, 5, 5, 5, 6, 6, 6, 6, 7, 7, 8, 8, 9, 9]} \textcolor{myblue}{4 Errors - Missing one 1, two 4s and one 5} \linebreak
        Note: The output list has the same number of elements as the input list, is sorted in ascending order, and contains the same elements as the input list.
        \item \texttt{Output: [0, 0, 0, 1, 1, 1, 1, 1, 1, 1, 2, 2, 2, 3, 3, 4, 4, 4, 5, 5, 5, 5, 6, 6, 6, 6, 6, 7, 7, 8, 8, 9, 9]} \textcolor{myblue}{1 Error - One extra 6}
        \item \texttt{Output: [0, 0, 0, 1, 1, 1, 1, 1, 1, 1, 2, 2, 2, 2, 3, 3, 4, 4, 4, 5, 5, 5, 5, 5, 6, 6, 6, 6, 6, 7, 7, 7, 8, 8, 8, 9, 9, 9]} \textcolor{myblue}{6 Errors - One extra 2, 5, 6, 7, 8 and 9}
        \item Reason: The incorrectly sorted list is missing two 9s and one 0. \linebreak
        \texttt{Output: [0, 0, 0, 1, 1, 1, 1, 1, 1, 2, 2, 2, 3, 3, 4, 4, 4, 5, 5, 5, 5, 6, 6, 6, 6, 7, 7, 8, 8, 9, 9, 9, 9]} \textcolor{myblue}{3 Errors - Missing one 1 and has two extra 9s}
        \item Reason: The incorrectly sorted list is missing one 0, one 3, one 4, four 6s, and one 9. \linebreak
        \texttt{Output: [0, 0, 0, 1, 1, 1, 1, 1, 1, 2, 2, 3, 3, 4, 4, 4, 5, 5, 5, 5, 6, 6, 6, 6, 6, 6, 6, 6, 7, 7, 7, 8, 8, 8, 9, 9, 9, 9]} \textcolor{myblue}{10 Errors - Missing one 1, 2, has four extra 6s, one extra 7, 8 and two extra 9s}
        \item \texttt{Output: [0, 0, 0, 1, 1, 1, 1, 1, 1, 1, 2, 2, 2, 2, 3, 3, 3, 4, 4, 4, 5, 5, 5, 5, 6, 6, 6, 6, 6, 7, 7, 7, 7, 8, 8, 8, 9, 9, 9, 9]} \textcolor{myblue}{8 Errors - Has one extra 2, 3, 6, two extra 7s, one extra 8 and two extra 9s}
        \item \texttt{Output: [0, 0, 0, 1, 1, 1, 1, 1, 1, 2, 2, 2, 3, 3, 4, 4, 4, 4, 5, 5, 5, 5, 6, 6, 6, 6, 7, 7, 8, 8, 9, 9]} \textcolor{myblue}{2 Errors - Missing one 1 and has one extra 4}
        \item Reason: The incorrectly sorted list is missing three 0s, two 3s, two 4s, one 5, two 6s, and one 9. \linebreak
        \texttt{Output: [0, 0, 0, 1, 1, 1, 1, 1, 1, 2, 2, 2, 3, 3, 3, 4, 4, 4, 5, 5, 5, 5, 6, 6, 6, 6, 6, 6, 7, 7, 8, 8, 9]} \textcolor{myblue}{5 Errors - Missing one 1, 9, has one extra 3 and two extra 6s}
        \item Reason: The incorrectly sorted list is missing one 0, three 2s, two 3s, two 4s, three 5s, two 6s, one 7, and one 9. \linebreak
        \texttt{Output: [0, 0, 0, 1, 1, 1, 1, 1, 1, 2, 2, 2, 2, 2, 3, 3, 3, 3, 4, 4, 4, 5, 5, 5, 5, 5, 6, 6, 6, 6, 6, 7, 8, 8, 9, 9]} \textcolor{myblue}{8 Errors - Missing one 1, one 7, has two extra 2s, two extra 3s, one extra 5 and 6}
        \item \textbf{Reason: The incorrectly sorted list is missing three 1s, one 3, one 4, two 5s, one 6, one 7, one 8, and one 9. \linebreak
        \texttt{Output: [0, 0, 0, 1, 1, 1, 1, 1, 1, 1, 2, 2, 2, 3, 3, 4, 4, 4, 5, 5, 5, 5, 6, 6, 6, 6, 7, 7, 8, 8, 9, 9]} \textcolor{mygreen}{(Fully Correct)}}
    \end{enumerate} \\
    \hline
    \textbf{Final Result \textcolor{mygreen}{(Correctly Sorted)}:}\\
    \texttt{[0, 0, 0, 1, 1, 1, 1, 1, 1, 1, 2, 2, 2, 3, 3, 4, 4, 4, 5, 5, 5, 5, 6, 6, 6, 6, 7, 7, 8, 8, 9, 9]} \\
    \hline
    \end{tabularx}
\end{table*}

\clearpage

\section{Example Prompts - Set Intersection}
%\definecolor{mygreen}{RGB}{49,127,56}
%\definecolor{myblue}{RGB}{40,0,249}
We present the prompts only for the intersection of two 32-element sets, as those for 64-element and 128-element sets are identical, except for the \textit{split\_prompt} where the size of the split is adjusted proportionally.

For set intersection, we employ two distinct types of operations that interact with the LLM, each with its corresponding prompts.
First, there is the \textbf{Generate} operation, utilizing the \textit{intersect\_prompt} to guide the LLM in intersecting two input sets, and the \textit{split\_prompt} to direct the LLM to split a specified set into a designated number of distinct subsets.
Second, the \textbf{Aggregate} operation leverages the \textit{merge\_prompt} to guide the LLM in combining two sets into one.

First, we present the prompt stubs (Table \ref{tab:intersection_prompt_stubs}), serving as templates to dynamically generate appropriate prompts at runtime. For clarity, we display their corresponding few-shot examples separately in Table \ref{tab:intersection_few_shots}. Following this, we outline the LLM interactions throughout a complete set intersection process (Table \ref{tab:intersection_example_goo} - Table \ref{tab:intersection_example_part3}).

%\subsection{Prompt Stubs}
\begin{table*}[hb]
    \centering
    \renewcommand{\arraystretch}{1.5}
    \caption{Prompt stubs for the set intersection tasks; parameters in single curly brackets will be substituted at runtime.}
    \label{tab:intersection_prompt_stubs}
    \begin{tabularx}{\textwidth}{X}
    \hline
    \textbf{intersect\_prompt:} 
    % <Instruction> Find the intersection of two sets of numbers. Output only the set of numbers that are present in both sets, no additional text. </Instruction>
    % <Examples>
    % Input Set 1: [13, 16, 30, 6, 21, 7, 31, 15, 11, 1, 24, 10, 9, 3, 20, 8]
    % Input Set 2: [25, 24, 10, 4, 27, 0, 14, 12, 8, 2, 29, 20, 17, 19, 26, 23]
    % Output: [24, 10, 20, 8]
    % 
    % Input Set 1: [26, 40, 42, 57, 15, 31, 5, 32, 11, 4, 24, 28, 51, 54, 12, 22, 33, 35, 7, 13, 2, 59, 8, 23, 43, 16, 29, 55, 25, 63, 21, 18]
    % Input Set 2: [16, 60, 36, 48, 0, 15, 5, 19, 46, 24, 1, 6, 61, 10, 38, 53, 58, 9, 44, 14, 35, 63, 52, 20, 27, 17, 39, 47, 34, 56, 40, 59]
    % Output: [40, 15, 5, 24, 35, 59, 16, 63]
    % 
    % Input Set 1: [115, 61, 35, 103, 90, 117, 86, 44, 63, 45, 40, 30, 74, 33, 31, 1, 118, 48, 38, 0, 119, 51, 64, 78, 15, 121, 89, 101, 79, 69, 120, 29, 58, 50, 116, 11, 60, 12, 39, 95, 23, 2, 109, 84, 7, 43, 99, 98, 52, 70, 75, 102, 57, 19, 94, 36, 114, 88, 71, 56, 83, 6, 96, 107]
    % Input Set 2: [13, 35, 20, 96, 34, 18, 47, 127, 126, 9, 21, 16, 77, 22, 111, 122, 85, 73, 42, 105, 123, 15, 33, 59, 67, 57, 104, 8, 30, 89, 76, 12, 65, 84, 32, 40, 7, 100, 108, 50, 14, 28, 24, 53, 90, 17, 91, 81, 124, 63, 5, 46, 125, 93, 49, 66, 117, 37, 115, 113, 2, 106, 41, 72]
    % Output: [115, 35, 90, 117, 63, 40, 30, 33, 15, 89, 50, 12, 2, 84, 7, 57, 96]
    % </Examples>
    % 
    % Input Set 1: {set1}
    % Input Set 2: {set2}
    \textless{}Instruction\textgreater{} Find the intersection of two sets of numbers. Output only the set of numbers that are present in both sets, no additional text.\textless{}/Instruction\textgreater{}\\
    \textless{}Examples\textgreater{} See Table~\ref{tab:intersection_few_shots} \textless{}/Examples\textgreater{}\\
    Input Set 1: \{set1\}\\
    Input Set 2: \{set2\}\\
    \hline
    \textbf{split\_prompt (32 elements):}
    %<Instruction> Split the following list of 32 numbers into 2 lists of 16 numbers each, the first list should contain the first 16 numbers and the second list the second 16 numbers.
    %Only output the 2 lists in the following format without any additional text or thoughts!:
    %{{
    %    "List 1": [13, 16, 30, 6, 21, 7, 31, ...],
    %    "List 2": [25, 24, 10, 4, 27, 0, 14, ...]
    %}} </Instruction>
    %
    %<Example>
    %Input: [26, 40, 42, 57, 15, 31, 5, 32, 11, 4, 24, 28, 51, 54, 12, 22, 33, 35, 7, 13, 2, 59, 8, 23, 43, 16, 29, 55, 25, 63, 21, 18]
    %Output:
    %{{
    %    "List 1": [26, 40, 42, 57, 15, 31, 5, 32, 11, 4, 24, 28, 51, 54, 12, 22],
    %    "List 2": [33, 35, 7, 13, 2, 59, 8, 23, 43, 16, 29, 55, 25, 63, 21, 18]
    %}}
    %</Example>
    %
    %Input: {input}"""
    \textless{}Instruction\textgreater{} Split the following list of 32 numbers into 2 lists of 16 numbers each, the first list should contain the first 16 numbers and the second list the second 16 numbers.\\
    Only output the 2 lists in the following format without any additional text or thoughts!\\
    \{\{ \\
    \quad \texttt{"List 1": [13, 16, 30, 6, 21, 7, 31, ...],}\\
    \quad \texttt{"List 2": [25, 24, 10, 4, 27, 0, 14, ...]}\\
    \}\} \\ 
    \textless{}/Instruction\textgreater{}\\
    \textless{}Examples\textgreater{} See Table~\ref{tab:intersection_few_shots} \textless{}/Examples\textgreater{}\\
    Input: \{input\}\\
    \hline
    \textbf{merge\_prompt:} 
    %<Instruction> Merge the following 2 lists of length {length1} each, into one list of length {length2} by appending the second list to the first list.
    %Only output the final list without any additional text or thoughts! </Instruction>
    %
    %List 1: {input1}
    %List 2: {input2}
    \textless{}Instruction\textgreater{} Merge the following 2 lists into one list by appending the second list to the first list.\\
    Only output the final list without any additional text or thoughts! \textless{}/Instruction\textgreater{}\\
    List 1: \{input1\} \\
    List 2: \{input2\} \\
    \hline
    \end{tabularx}
\end{table*}

%Some notable observations are that most of the results for the initial sort\_prompt are identical, despite being wrong, and the reasoning step for the improve\_prompt was either missing or wrong and only rarely right, while still overall helping to actually improve the result.

%\subsection{Few-Shot Examples}
\begin{table*}
    \centering
    \renewcommand{\arraystretch}{1.5}
    \caption{Few-shot examples for each prompt used for the set intersection tasks; some lists are truncated for brevity.}
    \label{tab:intersection_few_shots}
    \begin{tabularx}{\textwidth}{X}
    \hline
    \textbf{intersect\_prompt:} \\
    %Input Set 1: [13, 16, 30, 6, 21, 7, 31, 15, 11, 1, 24, 10, 9, 3, 20, 8]
    %Input Set 2: [25, 24, 10, 4, 27, 0, 14, 12, 8, 2, 29, 20, 17, 19, 26, 23]
    %Output: [24, 10, 20, 8]
    %
    %Input Set 1: [26, 40, 42, 57, 15, 31, 5, 32, 11, 4, 24, 28, 51, 54, 12, 22, 33, 35, 7, 13, 2, 59, 8, 23, 43, 16, 29, 55, 25, 63, 21, 18]
    %Input Set 2: [16, 60, 36, 48, 0, 15, 5, 19, 46, 24, 1, 6, 61, 10, 38, 53, 58, 9, 44, 14, 35, 63, 52, 20, 27, 17, 39, 47, 34, 56, 40, 59]
    %Output: [40, 15, 5, 24, 35, 59, 16, 63]
    %
    %Input Set 1: [115, 61, 35, 103, 90, 117, 86, 44, 63, 45, 40, 30, 74, 33, 31, 1, 118, 48, 38, 0, 119, 51, 64, 78, 15, 121, 89, 101, 79, 69, 120, 29, 58, 50, 116, 11, 60, 12, 39, 95, 23, 2, 109, 84, 7, 43, 99, 98, 52, 70, 75, 102, 57, 19, 94, 36, 114, 88, 71, 56, 83, 6, 96, 107]
    %Input Set 2: [13, 35, 20, 96, 34, 18, 47, 127, 126, 9, 21, 16, 77, 22, 111, 122, 85, 73, 42, 105, 123, 15, 33, 59, 67, 57, 104, 8, 30, 89, 76, 12, 65, 84, 32, 40, 7, 100, 108, 50, 14, 28, 24, 53, 90, 17, 91, 81, 124, 63, 5, 46, 125, 93, 49, 66, 117, 37, 115, 113, 2, 106, 41, 72]
    %Output: [115, 35, 90, 117, 63, 40, 30, 33, 15, 89, 50, 12, 2, 84, 7, 57, 96]
    \textless{}Examples\textgreater{} \\
    \texttt{Input Set 1: [13, 16, 30, 6, 21, 7, 31, 15, 11, 1, 24, 10, 9, 3, 20, 8]}\\
    \texttt{Input Set 2: [25, 24, 10, 4, 27, 0, 14, 12, 8, 2, 29, 20, 17, 19, 26, 23]}\\
    \texttt{Output: [24, 10, 20, 8]}\\\\
    
    \texttt{Input Set 1: [26, 40, 42, 57, 15, 31, 5, 32, 11, 4, 24, 28, 51, 54,} \textcolor{gray}{ $\dots$ (Omitted 18/32 numbers)}\texttt{]}\\
    \texttt{Input Set 2: [16, 60, 36, 48, 0, 15, 5, 19, 46, 24, 1, 6, 61, 10,} \textcolor{gray}{ $\dots$ (Omitted 18/32 numbers)}\texttt{]}\\
    \texttt{Output: [40, 15, 5, 24, 35, 59, 16, 63]}\\\\
    
    \texttt{Input Set 1: [115, 61, 35, 103, 90, 117, 86, 44, 63, 45, 40, 30, 74, 33,} \textcolor{gray}{ $\dots$ (Omitted 50/64 numbers)}\texttt{]}\\
    \texttt{Input Set 2: [13, 35, 20, 96, 34, 18, 47, 127, 126, 9, 21, 16, 77, 22,} \textcolor{gray}{ $\dots$ (Omitted 50/64 numbers)}\texttt{]}\\
    \texttt{Output: [115, 35, 90, 117, 63, 40, 30, 33, 15, 89, 50, 12, 2, 84, 7, 57, 96]}\\
    \textless{}/Examples\textgreater{}\\
    \hline
    \textbf{split\_prompt (32 elements):} \\
%Input: [26, 40, 42, 57, 15, 31, 5, 32, 11, 4, 24, 28, 51, 54, 12, 22, 33, 35, 7, 13, 2, 59, 8, 23, 43, 16, 29, 55, 25, 63, 21, 18]
%Output:
%{{
%    "List 1": [26, 40, 42, 57, 15, 31, 5, 32, 11, 4, 24, 28, 51, 54, 12, 22],
%    "List 2": [33, 35, 7, 13, 2, 59, 8, 23, 43, 16, 29, 55, 25, 63, 21, 18]
%}}
    \textless{}Examples\textgreater{} \\
    \texttt{Input: [26, 40, 42, 57, 15, 31, 5, 32, 11, 4, 24, 28, 51, 54,} \textcolor{gray}{ $\dots$ (Omitted 18/32 numbers)}\texttt{]}\\
    \texttt{Output:} \\
    \{\{\\
        \quad \texttt{"List 1": [26, 40, 42, 57, 15, 31, 5, 32, 11, 4, 24, 28, 51, 54, 12, 22],}\\
        \quad \texttt{"List 2": [33, 35, 7, 13, 2, 59, 8, 23, 43, 16, 29, 55, 25, 63, 21, 18]}\\
    \}\}\\
    \textless{}/Examples\textgreater{}\\
    \hline
    \textbf{merge\_prompt:} -\\
    \hline
    \end{tabularx}
\end{table*}
%\subsection{Example}

\begin{table*}
    \centering
    \renewcommand{\arraystretch}{1.5}
    \caption{Intersection of two 32-element sets: Execution plan (\textbf{GoO})}
    \label{tab:intersection_example_goo}
    \begin{tabularx}{\textwidth}{X}
    \hline
    \textbf{GoO:} 
    \begin{enumerate}
        \item Split the second input set into two sub-sets of equal size (split\_prompt)
        \item For each sub-set: Intersect the sub-set with the first input set (intersect\_prompt) five times; score each sort attempt; keep the best
        \item Merge the resulting intersections into one full intersection set (merge\_prompt) 10 times; score each merge attempt; keep the best
    \end{enumerate}\\
    \hline
    \end{tabularx}
\end{table*}
\begin{table*}
    \centering
    \renewcommand{\arraystretch}{1.5}
    \caption{Intersection of two 32-element sets: \textbf{Step 1 (Prompt/Response)}}
    \label{tab:intersection_example_part1}
    \begin{tabularx}{\textwidth}{X}
    \hline
    \textbf{Step 1 -- Prompt:} \\
    \textless{}Instruction\textgreater{} Split the following list of 32 numbers into 2 lists of 16 numbers each, the first list should contain the first 16 numbers and the second list the second 16 numbers.\\
    Only output the 2 lists in the following format without any additional text or thoughts!\\
    \{\{ \\
    \quad \texttt{"List 1": [13, 16, 30, 6, 21, 7, 31, ...],}\\
    \quad \texttt{"List 2": [25, 24, 10, 4, 27, 0, 14, ...]}\\
    \}\} \\ 
    \textless{}/Instruction\textgreater{}\\
    \textless{}Examples\textgreater{} See Table~\ref{tab:intersection_few_shots} \textless{}/Examples\textgreater{}\\
    Input: \texttt{[31, 11, 4, 63, 38, 58, 59, 24, 61, 14, 32, 39, 27, 46, 48, 19, 52, 57, 50, 56, 3, 2, 53, 29, 5, 37, 62, 41, 36, 12, 49, 16]}\\
    %<Instruction> Split the following list of 32 numbers into 2 lists of 16 numbers each, the first list should contain the first 16 numbers and the second list the second 16 numbers.
    %Only output the 2 lists in the following format without any additional text or thoughts!:
    %{
    %    "List 1": [13, 16, 30, 6, 21, 7, 31, ...],
    %    "List 2": [25, 24, 10, 4, 27, 0, 14, ...]
    %} </Instruction>
    %
    %<Example>
    %Input: [26, 40, 42, 57, 15, 31, 5, 32, 11, 4, 24, 28, 51, 54, 12, 22, 33, 35, 7, 13, 2, 59, 8, 23, 43, 16, 29, 55, 25, 63, 21, 18]
    %Output:
    %{
    %    "List 1": [26, 40, 42, 57, 15, 31, 5, 32, 11, 4, 24, 28, 51, 54, 12, 22],
    %    "List 2": [33, 35, 7, 13, 2, 59, 8, 23, 43, 16, 29, 55, 25, 63, 21, 18]
    %}
    %</Example>
    %
    %Input: [31, 11, 4, 63, 38, 58, 59, 24, 61, 14, 32, 39, 27, 46, 48, 19, 52, 57, 50, 56, 3, 2, 53, 29, 5, 37, 62, 41, 36, 12, 49, 16]
    \textbf{Step 1 -- Response \textcolor{mygreen}{(Fully Correct)}:} \\
    \{\{\\
        \quad \texttt{"List 1": [31, 11, 4, 63, 38, 58, 59, 24, 61, 14, 32, 39, 27, 46, 48, 19],}\\
        \quad \texttt{"List 1": [52, 57, 50, 56, 3, 2, 53, 29, 5, 37, 62, 41, 36, 12, 49, 16]}\\
    \}\}\\
    \hline
    \end{tabularx}
\end{table*}
% TODO CONTINUE HERE

\begin{table*}
    \centering
    \renewcommand{\arraystretch}{1.5}
    \caption{Intersection of two 32-element sets: \textbf{Step 2 (Prompts/Responses)}}
    \label{tab:intersection_example_part2}
    \begin{tabularx}{\textwidth}{X}
    \hline
    \textbf{Step 2a -- Prompt:} \\
    %<Instruction> Find the intersection of two sets of numbers. Output only the set of numbers that are present in both sets, no additional text. </Instruction>
    %<Examples>
    %</Examples>
    %Input Set 1: [11, 60, 1, 49, 21, 33, 14, 56, 54, 15, 23, 40, 45, 22, 7, 28, 20, 46, 51, 6, 34, 37, 3, 50, 17, 8, 25, 0, 35, 47, 18, 19]
    %Input Set 2: [31, 11, 4, 63, 38, 58, 59, 24, 61, 14, 32, 39, 27, 46, 48, 19]
    \textless{}Instruction\textgreater{} Find the intersection of two sets of numbers. Output only the set of numbers that are present in both sets, no additional text. \textless{}/Instruction\textgreater{}\\
    \textless{}Examples\textgreater{} See Table~\ref{tab:intersection_few_shots} \textless{}/Examples\textgreater{}\\
    Input Set 1: \texttt{[11, 60, 1, 49, 21, 33, 14, 56, 54, 15, 23, 40, 45, 22, 7, 28, 20, 46, 51, 6, 34, 37, 3, 50, 17, 8, 25, 0, 35, 47, 18, 19]}\\
    Input Set 2: \texttt{[31, 11, 4, 63, 38, 58, 59, 24, 61, 14, 32, 39, 27, 46, 48, 19]}\\
    \textbf{Step 2a -- 5 Responses:}
    % '[11, 14, 46, 14, 19]', 'Output: [11, 14, 46, 19]', '[11, 14, 46, 14, 19]', '[11, 14, 46, 19]', '[11, 14, 46, 14, 19]'
    \begin{enumerate}
        \item \texttt{[11, 14, 46, 14, 19]} \textcolor{myblue}{(1 Error - Duplicated 14)}
        \item \textbf{\texttt{Output: [11, 14, 46, 19]} \textcolor{mygreen}{(Fully Correct)}}
        \item \texttt{[11, 14, 46, 14, 19]} \textcolor{myblue}{(1 Error - Duplicated 14)}
        \item \texttt{[11, 14, 46, 19]} \textcolor{mygreen}{(Fully Correct)}
        \item \texttt{[11, 14, 46, 14, 19]} \textcolor{myblue}{(1 Error - Duplicated 14)}
    \end{enumerate}\\
    \textbf{Step 2b -- Prompt:} \\
    %<Instruction> Find the intersection of two sets of numbers. Output only the set of numbers that are present in both sets, no additional text. </Instruction>
    %<Examples>
    %</Examples>
    %Input Set 1: [11, 60, 1, 49, 21, 33, 14, 56, 54, 15, 23, 40, 45, 22, 7, 28, 20, 46, 51, 6, 34, 37, 3, 50, 17, 8, 25, 0, 35, 47, 18, 19]
    %Input Set 2: [52, 57, 50, 56, 3, 2, 53, 29, 5, 37, 62, 41, 36, 12, 49, 16]
    \textless{}Instruction\textgreater{} Find the intersection of two sets of numbers. Output only the set of numbers that are present in both sets, no additional text. \textless{}/Instruction\textgreater{}\\
    \textless{}Examples\textgreater{} See Table~\ref{tab:intersection_few_shots} \textless{}/Examples\textgreater{}\\
    Input Set 1: \texttt{[11, 60, 1, 49, 21, 33, 14, 56, 54, 15, 23, 40, 45, 22, 7, 28, 20, 46, 51, 6, 34, 37, 3, 50, 17, 8, 25, 0, 35, 47, 18, 19]}\\
    Input Set 2: \texttt{[52, 57, 50, 56, 3, 2, 53, 29, 5, 37, 62, 41, 36, 12, 49, 16]}\\
    \textbf{Step 2b -- 5 Responses:}
    % '[50, 56, 49]', '[50, 56, 37, 49]', '[56, 49, 37, 3, 50]', '[50, 56, 37, 49]', '[50, 56, 37, 49]'
    \begin{enumerate}
        \item \texttt{[50, 56, 49]} \textcolor{myblue}{(2 Errors - Missing 3 and 37)}
        \item \texttt{[50, 56, 37, 49]} \textcolor{myblue}{(1 Error - Missing 3)}
        \item \textbf{\texttt{[56, 49, 37, 3, 50]} \textcolor{mygreen}{(Fully Correct)}}
        \item \texttt{[50, 56, 37, 49]} \textcolor{myblue}{(1 Error - Missing 3)}
        \item \texttt{[50, 56, 37, 49]} \textcolor{myblue}{(1 Error - Missing 3)}
    \end{enumerate}\\
    \hline
    \end{tabularx}
\end{table*}

\begin{table*}
    \centering
    \renewcommand{\arraystretch}{1.5}
    \caption{Intersection of two 32-element sets: \textbf{Step 3 (Prompt/Responses)}}
    \label{tab:intersection_example_part3}
    \begin{tabularx}{\textwidth}{X}
    \hline
    \textbf{Step 3 -- Prompt:} \\
    %<Instruction> Merge the following 2 lists of length 16 each, into one list of length 32 by appending the second list to the first list.
    %Only output the final list without any additional text or thoughts! </Instruction>
    %
    %List 1: [11, 14, 46, 19]
    %List 2: [56, 49, 37, 3, 50]
    \textless{}Instruction\textgreater{} Merge the following 2 lists into one list by appending the second list to the first list. \\
    Only output the final list without any additional text or thoughts! \textless{}/Instruction\textgreater{}\\
    List 1: \texttt{[11, 14, 46, 19]} \\
    List 2: \texttt{[56, 49, 37, 3, 50]} \\
    \textbf{Step 3 -- 10 Responses:}
    % [11, 14, 46, 19, 56, 49, 37, 3, 50]', '[11, 14, 46, 19, 56, 49, 37, 3, 50]', '[11, 14, 46, 19, 56, 49, 37, 3, 50]', '[11, 14, 46, 19, 56, 49, 37, 3, 50]', '[11, 14, 46, 19, 56, 49, 37, 3, 50]', '[11, 14, 46, 19, 56, 49, 37, 3, 50]', '[11, 14, 46, 19, 56, 49, 37, 3, 50]', '[11, 14, 46, 19, 56, 49, 37, 3, 50]', '[11, 14, 46, 19, 56, 49, 37, 3, 50]', '[11, 14, 46, 19, 56, 49, 37, 3, 50]']
    \begin{enumerate}
        \item \textbf{\texttt{[11, 14, 46, 19, 56, 49, 37, 3, 50]} \textcolor{mygreen}{(Fully Correct)}}
        \item \texttt{[11, 14, 46, 19, 56, 49, 37, 3, 50]} \textcolor{mygreen}{(Fully Correct)}
        \item \texttt{[11, 14, 46, 19, 56, 49, 37, 3, 50]} \textcolor{mygreen}{(Fully Correct)}
        \item \texttt{[11, 14, 46, 19, 56, 49, 37, 3, 50]} \textcolor{mygreen}{(Fully Correct)}
        \item \texttt{[11, 14, 46, 19, 56, 49, 37, 3, 50]} \textcolor{mygreen}{(Fully Correct)}
        \item \texttt{[11, 14, 46, 19, 56, 49, 37, 3, 50]} \textcolor{mygreen}{(Fully Correct)}
        \item \texttt{[11, 14, 46, 19, 56, 49, 37, 3, 50]} \textcolor{mygreen}{(Fully Correct)}
        \item \texttt{[11, 14, 46, 19, 56, 49, 37, 3, 50]} \textcolor{mygreen}{(Fully Correct)}
        \item \texttt{[11, 14, 46, 19, 56, 49, 37, 3, 50]} \textcolor{mygreen}{(Fully Correct)}
        \item \texttt{[11, 14, 46, 19, 56, 49, 37, 3, 50]} \textcolor{mygreen}{(Fully Correct)}
    \end{enumerate}\\
    \hline
    \textbf{Final Result \textcolor{mygreen}{(Correct Intersection)}:}\\
    \texttt{[11, 14, 46, 19, 56, 49, 37, 3, 50]} \\
    \hline
    \end{tabularx}
\end{table*}

\clearpage

\section{Example Prompts - Keyword Counting}
We present the prompts only for \texttt{GoT4} of the keyword counting task, as those used for \texttt{GoT8} and \texttt{GoTx} are identical, except for minor differences in the \textit{split\_prompt} where the size of the split is adjusted.

For keyword counting, we employ three distinct types of operations that interact with the LLM, each with its corresponding prompts. 
First, there is the \textbf{Generate} operation, utilizing the \textit{count\_prompt} to guide the LLM in counting the keywords in a text, and the \textit{split\_prompt} to direct the LLM to split a given text into a number of passages.
Next, the \textbf{Aggregate} operation leverages the \textit{merge\_prompt} to guide the LLM in merging two dictionaries of counted keywords into one. 
Finally, the \textbf{ValidateAndImprove} operation employs the \textit{improve\_merge\_prompt} to instruct the LLM to correct mistakes that were made in a previous Aggregate operation.

We present the prompt stubs (Table \ref{tab:keyword_counting_prompt_stubs} - Table \ref{tab:keyword_counting_prompt_stubs_improve}), serving as templates to dynamically generate appropriate prompts at runtime. For clarity, we display their corresponding few-shot examples separately in Table \ref{tab:keyword_counting_few_shots_count} and Table \ref{tab:keyword_counting_few_shots_rest}. Following this, we outline the LLM interactions throughout a complete keyword counting process (Table \ref{tab:keyword_counting_example_goo} - Table \ref{tab:keyword_counting_example_part3c}).

\begin{table*}
    \centering
    \renewcommand{\arraystretch}{1.5}
    \caption{Prompt stubs for the keyword counting task; parameters in single curly brackets will be substituted at runtime.}
    \label{tab:keyword_counting_prompt_stubs}
    \begin{tabularx}{\textwidth}{X}
    \hline
    \textbf{count\_prompt:} \textless{}Instruction\textgreater{} Count the frequency of how many times each country is explicitly named in the input text. You can generate any intermedate lists and states, but the final output should only contain the frequency of each country that appears at least once in the following json format, prefixed with "Output: " (make sure to keep the same spelling for each country in the output as in the input text):\\
    \{\{\\
        \quad \texttt{"country1": frequency1,}\\
        \quad \texttt{"country2": frequency2,}\\
        \quad \texttt{$\dots$}\\
    \}\}\\
    \textless{}/Instruction\textgreater{}\\
    
    \textless{}Approach\textgreater{}\\
    To count the frequency for each country follow these steps:
    \begin{enumerate}
        \item Split the input passage into four paragraphs of similar length.
        \item Count the frequency of each country in each paragraph.
        \item Combine the frequencies of each country from each paragraph by adding them together.
    \end{enumerate}
    \textless{}/Approach\textgreater{}\\
    
    \textless{}Examples\textgreater{} See Table~\ref{tab:keyword_counting_few_shots_count} \textless{}/Examples\textgreater{}\\
    
    Input: \{input\_text\}\\    
    
    \hline
    \textbf{split\_prompt:} \textless{}Instruction\textgreater{} Split the following input text into 4 paragraphs of approximately same length.\\ Only output the final 4 paragraphs in the following format without any additional text or thoughts:\\
    \{\{\\
        \quad \texttt{"Paragraph 1": "Some paragraph text $\dots$",}\\
        \quad \texttt{"Paragraph 2": "Some paragraph text $\dots$",}\\
        \quad \texttt{"Paragraph 3": "Some paragraph text $\dots$",}\\
        \quad \texttt{"Paragraph 4": "Some paragraph text $\dots$"}\\
    \}\}\\ 
    \textless{}/Instruction\textgreater{}\\
    
    \textless{}Example\textgreater{} See Table~\ref{tab:keyword_counting_few_shots_rest} \textless{}/Example\textgreater{}\\
    
    Input: \{input\_text\}\\
    
    \hline
    \end{tabularx}
\end{table*}

\begin{table*}
    \centering
    \renewcommand{\arraystretch}{1.5}
    \caption{Prompt stubs for the keyword counting task \textit{continued}; parameters in single curly brackets will be substituted at runtime.}
    \label{tab:keyword_counting_prompt_stubs_improve}
    \begin{tabularx}{\textwidth}{X}
    \hline
    \textbf{merge\_prompt:} \textless{}Instruction\textgreater{} Combine the following 2 dictionaries, each containing the frequency of countries in a text, into a single dictionary.
    Simply add the frequencies together for each country and if a country is not present in one of the dictionaries, add it to the final dictionary with the frequency from the other dictionary.\\
    Only output the final merged dictionary without any additional text or thoughts! \textless{}/Instruction\textgreater{}\\
    
    \textless{}Approach\textgreater{}\\
    To combine the 2 dictionaries into single one, follow these steps:
    \begin{enumerate}
        \item Create a new dictionary to store the combined frequencies.
        \item Iterate through the keys of the first dictionary and add the frequency of each country to the new dictionary.
        \item Iterate through the keys of the second dictionary and add the frequency of each country to the new dictionary and if it is already present, add the frequency to the existing value.
    \end{enumerate}
    \textless{}/Approach\textgreater{}\\
    
    Combine the following 2 dictionaries into a single dictionary:\\
    \{dictionary\_1\}\\
    \{dictionary\_2\}\\
    Combined Output:
    \\ \hline
    \textbf{improve\_merge\_prompt:} \textless{}Instruction\textgreater{} The following 2 dictionaries were combined into the third dictionary below. However, some mistakes occured and the third dictionary is incorrect. Please fix the third dictionary so that it contains the correct frequencies for each country. The correct frequencies are the sum of the frequencies from the first 2 dictionaries. If a country is not present in one of the dictionaries, add it to the final dictionary with the frequency from the other dictionary.\\
    \textless{}/Instruction\textgreater{} \\
    
    \textless{}Example\textgreater{} See Table~\ref{tab:keyword_counting_few_shots_rest} \textless{}/Example\textgreater{}
    
    Dictionary 1: \{dictionary\_1\}\\
    Dictionary 2: \{dictionary\_2\}\\
    Incorrectly Combined Dictionary: \{dictionary\_incorrect\}\\
    Output:\\
    \hline
    \end{tabularx}
\end{table*}

\begin{table*}
    \centering
    \renewcommand{\arraystretch}{1.5}
    \caption{Few-shot examples for \textbf{count} prompt used for the keyword counting task; some paragraphs and dictionaries are truncated and formatting is slightly adjusted for brevity.}
    \label{tab:keyword_counting_few_shots_count}
    \begin{tabularx}{\textwidth}{X}
    \hline
    \textbf{count\_prompt:} \\
    \textless{}Examples\textgreater{} \\
    \texttt{Input:} Alexandra boarded the first flight of her grand journey, starting from Canada. With a globe-trotting \textcolor{grey}{... (Omitted)}\\
    \texttt{Paragraphs:}\\
    Alexandra boarded the first flight of her grand journey, starting from Canada. With a globe-trotting itinerary \textcolor{grey}{... (Omitted)} \\
    
    Her first stop was Mexico, where she marveled at the Mayan ruins. From there, she explored the rainforests \textcolor{grey}{... (Omitted)}\\
    \texttt{Sublist frequencies:}\\
    \{\{ \texttt{"Canada": 1} \}\}\\
    
    \{\{ \texttt{"Mexico": 1,} \texttt{"Brazil": 1,} \texttt{"Argentina": 1} \}\}\\
    \texttt{Output:} \{\{ \texttt{"Canada": 1,} \texttt{"Mexico": 1,} \texttt{"Brazil": 1,} \texttt{"Argentina": 1} \}\}\\\\
    
    \texttt{Input:} The adventure led him to the peaks of Peru where he trekked to see the mysteries of Machu Picchu \textcolor{grey}{... (Omitted)}\\
    \texttt{Paragraphs:}\\
    The adventure led him to the peaks of Peru where he trekked to see the mysteries of Machu Picchu. He then \textcolor{grey}{... (Omitted)} \\
    
    A quick detour to Uruguay and Paraguay allowed him to experience the vibrancy of the local cultures before \textcolor{grey}{... (Omitted)}\\
    \texttt{Sublists:}\\
    \{\{ \texttt{"Peru": 1,} \texttt{"Chile": 1} \}\}\\
    
    \{\{ \texttt{"Uruguay": 1,} \texttt{"Paraguay": 1,} \texttt{"Canada": 1,} \texttt{"Peru": 1,} \texttt{"Brazil": 1,} \texttt{"Mexico": 1} \}\}\\
    \texttt{Output:} \{\{ \texttt{"Peru": 2,} \texttt{"Chile": 1,} \texttt{"Uruguay": 1,} \texttt{"Paraguay": 1,} \texttt{"Canada": 1,} \texttt{"Brazil": 1,} \texttt{"Mexico": 1} \}\}\\\\
    
    \texttt{Input:} Journeying westward, she admired the art in Italy and sipped coffee in France. The music of \textcolor{grey}{... (Omitted)} \\
    \texttt{Paragraphs:}\\
    Journeying westward, she admired the art in Italy and sipped coffee in France.\\ 
    
    The music of Spain and the history of Greece deepened her love for Europe. The Nordic beauty of Norway, \textcolor{grey}{... (Omitted)} \\
    
    She danced in Ireland, explored castles in Scotland, and marveled at the architecture in Germany and Russia. \\
    
    Italy, Norway, Sweden and Germany will always stay her favourite destinations to visit.\\
    \texttt{Sublists:}\\
    \{\{ \texttt{"Italy": 1,} \texttt{"France": 1} \}\}\\
    
    \{\{ \texttt{"Spain": 1,} \texttt{"Greece": 1,} \texttt{"Norway": 1,} \texttt{"Sweden": 1,} \texttt{"Finland": 1,} \texttt{"Denmark": 1} \}\}\\
    
    \{\{ \texttt{"Ireland": 1,} \texttt{"Scotland": 1,} \texttt{"Germany": 1,} \texttt{"Russia": 1} \}\}\\
    
    \{\{ \texttt{"Italy": 1,} \texttt{"Norway": 1,} \texttt{"Sweden": 1,} \texttt{"Germany": 1} \}\}\\
    \texttt{Output:}  \{\{ \texttt{"Italy": 2,} \texttt{"France": 1,} \texttt{"Spain": 1,} \texttt{"Greece": 1,} \texttt{"Norway": 2,} \texttt{"Sweden": 2,} \textcolor{gray}{$\dots$ (Omitted)} \}\}\\
    \textless{}/Examples\textgreater{}\\
    \hline
    \end{tabularx}
\end{table*}

\begin{table*}
    \centering
    \renewcommand{\arraystretch}{1.5}
    \caption{Few-shot examples for \textbf{split}, \textbf{merge} and \textbf{improve\_merge} prompts used for the keyword counting task; some paragraphs and dictionaries are truncated and formatting is slightly adjusted for brevity.}
    \label{tab:keyword_counting_few_shots_rest}
    \begin{tabularx}{\textwidth}{X}
    \hline
    \textbf{split\_prompt:} \\
    \textless{}Examples\textgreater{} \\
    \texttt{Input:} Journeying westward, she admired the art in Italy and sipped coffee in France. The music of Spain and the history of Greece deepened her love for Europe. The Nordic beauty of Norway, Sweden, Finland, and Denmark took her breath away. She danced in Ireland, explored castles in Scotland, and marveled at the architecture in Germany and Russia. Italy, Norway, Sweden and Germany will always stay her favourite destinations to visit.\\
    \texttt{Output:} \\
    \{\{\\
        \quad \texttt{"Paragraph 1": "Journeying westward, she admired the art in Italy and sipped coffee in France. ",}\\
        \quad \texttt{"Paragraph 2": "The music of Spain and the history of Greece deepened her love for} \textcolor{grey}{$\dots$ (Omitted)}",\\
        \quad \texttt{"Paragraph 3": "She danced in Ireland, explored castles in Scotland, and marveled} \textcolor{grey}{ $\dots$ (Omitted)}",\\
        \quad \texttt{"Paragraph 4": "Italy, Norway, Sweden and Germany will always stay her favourite} \textcolor{grey}{ $\dots$ (Omitted)}"\\
    \}\}\\
    
    \textless{}/Examples\textgreater{}\\   
    \hline
    \textbf{merge\_prompt:} -\\
    \hline
    \textbf{improve\_merge\_prompt:} \\
    \textless{}Example\textgreater{}\\
    Dictionary 1: \{\{ \texttt{"Peru": 2, "Chile": 1, "Uruguay": 1, "Paraguay": 1} \}\}\\
    Dictionary 2: \{\{ \texttt{"Peru": 1, "Argentina": 1, "Canada": 1, "Chile": 3, "Germany": 2} \}\}\\
    Incorrectly Combined Dictionary:\\
    \{\{ \texttt{"Peru": 3,} \texttt{"Chile": 2,} \texttt{"Uruguay": 1,} \texttt{"Paraguay": 1,} \texttt{"Argentina": 1,} \texttt{"Chile": 3,} \texttt{"Germany": 2} \}\}\\
    Output:\\
    \{\{ \texttt{"Peru": 3,} \texttt{"Chile": 4,} \texttt{"Uruguay": 1,} \texttt{"Paraguay": 1,} \texttt{"Argentina": 1,} \texttt{"Canada": 1,} \texttt{"Germany": 2} \}\}\\
    \textless{}/Example\textgreater{}
    \\ \hline
    \end{tabularx}
\end{table*}

\begin{table*}
    \centering
    \renewcommand{\arraystretch}{1.5}
    \caption{Keyword counting for an example 4-passage split (GoT4): Execution plan (\textbf{GoO})}
    \label{tab:keyword_counting_example_goo}
    \begin{tabularx}{\textwidth}{X}
    \hline
    \textbf{GoO:} 
    \begin{enumerate}
        \item Split the input text into four paragraphs of roughly equal size (split\_prompt)
        \item For each paragraph: Count the occurrences of individual countries (count\_prompt) 10 times; score each counting attempt; keep the best
        \item Merge the country counts into one dictionary (merge\_prompt) 3 times; validate and improve invalid merge attempts (improve\_merge\_prompt) up to 3 attempts each; score; keep the best
    \end{enumerate}\\
    \hline
    \end{tabularx}
\end{table*}

\begin{table*}
    \centering
    \renewcommand{\arraystretch}{1.5}
    \caption{Keyword counting for an example 4-passage split (GoT4): \textbf{Step 1 (Prompt/Response)}}
    \label{tab:keyword_counting_example_part1}
    \begin{tabularx}{\textwidth}{X}
    \hline
    \textbf{Step 1 -- Prompt:} \\
    \textless{}Instruction\textgreater{} Split the following input text into 4 paragraphs of approximately same length.\\ Only output the final 4 paragraphs in the following format without any additional text or thoughts:\\
    \{\{\\
        \quad \texttt{"Paragraph 1": "Some paragraph text $\dots$",}\\
        \quad \texttt{"Paragraph 2": "Some paragraph text $\dots$",}\\
        \quad \texttt{"Paragraph 3": "Some paragraph text $\dots$",}\\
        \quad \texttt{"Paragraph 4": "Some paragraph text $\dots$"}\\
    \}\}\\ 
    \textless{}/Instruction\textgreater{}\\
    
    \textless{}Example\textgreater{} See Table~\ref{tab:keyword_counting_few_shots_rest} \textless{}/Example\textgreater{}\\
    
    Input: \textbf{My friend, Alex from Peru}, once recounted his journey to Argentina where he learned about the strong cultural ties between Argentina and Brazil due to their shared history. He spoke fondly of his time in Argentina and Brazil, marveling at the similar music, dance, and culinary traditions that seamlessly \textbf{bridged the borders of these countries. It reminded me of a} documentary I'd watched about the ancient Silk Road that spanned across Iran and China. It touched upon the shared historical narratives of Iran and China, highlighting how goods, ideas, and culture flowed between these two countries for centuries. Intriguingly, the documentary also brought up some parallels between this eastern route and the vibrant exchange between Italy and France during the Renaissance. The cultural and intellectual bonds between Italy and France were indeed profound, just as the resilient trade relations that existed \textbf{between the United States and Canada. The United States and Canada, apart} from their geographical proximity, shared an economic bond that strongly influenced their policies and international standing. Similarly, the ties between Australia and New Zealand shed light on their gestalt identity in the Pacific region. Despite their unique characteristics, Australia and New Zealand were often viewed as a single entity due to their remarkably similar backgrounds in terms of culture, language, and colonial history. Inspired by these chronicles of interconnectedness, I decided to delve deeper into history and found a fascinating account of how Ukraine and Poland had influenced each other \textbf{through years of coexistence. Despite their tumultuous past}, Ukraine and Poland shared a cultural tapestry that was deeply woven into their histories. It was quite similar to the complex relationship between North Korea and South Korea, which, despite their political differences, shared common heritage and traditions. Thus, from Argentina to South Korea, the world was an intricate web of countries intertwined with each other through shared histories, cultures, \textbf{and sometimes, shared destinies.}\\
    
    \textbf{Step 1 -- Response \textcolor{mygreen}{(Fully Correct)}:} \\
    \{\{\\
        \quad \texttt{"Paragraph 1": "My friend, Alex from Peru \textcolor{grey}{... (Omitted) ...} bridged the borders of these countries.",}\\
        \quad \texttt{"Paragraph 2": "It reminded me of a \textcolor{grey}{... (Omitted) ...} between the United States and Canada.",}\\
        \quad \texttt{"Paragraph 3": "The United States and Canada, apart \textcolor{grey}{... (Omitted) ...} through years of coexistence.",}\\
        \quad \texttt{"Paragraph 4": "Despite their tumultuous past \textcolor{grey}{... (Omitted) ...} and sometimes, shared destinies."}\\
    \}\}\\
    \hline
    \end{tabularx}
\end{table*}

\begin{table*}
    \centering
    \renewcommand{\arraystretch}{1.5}
    \caption{Keyword counting for an example 4-passage split (GoT4): \textbf{Step 2a (Prompt/Responses)}}
    \label{tab:keyword_counting_example_part2a}
    \begin{tabularx}{\textwidth}{X}
    \hline
    \textbf{Step 2a -- Prompt:} \\
    \textless{}Instruction\textgreater{} Count the frequency of how many times each country is explicitly named in the input text. You can generate any intermedate lists and states, but the final output should only contain the frequency of each country that appears at least once in the following json format, prefixed with "Output: " (make sure to keep the same spelling for each country in the output as in the input text):\\
    \{\{\\
        \quad \texttt{"country1": frequency1,}\\
        \quad \texttt{"country2": frequency2,}\\
        \quad \texttt{$\dots$}\\
    \}\}\\
    \textless{}/Instruction\textgreater{}\\
    
    \textless{}Approach\textgreater{}\\
    To count the frequency for each country follow these steps:
    \begin{enumerate}
        \item Split the input passage into four paragraphs of similar length.
        \item Count the frequency of each country in each paragraph.
        \item Combine the frequencies of each country from each paragraph by adding them together.
    \end{enumerate}
    \textless{}/Approach\textgreater{}\\
    
    \textless{}Examples\textgreater{} See Table~\ref{tab:keyword_counting_few_shots_count} \textless{}/Examples\textgreater{}\\
    
    Input: My friend, Alex from Peru, once recounted his journey to Argentina where he learned about the strong cultural ties between Argentina and Brazil due to their shared history. He spoke fondly of his time in Argentina and Brazil, marveling at the similar music, dance, and culinary traditions that seamlessly bridged the borders of these countries.\\    
    \textbf{Step 2a -- 10 Responses:}
    \begin{enumerate}
        \item \{\{ \texttt{"Peru": 1, "Argentina": 1, "Brazil": 1} \}\} \textcolor{myblue}{(3 Errors - Missing two "Argentina" and one "Brazil")}
        \item \textbf{\{\{ \texttt{"Peru": 1, "Argentina": 2, "Brazil": 2} \}\} \textcolor{myblue}{(1 Error - Missing one "Argentina")}}
        \item \{\{ \texttt{"Peru": 1, "Argentina": 2, "Brazil": 2} \}\} \textcolor{myblue}{(1 Error - Missing one "Argentina")}
        \item \{\{ \texttt{"Peru": 1, "Argentina": 1, "Brazil": 1} \}\} \textcolor{myblue}{(3 Errors - Missing two "Argentina" and one "Brazil")}
        \item \{\{ \texttt{"Peru": 1, "Argentina": 2, "Brazil": 2} \}\} \textcolor{myblue}{(1 Error - Missing one "Argentina")}
        \item \{\{ \texttt{"Peru": 1, "Argentina": 2, "Brazil": 2} \}\} \textcolor{myblue}{(1 Error - Missing one "Argentina")}
        \item \{\{ \texttt{"Peru": 1, "Argentina": 1, "Brazil": 1} \}\} \textcolor{myblue}{(3 Errors - Missing two "Argentina" and one "Brazil")}
        \item \{\{ \texttt{"Peru": 1, "Argentina": 1, "Brazil": 1} \}\} \textcolor{myblue}{(3 Errors - Missing two "Argentina" and one "Brazil")}
        \item \{\{ \texttt{"Peru": 1, "Argentina": 1, "Brazil": 1} \}\} \textcolor{myblue}{(3 Errors - Missing two "Argentina" and one "Brazil")}
        \item \{\{ \texttt{"Peru": 1, "Argentina": 1, "Brazil": 1} \}\} \textcolor{myblue}{(3 Errors - Missing two "Argentina" and one "Brazil")}
    \end{enumerate}\\
    \hline
    \end{tabularx}
\end{table*}

\begin{table*}
    \centering
    \renewcommand{\arraystretch}{1.5}
    \caption{Keyword counting for an example 4-passage split (GoT4): \textbf{Step 2b (Prompt/Responses)}}
    \label{tab:keyword_counting_example_part2b}
    \begin{tabularx}{\textwidth}{X}
    \hline
    \textbf{Step 2b -- Prompt:} \\
    \textless{}Instruction\textgreater{} Count the frequency of how many times each country is explicitly named in the input text. You can generate any intermedate lists and states, but the final output should only contain the frequency of each country that appears at least once in the following json format, prefixed with "Output: " (make sure to keep the same spelling for each country in the output as in the input text):\\
    \{\{\\
        \quad \texttt{"country1": frequency1,}\\
        \quad \texttt{"country2": frequency2,}\\
        \quad \texttt{$\dots$}\\
    \}\}\\
    \textless{}/Instruction\textgreater{}\\
    
    \textless{}Approach\textgreater{}\\
    To count the frequency for each country follow these steps:
    \begin{enumerate}
        \item Split the input passage into four paragraphs of similar length.
        \item Count the frequency of each country in each paragraph.
        \item Combine the frequencies of each country from each paragraph by adding them together.
    \end{enumerate}
    \textless{}/Approach\textgreater{}\\
    
    \textless{}Examples\textgreater{} See Table~\ref{tab:keyword_counting_few_shots_count} \textless{}/Examples\textgreater{}\\
    
    Input: It reminded me of a documentary I'd watched about the ancient Silk Road that spanned across Iran and China. It touched upon the shared historical narratives of Iran and China, highlighting how goods, ideas, and culture flowed between these two countries for centuries. Intriguingly, the documentary also brought up some parallels between this eastern route and the vibrant exchange between Italy and France during the Renaissance. The cultural and intellectual bonds between Italy and France were indeed profound, just as the resilient trade relations that existed between the United States and Canada.\\    
    \textbf{Step 2b -- 10 Responses:}
    \begin{enumerate}
        \item \{\{ \texttt{"Iran": 1, "China": 1, "Italy": 1, "France": 1, "United States": 1, "Canada": 1} \}\} \newline \textcolor{myblue}{(4 Errors - Missing one "Iran", one "China", one "Italy" and one "France")}
        \item \{\{ \texttt{"Iran": 1, "China": 1, "Italy": 1, "France": 1, "United States": 1, "Canada": 1} \}\} \newline \textcolor{myblue}{(4 Errors - Missing one "Iran", one "China", one "Italy" and one "France")}
        \item \{\{ \texttt{"Iran": 3, "China": 2, "Italy": 1, "France": 1, "United States": 1, "Canada": 1} \}\} \newline \textcolor{myblue}{(3 Errors - Surplus of one "Iran", missing one "Italy" and one "France")}
        \item \{\{ \texttt{"Iran": 1, "China": 1, "Italy": 1, "France": 1, "United States": 1, "Canada": 1} \}\} \newline \textcolor{myblue}{(4 Errors - Missing one "Iran", one "China", one "Italy" and one "France")}
        \item \{\{ \texttt{"Iran": 2, "China": 2, "Italy": 1, "France": 1, "United States": 1, "Canada": 1} \}\} \newline \textcolor{myblue}{(2 Errors - Missing one "Italy" and one "France")}
        \item \{\{ \texttt{"Iran": 1, "China": 1, "Italy": 1, "France": 1, "United States": 1, "Canada": 1} \}\} \newline \textcolor{myblue}{(4 Errors - Missing one "Iran", one "China", one "Italy" and one "France")}
        \item \{\{ \texttt{"Iran": 1, "China": 1, "Italy": 1, "France": 1, "United States": 1, "Canada": 1} \}\} \newline \textcolor{myblue}{(4 Errors - Missing one "Iran", one "China", one "Italy" and one "France")}
        \item \{\{ \texttt{"Iran": 1, "China": 1, "Italy": 1, "France": 1, "United States": 1, "Canada": 1} \}\} \newline \textcolor{myblue}{(4 Errors - Missing one "Iran", one "China", one "Italy" and one "France")}
        \item \{\{ \texttt{"Iran": 1, "China": 1, "Italy": 1, "France": 1, "United States": 1, "Canada": 1} \}\} \newline \textcolor{myblue}{(4 Errors - Missing one "Iran", one "China", one "Italy" and one "France")}
        \item \textbf{\{\{ \texttt{"Iran": 2, "China": 2, "Italy": 1, "France": 2, "United States": 1, "Canada": 1} \}\} \newline \textcolor{myblue}{(1 Error - Missing one "Italy")}}
    \end{enumerate}\\
    \hline
    \end{tabularx}
\end{table*}

\begin{table*}
    \centering
    \renewcommand{\arraystretch}{1.5}
    \caption{Keyword counting for an example 4-passage split (GoT4): \textbf{Step 2c (Prompt/Responses)}}
    \label{tab:keyword_counting_example_part2c}
    \begin{tabularx}{\textwidth}{X}
    \hline
    \textbf{Step 2c -- Prompt:} \\
    \textless{}Instruction\textgreater{} Count the frequency of how many times each country is explicitly named in the input text. You can generate any intermedate lists and states, but the final output should only contain the frequency of each country that appears at least once in the following json format, prefixed with "Output: " (make sure to keep the same spelling for each country in the output as in the input text):\\
    \{\{\\
        \quad \texttt{"country1": frequency1,}\\
        \quad \texttt{"country2": frequency2,}\\
        \quad \texttt{$\dots$}\\
    \}\}\\
    \textless{}/Instruction\textgreater{}\\
    
    \textless{}Approach\textgreater{}\\
    To count the frequency for each country follow these steps:
    \begin{enumerate}
        \item Split the input passage into four paragraphs of similar length.
        \item Count the frequency of each country in each paragraph.
        \item Combine the frequencies of each country from each paragraph by adding them together.
    \end{enumerate}
    \textless{}/Approach\textgreater{}\\
    
    \textless{}Examples\textgreater{} See Table~\ref{tab:keyword_counting_few_shots_count} \textless{}/Examples\textgreater{}\\
    
    Input: The United States and Canada, apart from their geographical proximity, shared an economic bond that strongly influenced their policies and international standing. Similarly, the ties between Australia and New Zealand shed light on their gestalt identity in the Pacific region. Despite their unique characteristics, Australia and New Zealand were often viewed as a single entity due to their remarkably similar backgrounds in terms of culture, language, and colonial history. Inspired by these chronicles of interconnectedness, I decided to delve deeper into history and found a fascinating account of how Ukraine and Poland had influenced each other through years of coexistence.\\    
    \textbf{Step 2c -- 10 Responses:}
    \begin{enumerate}
        \item \textbf{\{\{ \texttt{"United States": 1, "Canada": 1, "Australia": 2, "New Zealand": 2, "Ukraine": 1, "Poland": 1} \}\} \newline \textcolor{mygreen}{(Fully Correct)}}
        \item \{\{ \texttt{"United States": 1, "Canada": 1, "Australia": 2, "New Zealand": 2, "Ukraine": 1, "Poland": 1} \}\} \newline \textcolor{mygreen}{(Fully Correct)}
        \item \{\{ \texttt{"United States": 1, "Canada": 1, "Australia": 1, "New Zealand": 1, "Australia and New Zealand": 1, "Ukraine": 1, "Poland": 1} \}\} \newline \textcolor{myblue}{(3 Errors - Missing one "Australia, one "New Zealand" and surplus of one "Australia and New Zealand")}
        \item \{\{ \texttt{"United States": 1, "Canada": 1, "Australia": 2, "New Zealand": 2, "Ukraine": 1, "Poland": 1} \}\} \newline \textcolor{mygreen}{(Fully Correct)}
        \item \{\{ \texttt{"United States": 1, "Canada": 1, "Australia": 1, "New Zealand": 1, "Australia and New Zealand": 1, "Ukraine": 1, "Poland": 1} \}\} \newline \textcolor{myblue}{(3 Errors - Missing one "Australia, one "New Zealand" and surplus of one "Australia and New Zealand")}
        \item \{\{ \texttt{"United States": 1, "Canada": 1, "Australia": 1, "New Zealand": 1, "Ukraine": 1, "Poland": 1} \}\} \newline \textcolor{myblue}{(2 Errors - Missing one "Australia" and one "New Zealand")}
        \item \{\{ \texttt{"United States": 1, "Canada": 1, "Australia": 2, "New Zealand": 2, "Ukraine": 1, "Poland": 1} \}\} \newline \textcolor{mygreen}{(Fully Correct)}
        \item \{\{ \texttt{"United States": 1, "Canada": 1, "Australia": 1, "New Zealand": 1, "Ukraine": 1, "Poland": 1} \}\} \newline \textcolor{myblue}{(2 Errors - Missing one "Australia" and one "New Zealand")}
        \item \{\{ \texttt{"United States": 1, "Canada": 1, "Australia": 2, "New Zealand": 2, "Ukraine": 1, "Poland": 1} \}\} \newline \textcolor{mygreen}{(Fully Correct)}
        \item \{\{ \texttt{"United States": 1, "Canada": 1, "Australia": 2, "New Zealand": 2, "Ukraine": 1, "Poland": 1} \}\} \newline \textcolor{mygreen}{(Fully Correct)}
    \end{enumerate}\\
    \hline
    \end{tabularx}
\end{table*}

\begin{table*}
    \centering
    \renewcommand{\arraystretch}{1.5}
    \caption{Keyword counting for an example 4-passage split (GoT4): \textbf{Step 2d (Prompt/Responses)}}
    \label{tab:keyword_counting_example_part2d}
    \begin{tabularx}{\textwidth}{X}
    \hline
    \textbf{Step 2d -- Prompt:} \\
    \textless{}Instruction\textgreater{} Count the frequency of how many times each country is explicitly named in the input text. You can generate any intermedate lists and states, but the final output should only contain the frequency of each country that appears at least once in the following json format, prefixed with "Output: " (make sure to keep the same spelling for each country in the output as in the input text):\\
    \{\{\\
        \quad \texttt{"country1": frequency1,}\\
        \quad \texttt{"country2": frequency2,}\\
        \quad \texttt{$\dots$}\\
    \}\}\\
    \textless{}/Instruction\textgreater{}\\
    
    \textless{}Approach\textgreater{}\\
    To count the frequency for each country follow these steps:
    \begin{enumerate}
        \item Split the input passage into four paragraphs of similar length.
        \item Count the frequency of each country in each paragraph.
        \item Combine the frequencies of each country from each paragraph by adding them together.
    \end{enumerate}
    \textless{}/Approach\textgreater{}\\
    
    \textless{}Examples\textgreater{} See Table~\ref{tab:keyword_counting_few_shots_count} \textless{}/Examples\textgreater{}\\
    
    Input: Despite their tumultuous past, Ukraine and Poland shared a cultural tapestry that was deeply woven into their histories. It was quite similar to the complex relationship between North Korea and South Korea, which, despite their political differences, shared common heritage and traditions. Thus, from Argentina to South Korea, the world was an intricate web of countries intertwined with each other through shared histories, cultures, and sometimes, shared destinies.\\    
    \textbf{Step 2d -- 10 Responses:}
    \begin{enumerate}
        \item \textbf{\{\{ \texttt{"Ukraine": 1, "Poland": 1, "North Korea": 1, "South Korea": 2, "Argentina": 1 } \}\} \textcolor{mygreen}{(Fully Correct)}}
        \item \{\{ \texttt{"Ukraine": 1, "Poland": 1, "North Korea": 1, "South Korea": 2, "Argentina": 1 } \}\} \textcolor{mygreen}{(Fully Correct)}
        \item \{\{ \texttt{"Ukraine": 1, "Poland": 1, "North Korea": 1, "South Korea": 2, "Argentina": 1 } \}\} \textcolor{mygreen}{(Fully Correct)}
        \item \{\{ \texttt{"Ukraine": 1, "Poland": 1, "North Korea": 1, "South Korea": 2, "Argentina": 1 } \}\} \textcolor{mygreen}{(Fully Correct)}
        \item \{\{ \texttt{"Ukraine": 1, "Poland": 1, "North Korea": 1, "South Korea": 2, "Argentina": 1 } \}\} \textcolor{mygreen}{(Fully Correct)}
        \item \{\{ \texttt{"Ukraine": 1, "Poland": 1, "North Korea": 1, "South Korea": 2, "Argentina": 1 } \}\} \textcolor{mygreen}{(Fully Correct)}
        \item \{\{ \texttt{"Ukraine": 1, "Poland": 1, "North Korea": 1, "South Korea": 2, "Argentina": 1 } \}\} \textcolor{mygreen}{(Fully Correct)}
        \item \{\{ \texttt{"Ukraine": 1, "Poland": 1, "North Korea": 1, "South Korea": 2, "Argentina": 1 } \}\} \textcolor{mygreen}{(Fully Correct)}
        \item \{\{ \texttt{"Ukraine": 1, "Poland": 1, "North Korea": 1, "South Korea": 2, "Argentina": 1 } \}\} \textcolor{mygreen}{(Fully Correct)}
        \item \{\{ \texttt{"Ukraine": 1, "Poland": 1, "North Korea": 1, "South Korea": 2, "Argentina": 1 } \}\} \textcolor{mygreen}{(Fully Correct)}
    \end{enumerate}\\
    \hline
    \end{tabularx}
\end{table*}

\begin{table*}
    \centering
    \renewcommand{\arraystretch}{1.5}
    \caption{Keyword counting for an example 4-passage split (GoT4): \textbf{Step 3a (Prompt/Responses)}}
    \label{tab:keyword_counting_example_part3a}
    \begin{tabularx}{\textwidth}{X}
    \hline
    \textbf{Step 3a -- Prompt:} \\
    \textless{}Instruction\textgreater{} Combine the following 2 dictionaries, each containing the frequency of countries in a text, into a single dictionary.
    Simply add the frequencies together for each country and if a country is not present in one of the dictionaries, add it to the final dictionary with the frequency from the other dictionary.\\
    Only output the final merged dictionary without any additional text or thoughts! \textless{}/Instruction\textgreater{}\\
    
    \textless{}Approach\textgreater{}\\
    To combine the 2 dictionaries into single one, follow these steps:
    \begin{enumerate}
        \item Create a new dictionary to store the combined frequencies.
        \item Iterate through the keys of the first dictionary and add the frequency of each country to the new dictionary.
        \item Iterate through the keys of the second dictionary and add the frequency of each country to the new dictionary and if it is already present, add the frequency to the existing value.
    \end{enumerate}
    \textless{}/Approach\textgreater{}\\
    
    Combine the following 2 dictionaries into a single dictionary:\\
    \{\{ \texttt{"Peru": 1, "Argentina": 2, "Brazil": 2} \}\}\\
    \{\{ \texttt{"Iran": 2, "China": 2, "Italy": 1, "France": 2, "United States": 1, "Canada": 1} \}\}\\
    Combined Output:\\    
    \textbf{Step 3a -- 3 Responses:}
    \begin{enumerate}
        \item \textbf{\{\{ \texttt{"Peru": 1, "Argentina": 2, "Brazil": 2, "Iran": 2, "China": 2, "Italy": 1, "France": 2, "United States": 1, "Canada": 1 } \}\} \textcolor{mygreen}{(Valid - No Improvement)} \textcolor{myblue}{(2 Errors - from counting)}}
        \item \{\{ \texttt{"Peru": 1, "Argentina": 2, "Brazil": 2, "Iran": 2, "China": 2, "Italy": 1, "France": 2, "United States": 1, "Canada": 1 } \}\} \textcolor{mygreen}{(Valid - No Improvement)} \textcolor{myblue}{(2 Errors - from counting)}
        \item \{\{ \texttt{"Peru": 1, "Argentina": 2, "Brazil": 2, "Iran": 2, "China": 2, "Italy": 1, "France": 2, "United States": 1, "Canada": 1 } \}\} \textcolor{mygreen}{(Valid - No Improvement)} \textcolor{myblue}{(2 Errors - from counting)}
    \end{enumerate}\\
    \hline
    \end{tabularx}
\end{table*}

\begin{table*}
    \centering
    \renewcommand{\arraystretch}{1.5}
    \caption{Keyword counting for an example 4-passage split (GoT4): \textbf{Step 3b (Prompt/Responses)}}
    \label{tab:keyword_counting_example_part3b}
    \begin{tabularx}{\textwidth}{X}
    \hline
    \textbf{Step 3b -- Prompt:} \\
    \textless{}Instruction\textgreater{} Combine the following 2 dictionaries, each containing the frequency of countries in a text, into a single dictionary.
    Simply add the frequencies together for each country and if a country is not present in one of the dictionaries, add it to the final dictionary with the frequency from the other dictionary.\\
    Only output the final merged dictionary without any additional text or thoughts! \textless{}/Instruction\textgreater{}\\
    
    \textless{}Approach\textgreater{}\\
    To combine the 2 dictionaries into single one, follow these steps:
    \begin{enumerate}
        \item Create a new dictionary to store the combined frequencies.
        \item Iterate through the keys of the first dictionary and add the frequency of each country to the new dictionary.
        \item Iterate through the keys of the second dictionary and add the frequency of each country to the new dictionary and if it is already present, add the frequency to the existing value.
    \end{enumerate}
    \textless{}/Approach\textgreater{}\\
    
    Combine the following 2 dictionaries into a single dictionary:\\
    \{\{ \texttt{"United States": 1, "Canada": 1, "Australia": 2, "New Zealand": 2, "Ukraine": 1, "Poland": 1} \}\}\\
    \{\{ \texttt{"Ukraine": 1, "Poland": 1, "North Korea": 1, "South Korea": 2, "Argentina": 1 } \}\}\\
    Combined Output:\\ 
    \textbf{Step 3b -- 3 Responses:}
    \begin{enumerate}
        \item \textbf{\{\{ \texttt{"United States": 1, "Canada": 1, "Australia": 2, "New Zealand": 2, "Ukraine": 2, "Poland": 2, "North Korea": 1, "South Korea": 2, "Argentina": 1 } \}\} \textcolor{mygreen}{(Valid - No Improvement)} \textcolor{mygreen}{(Fully Correct)}}
        \item \{\{ \texttt{"United States": 1, "Canada": 1, "Australia": 2, "New Zealand": 2, "Ukraine": 2, "Poland": 2, "North Korea": 1, "South Korea": 2, "Argentina": 1 } \}\} \textcolor{mygreen}{(Valid - No Improvement)} \textcolor{mygreen}{(Fully Correct)}
        \item \{\{ \texttt{"United States": 1, "Canada": 1, "Australia": 2, "New Zealand": 2, "Ukraine": 2, "Poland": 2, "North Korea": 1, "South Korea": 2, "Argentina": 1 } \}\} \textcolor{mygreen}{(Valid - No Improvement)} \textcolor{mygreen}{(Fully Correct)}
    \end{enumerate}\\
    \hline
    \end{tabularx}
\end{table*}

\begin{table*}
    \centering
    \renewcommand{\arraystretch}{1.5}
    \caption{Keyword counting for an example 4-passage split (GoT4): \textbf{Step 3c (Prompt/Responses)}}
    \label{tab:keyword_counting_example_part3c}
    \begin{tabularx}{\textwidth}{X}
    \hline
    \textbf{Step 3c -- Prompt:} \\
    \textless{}Instruction\textgreater{} Combine the following 2 dictionaries, each containing the frequency of countries in a text, into a single dictionary.
    Simply add the frequencies together for each country and if a country is not present in one of the dictionaries, add it to the final dictionary with the frequency from the other dictionary.\\
    Only output the final merged dictionary without any additional text or thoughts! \textless{}/Instruction\textgreater{}\\
    
    \textless{}Approach\textgreater{}\\
    To combine the 2 dictionaries into single one, follow these steps:
    \begin{enumerate}
        \item Create a new dictionary to store the combined frequencies.
        \item Iterate through the keys of the first dictionary and add the frequency of each country to the new dictionary.
        \item Iterate through the keys of the second dictionary and add the frequency of each country to the new dictionary and if it is already present, add the frequency to the existing value.
    \end{enumerate}
    \textless{}/Approach\textgreater{}\\
    
    Combine the following 2 dictionaries into a single dictionary:\\
    \{\{ \texttt{"Peru": 1, "Argentina": 2, "Brazil": 2, "Iran": 2, "China": 2, "Italy": 1, "France": 2, "United States": 1, "Canada": 1 } \}\}\\
    \{\{ \texttt{"United States": 1, "Canada": 1, "Australia": 2, "New Zealand": 2, "Ukraine": 2, "Poland": 2, "North Korea": 1, "South Korea": 2, "Argentina": 1 } \}\}\\
    Combined Output:\\ 
    \textbf{Step 3c -- 3 Responses:}
    \begin{enumerate}
        \item \textbf{\{\{ \texttt{"Peru": 1, "Argentina": 3, "Brazil": 2, "Iran": 2, "China": 2, "Italy": 1, "France": 2, "United States": 2, "Canada": 2, "Australia": 2, "New Zealand": 2, "Ukraine": 2, "Poland": 2, "North Korea": 1, "South Korea": 2 } \}\} \textcolor{mygreen}{(Valid - No Improvement)} \textcolor{myblue}{(2 Errors - from counting)}}
        \item \{\{ \texttt{"Peru": 1, "Argentina": 3, "Brazil": 2, "Iran": 2, "China": 2, "Italy": 1, "France": 2, "United States": 2, "Canada": 2, "Australia": 2, "New Zealand": 2, "Ukraine": 2, "Poland": 2, "North Korea": 1, "South Korea": 2 } \}\} \textcolor{mygreen}{(Valid - No Improvement)} \textcolor{myblue}{(2 Errors - from counting)}
        \item \{\{ \texttt{"Peru": 1, "Argentina": 3, "Brazil": 2, "Iran": 2, "China": 2, "Italy": 1, "France": 2, "United States": 2, "Canada": 2, "Australia": 2, "New Zealand": 2, "Ukraine": 2, "Poland": 2, "North Korea": 1, "South Korea": 2 } \}\} \textcolor{mygreen}{(Valid - No Improvement)} \textcolor{myblue}{(2 Errors - from counting)}
    \end{enumerate}\\
    \hline
    \textbf{Final Result \textcolor{myblue}{(2 Errors)}:}\\
    \{\{ \texttt{"Peru": 1, "Argentina": 3, "Brazil": 2, "Iran": 2, "China": 2, "Italy": 1, "France": 2, "United States": 2, "Canada": 2, "Australia": 2, "New Zealand": 2, "Ukraine": 2, "Poland": 2, "North Korea": 1, "South Korea": 2 } \}\} \\
    \hline
    \end{tabularx}
\end{table*}

\clearpage

\section{Example Prompts - Document Merging}
We present the prompts only for \texttt{GoT} of the document merging task, as \texttt{GoT2} only differs in the fact that it merges the 4 NDAs in 2 steps rather than 1. 
For document merging, we employ four distinct types of operations that interact with the LLM, each with its corresponding prompts. 
First, there is the \textbf{Generate} operation, utilizing the \textit{merge\_prompt} to instruct the LLM to merge the 4 NDAs into 1. 
Second, the \textbf{Score} operations instructs the LLM to score a given merged NDA using the \textit{score\_prompt}.
Next, the \textbf{Aggregate} operation employs the \textit{aggregate\_prompt} to instruct the LLM to aggregate multiple merge attempts into a single, better one.
Finally, the \textbf{Improve} operation leverages the \textit{improve\_prompt} to instruct the LLM to improve a merged NDA. 

First, we present the prompt stubs (Table \ref{tab:doc_prompt_stubs} - Table \ref{tab:doc_prompt_stubs_continued}), serving as templates to dynamically generate appropriate prompts at runtime. Following this, we outline the LLM interactions throughout a complete merging process (Table \ref{tab:doc_example_nda12} - Table \ref{tab:doc_example_finale}).
However, instead of displaying each input/generated NDA in every prompt/response, we present the 4 input NDAs in Table \ref{tab:doc_example_nda12} - Table \ref{tab:doc_example_nda4} and the final merged NDA in Table \ref{tab:doc_example_finale}.
Furthermore, as scoring is done using the LLM as well, we will present these interactions for the best performing merged NDAs (Tables \ref{tab:doc_example_scoring} - \ref{tab:doc_example_scoring_r} and Tables \ref{tab:doc_example_scoring_s3} - \ref{tab:doc_example_scoring_s3_r}).
Lastly, most responses are limited to a few lines only, as they don't offer any further insights and would otherwise span multiple pages.
However, we refer the interested reader to the results in the corresponding code repository\footnote{https://github.com/spcl/graph-of-thoughts} for full logs and further examples. 

\begin{table*}
    \centering
    \renewcommand{\arraystretch}{1.5}
    \caption{Prompt stubs for the document merging task; parameters in single curly brackets will be substituted at runtime.}
    \label{tab:doc_prompt_stubs}
    \begin{tabularx}{\textwidth}{X}
    \hline
    \textbf{merge\_prompt:} 
    Merge the following 4 NDA documents \textless{}Doc1\textgreater{} - \textless{}Doc4\textgreater{} into a single NDA, maximizing retained information and minimizing redundancy. Output only the created NDA between the tags \textless{}Merged\textgreater{} and \textless{}/Merged\textgreater{}, without any additional text. \\
    Here are NDAs \textless{}Doc1\textgreater{} - \textless{}Doc4\textgreater{}: \\
    \textless{}Doc1\textgreater{}
    \{doc1\}
    \textless{}/Doc1\textgreater{}\\
    \textless{}Doc2\textgreater{}
    \{doc2\}
    \textless{}/Doc2\textgreater{}\\
    \textless{}Doc3\textgreater{}
    \{doc3\}
    \textless{}/Doc3\textgreater{}\\
    \textless{}Doc4\textgreater{}
    \{doc4\}
    \textless{}/Doc4\textgreater{}\\
    \hline
    \textbf{score\_prompt:}
    The following NDA \textless{}S\textgreater{} merges NDAs \textless{}Doc1\textgreater{} - \textless{}Doc4\textgreater{}. \\
    Please score the merged NDA \textless{}S\textgreater{} in terms of how much redundant information is contained, independent of the original NDAs, as well as how much information is retained from the original NDAs. \\
    A score of 10 for redundancy implies that absolutely no information is redundant, while a score of 0 implies that at least half of the information is redundant (so everything is at least mentioned twice). \\
    A score of 10 for retained information implies that all information from the original NDAs is retained, while a score of 0 implies that no information is retained. \\
    You may provide reasoning for your scoring, but the final score for redundancy should be between the tags \textless{}Redundancy\textgreater{} and \textless{}/Redundancy\textgreater{}, and the final score for retained information should be between the tags \textless{}Retained\textgreater{} and \textless{}/Retained\textgreater{}, without any additional text within any of those tags. \\

    Here are NDAs \textless{}Doc1\textgreater{} - \textless{}Doc4\textgreater{}: \\
    \textless{}Doc1\textgreater{}
    \{doc1\}
    \textless{}/Doc1\textgreater{}\\
    \textless{}Doc2\textgreater{}
    \{doc2\}
    \textless{}/Doc2\textgreater{}\\
    \textless{}Doc3\textgreater{}
    \{doc3\}
    \textless{}/Doc3\textgreater{}\\
    \textless{}Doc4\textgreater{}
    \{doc4\}
    \textless{}/Doc4\textgreater{}\\
    Here is the merged NDA \textless{}S\textgreater{}:\\
    \textless{}S\textgreater{}
    \{s\}
    \textless{}/S\textgreater{}\\
    \hline
    \textbf{aggregate\_prompt:} 
    The following NDAs \textless{}S1\textgreater{} - \textless{}S\{num\_ndas\_summaries\}\textgreater{} each merge the initial NDAs \textless{}Doc1\textgreater{} - \textless{}Doc4\textgreater{}.\\
    Combine the merged NDAs \textless{}S1\textgreater{} - \textless{}S\{num\_ndas\_summaries\}\textgreater{} into a new one, maximizing their advantages and overall information retention, while minimizing redundancy.\\
    Output only the new NDA between the tags \textless{}Merged\textgreater{} and \textless{}/Merged\textgreater{}, without any additional text.\\\\

    Here are the original NDAs \textless{}Doc1\textgreater{} - \textless{}Doc4\textgreater{}: \\
    \textless{}Doc1\textgreater{}
    \{doc1\}
    \textless{}/Doc1\textgreater{}\\
    \textless{}Doc2\textgreater{}
    \{doc2\}
    \textless{}/Doc2\textgreater{}\\
    \textless{}Doc3\textgreater{}
    \{doc3\}
    \textless{}/Doc3\textgreater{}\\
    \textless{}Doc4\textgreater{}
    \{doc4\}
    \textless{}/Doc4\textgreater{}\\

    Here are the merged NDAs \textless{}S1\textgreater{} - \textless{}S\{num\_ndas\_summaries\}\textgreater: \\
    \textless{}S1\textgreater{}
    \{s1\}
    \textless{}/S1\textgreater{}\\
    $\dots$\\
    \textless{}S\{num\_ndas\_summaries\}\textgreater{}
    \{s\{num\_ndas\_summaries\}\}
    \textless{}/S\{num\_ndas\_summaries\}\textgreater{}\\
    \hline
    \end{tabularx}
\end{table*}

\begin{table*}
    \centering
    \renewcommand{\arraystretch}{1.5}
    \caption{Prompt stubs for the document merging task \textit{continued}; parameters in single curly brackets will be substituted at runtime.}
    \label{tab:doc_prompt_stubs_continued}
    \begin{tabularx}{\textwidth}{X}
    \hline
    \textbf{improve\_prompt:}
    The following NDA \textless{}S\textgreater{} merges initial NDAs \textless{}Doc1\textgreater{} - \textless{}Doc4\textgreater{}. \\
    Please improve the merged NDA \textless{}S\textgreater{} by adding more information and removing redundancy.
    Output only the improved NDA, placed between the tags \textless{}Merged\textgreater{} and \textless{}/Merged\textgreater{}, without any additional text. \\
    Here are NDAs \textless{}Doc1\textgreater{} - \textless{}Doc4\textgreater{}: \\
    \textless{}Doc1\textgreater{}
    \{doc1\}
    \textless{}/Doc1\textgreater{}\\
    \textless{}Doc2\textgreater{}
    \{doc2\}
    \textless{}/Doc2\textgreater{}\\
    \textless{}Doc3\textgreater{}
    \{doc3\}
    \textless{}/Doc3\textgreater{}\\
    \textless{}Doc4\textgreater{}
    \{doc4\}
    \textless{}/Doc4\textgreater{}\\
    Here is the merged NDA \textless{}S\textgreater{}:\\
    \textless{}S\textgreater{}
    \{s\}
    \textless{}/S\textgreater{}\\
    \hline
    \end{tabularx}
\end{table*}

\begin{table*}
    \centering
    \renewcommand{\arraystretch}{1.5}
    \caption{Input NDA 1 and 2}
    \label{tab:doc_example_nda12}
    \begin{tabularx}{\textwidth}{X}
    \hline
    \textless{}Doc1\textgreater{}\\
    NON-DISCLOSURE AGREEMENT (NDA) 
    \begin{enumerate}
        \item Agreement between [Your Company Name] and [Recipient Name] on [Date].
        \item Information sharing for the purpose of [specific project or purpose].
        \item "Confidential Information" includes all potentially commercially valuable information, specifically software development tactics, processes, and in-house research results.
        \item Receiving party is obligated to protect the Confidential Information, use it solely for the disclosed purpose, and not disclose it without consent.
        \item Breach penalties include injunctive relief, other remedies, and a \$200,000 fee per breach.
        \item The Agreement applies to the Parties and their successors and assigns. It contains all related agreements and lack of enforcement doesn't imply waiver.
        \item The Agreement is under the laws of [State].
        \item Signed by [Your Company Name] and [Recipient Name] at the above date.
    \end{enumerate}
    \textless{}/Doc1\textgreater{}\\
    \hline
    \textless{}Doc2\textgreater{}\\
    NON-DISCLOSURE AGREEMENT (NDA)\\
    Effective from [Effective Date], this NDA involves [Your Company Name] ("Disclosing Party"), and [Recipient Name] ("Receiving Party").
    \begin{enumerate}
        \item Purpose: The Disclosing Party will disclose confidential information related to [Topic of Research] to the Receiving Party for [Purpose].
        \item Confidential Information: Defined as all non-public reports, data, designs, and other materials provided by the Disclosing Party to the Receiving Party.
        \item Receiving Party's Obligations:
        \begin{enumerate}[label=\alph*.]
            \item Use, reproduce, or distribute the confidential information only for the agreed purpose.
            \item Restrict access to the information to necessary parties, ensuring they abide by strict confidentiality.
            \item Return or destroy all confidential information upon request or at the end of the agreement.
        \end{enumerate}
        \item Exclusions: Information will not be classified as confidential if it is already known to the Receiving Party, publicly known, or independently developed by the Receiving Party.
        \item Non-Competition: The Receiving Party will not engage in any competing business against the Disclosing Party during the agreement and one year after its termination.
        \item Term and Termination: The agreement is valid for [e.g., "two years"], unless terminated earlier with [e.g., "30 days"] written notice. The Receiving Party's non-disclosure and non-competition obligations persist post-termination.

        \item General Provisions:
        \begin{enumerate}[label=\alph*.]
            \item Governing Law: [Your State]'s laws apply.
            \item Amendments: Only valid if written and signed by both parties.
            \item Entire Agreement: This contract overrules previous related agreements.
        \end{enumerate}
    \end{enumerate}
    Signed as of the Effective Date by [Your Company Name] - Disclosing Party [Recipient Name] - Receiving Party. \\
    \textless{}/Doc2\textgreater{}\\
    \hline
    \end{tabularx}
\end{table*}

\begin{table*}
    \centering
    \renewcommand{\arraystretch}{1.5}
    \caption{Input NDA 3}
    \label{tab:doc_example_nda3}
    \begin{tabularx}{\textwidth}{X}
    \hline
    \textless{}Doc3\textgreater{}\\
    CONFIDENTIALITY \& NON-DISCLOSURE AGREEMENT \\
    Entities Involved: \\
    Effective [Date], between [AquaBlue Innovations], established in [State], and [PineTree Solutions], a registered entity.\\
    Objective:\\
    To safeguard classified data during talks of a potential technological alliance.\\
    Specification of Protected Information:\\
    Particularly:
    \begin{enumerate}[label=\alph*.]
        \item System designs and architectural schematics.
        \item Proprietary computational algorithms.
    \end{enumerate}
    Receiver's Obligations:
    \begin{enumerate}[label=\alph*.]
        \item Maintain strict non-disclosure using best practices.
        \item Employ solely for the aforementioned aim.
        \item No unveiling without explicit authorization.
    \end{enumerate}
    Violation Ramifications:\\
    A charge of \$280,000 for every infringement, plus possible legal proceedings.\\
    General Terms:\\
    Binding for both parties and any successors. This encapsulates the entire accord.\\
    Legal Reference:\\
    Governed as per [State]'s legal framework.\\
    Attestation:\\
    Duly signed on [Date].
    
    [AquaBlue Innovations] [PineTree Solutions]\\
    \textless{}/Doc3\textgreater{}\\
    \hline
    \end{tabularx}
\end{table*}

\begin{table*}
    \centering
    \renewcommand{\arraystretch}{1.5}
    \caption{Input NDA 4}
    \label{tab:doc_example_nda4}
    \begin{tabularx}{\textwidth}{X}
    \hline
    \textless{}Doc4\textgreater{}\\
    SECRECY \& DISCLOSURE AGREEMENT \\
    Contracting Parties: \\
    Dated [Date], drawn between [AquaBlue Innovations], a [State]-based corporation, and [PineTree Solutions], a licensed organization.\\
    Aim:\\
    To protect exclusive insights amidst dialogues for a technological partnership.\\
    Categorization of Sensitive Data:\\
    Includes:
    \begin{enumerate}[label=\alph*.]
        \item Internal software blueprints.
        \item Intellectual property awaiting patents.
    \end{enumerate}
    Commitments of Recipient:
    \begin{enumerate}[label=\alph*.]
        \item Uphold confidentiality, ensuring data integrity.
        \item Utilize strictly for collaborative ventures.
        \item No exposure without prior consensus.
    \end{enumerate}
    Repercussions for Non-Compliance: \\
    \$295,000 fine for each transgression, and the option for legal recourse. \\
    Overall Provisions:\\
    Legally enforceable for signatories and successors. Complete and sole agreement.\\
    Juridical Standpoint:\\
    Under the auspices of [State] laws.\\
    Ratification:\\
    Confirmed and endorsed on [Date].

    [AquaBlue Innovations] [PineTree Solutions]\\
    \textless{}/Doc4\textgreater{}\\
    \hline
    \end{tabularx}
\end{table*}

\begin{table*}
    \centering
    \renewcommand{\arraystretch}{1.5}
    \caption{Merging 4 example NDAs: Execution plan (\textbf{GoO})}
    \label{tab:doc_example_goo}
    \begin{tabularx}{\textwidth}{X}
    \hline
    \textbf{GoO:} 
    \begin{enumerate}
        \item Merge the 4 NDAs into a single one 5 times; Score each attempt and keep the best 3
        \item Aggregate the merge attempts into a single one 5 times; Score each aggregation attempt and keep the overall best attempt (including Step 1)
        \item Improve the merged NDA 10 times; Score each and keep the best
    \end{enumerate}\\
    \hline
    \end{tabularx}
\end{table*}

\begin{table*}
    \centering
    \renewcommand{\arraystretch}{1.5}
    \caption{Merging 4 example NDAs: \textbf{Step 1 (Prompt)}}
    \label{tab:doc_example_step1p}
    \begin{tabularx}{\textwidth}{X}
    \hline
    %\textbf{Step 1 -- Prompt:} \\
    Merge the following 4 NDA documents \textless{}Doc1\textgreater{} - \textless{}Doc4\textgreater{} into a single NDA, maximizing retained information and minimizing redundancy. Output only the created NDA between the tags \textless{}Merged\textgreater{} and \textless{}/Merged\textgreater{}, without any additional text.\\
    Here are NDAs \textless{}Doc1\textgreater{} - \textless{}Doc4\textgreater{}:\\
    \textless{}Doc1\textgreater{}\\
    NON-DISCLOSURE AGREEMENT (NDA) 
    \begin{enumerate}
        \item Agreement between [Your Company Name] and [Recipient Name] on [Date].
        \item Information sharing for the purpose of [specific project or purpose].
    \end{enumerate}
    \textcolor{grey}{$\dots$ (Omitted, see Table \ref{tab:doc_example_nda12})} \\
    \textless{}/Doc1\textgreater{}\\
    \textless{}Doc2\textgreater{}\\
    NON-DISCLOSURE AGREEMENT (NDA)\\
    Effective from [Effective Date], this NDA involves [Your Company Name] ("Disclosing Party"), and [Recipient Name] ("Receiving Party").
    \begin{enumerate}
        \item Purpose: The Disclosing Party will disclose confidential information related to [Topic of Research] to the Receiving Party for [Purpose].
    \end{enumerate}
    \textcolor{grey}{$\dots$ (Omitted, see Table \ref{tab:doc_example_nda12})}\\
    \textless{}/Doc2\textgreater{}\\
    \textless{}Doc3\textgreater{}\\
    CONFIDENTIALITY \& NON-DISCLOSURE AGREEMENT \\
    Entities Involved: \\
    Effective [Date], between [AquaBlue Innovations], established in [State], and [PineTree Solutions], a registered entity.\\
    Objective:\\
    To safeguard classified data during talks of a potential \textcolor{grey}{$\dots$ (Omitted, see Table \ref{tab:doc_example_nda3})} \\
    \textless{}/Doc3\textgreater{}\\
    \textless{}Doc4\textgreater{}\\
    SECRECY \& DISCLOSURE AGREEMENT \\
    Contracting Parties: \\
    Dated [Date], drawn between [AquaBlue Innovations], a [State]-based corporation, and [PineTree Solutions], a licensed organization.\\
    Aim:\\
    To protect exclusive insights amidst \textcolor{grey}{$\dots$ (Omitted, see Table \ref{tab:doc_example_nda4})} \\
    \textless{}/Doc4\textgreater{}\\
    \hline
    \end{tabularx}
\end{table*}

\begin{table*}
    \centering
    \renewcommand{\arraystretch}{1.5}
    \caption{Merging 4 example NDAs; we only present the highest scoring response fully in Tables \ref{tab:doc_example_step1b1}-\ref{tab:doc_example_step1b2}: \textbf{Step 1 (Responses)}}
    \label{tab:doc_example_step1r}
    \begin{tabularx}{\textwidth}{X}
    \hline
    Note that due to the shortening of responses, the appear to be equal, but small differences can be found when observing the fully generated summaries. The only small differences are reflected in the very similar scores (see Table \ref{tab:doc_example_scoring}-\ref{tab:doc_example_scoring_r} for the scoring of the best response).\\
    \textbf{Response (1/5):} \\
    \textless{}Merged\textgreater{}\\
    NON-DISCLOSURE AGREEMENT (NDA)
    \begin{enumerate}
        \item Agreement between [Your Company Name] and [Recipient Name] on [Date].
        \item Information sharing for the purpose of [specific project or purpose].
    \end{enumerate}
    \textcolor{grey}{$\dots$ (Omitted)}\\
    \textcolor{myblue}{Score: $6.60$}\\
    \textbf{Response (2/5):}\\
    \textless{}Merged\textgreater{}\\
    NON-DISCLOSURE AGREEMENT (NDA)
    \begin{enumerate}
        \item Agreement between [Your Company Name] and [Recipient Name] on [Date].
        \item Information sharing for the purpose of [specific project or purpose].
    \end{enumerate}
    \textcolor{grey}{$\dots$ (Omitted, see Tables \ref{tab:doc_example_step1b1} - \ref{tab:doc_example_step1b2})}\\
    \textcolor{mygreen}{Score: $6.87$}\\
    \textbf{Response (3/5):} \\
    \textless{}Merged\textgreater{}\\
    NON-DISCLOSURE AGREEMENT (NDA)
    \begin{enumerate}
        \item Agreement between [Your Company Name] and [Recipient Name] on [Date].
        \item Information sharing for the purpose of [specific project or purpose].
    \end{enumerate}
    \textcolor{grey}{$\dots$ (Omitted)}\\
    \textcolor{myblue}{Score: $6.60$}\\
    \textbf{Response (4/5):} \\
    \textless{}Merged\textgreater{}\\
    NON-DISCLOSURE AGREEMENT (NDA)
    \begin{enumerate}
        \item Agreement between [Your Company Name] and [Recipient Name] on [Date].
        \item Information sharing for the purpose of [specific project or purpose].
    \end{enumerate}
    \textcolor{grey}{$\dots$ (Omitted)}\\
    \textcolor{myblue}{Score: $5.78$}\\
    \textbf{Response (5/5):} \\
    \textless{}Merged\textgreater{}\\
    NON-DISCLOSURE AGREEMENT (NDA)
    \begin{enumerate}
        \item Agreement between [Your Company Name] and [Recipient Name] on [Date].
        \item Information sharing for the purpose of [specific project or purpose].
    \end{enumerate}
    \textcolor{grey}{$\dots$ (Omitted)}\\
    \textcolor{myblue}{Score: $6.50$}\\
    \hline
    \end{tabularx}
\end{table*}

\begin{table*}
    \centering
    \renewcommand{\arraystretch}{1.5}
    \caption{Merging 4 example NDAs: \textbf{Step 1 (Highest Scoring Response Part 1)}}
    \label{tab:doc_example_step1b1}
    \begin{tabularx}{\textwidth}{X}
    \hline
    NON-DISCLOSURE AGREEMENT (NDA)
    \begin{enumerate}
        \item Agreement between [Your Company Name] and [Recipient Name] on [Date].
        \item Information sharing for the purpose of [specific project or purpose].
        \item "Confidential Information" includes all potentially commercially valuable information, specifically software development tactics, processes, and in-house research results.
        \item Receiving party is obligated to protect the Confidential Information, use it solely for the disclosed purpose, and not disclose it without consent.
        \item Breach penalties include injunctive relief, other remedies, and a \$200,000 fee per breach.
        \item The Agreement applies to the Parties and their successors and assigns. It contains all related agreements and lack of enforcement doesn't imply waiver.
        \item The Agreement is under the laws of [State].
        \item Signed by [Your Company Name] and [Recipient Name] at the above date.
    \end{enumerate}
    Effective from [Effective Date], this NDA involves [Your Company Name] ("Disclosing Party"), and [Recipient Name] ("Receiving Party"). \\
    Purpose: The Disclosing Party will disclose confidential information related to [Topic of Research] to the Receiving Party for [Purpose]. \\
    Confidential Information: Defined as all non-public reports, data, designs, and other materials provided by the Disclosing Party to the Receiving Party. \\
    Receiving Party's Obligations:
    \begin{enumerate}[label=\alph*.]
        \item Use, reproduce, or distribute the confidential information only for the agreed purpose.
        \item Restrict access to the information to necessary parties, ensuring they abide by strict confidentiality.
        \item Return or destroy all confidential information upon request or at the end of the agreement.
    \end{enumerate}
    Exclusions: Information will not be classified as confidential if it is already known to the Receiving Party, publicly known, or independently developed by the Receiving Party.\\
    Non-Competition: The Receiving Party will not engage in any competing business against the Disclosing Party during the agreement and one year after its termination. \\
    Term and Termination: The agreement is valid for [e.g., "two years"], unless terminated earlier with [e.g., "30 days"] written notice. The Receiving Party's non-disclosure and non-competition obligations persist post-termination. \\
    General Provisions:
    \begin{enumerate}[label=\alph*.]
        \item Governing Law: [Your State]'s laws apply.
        \item Amendments: Only valid if written and signed by both parties.
        \item Entire Agreement: This contract overrules previous related agreements.
    \end{enumerate}
    Signed as of the Effective Date by [Your Company Name] - Disclosing Party [Recipient Name] - Receiving Party. \\
    Entities Involved: \\
    Effective [Date], between [AquaBlue Innovations], established in [State], and [PineTree Solutions], a registered entity. \\
    Objective: \\
    To safeguard classified data during talks of a potential technological alliance.\\
    \hline
    \end{tabularx}
\end{table*}
\begin{table*}
    \centering
    \renewcommand{\arraystretch}{1.5}
    \caption{Merging 4 example NDAs: \textbf{Step 1 (Highest Scoring Response Part 2)}}
    \label{tab:doc_example_step1b2}
    \begin{tabularx}{\textwidth}{X}
    \hline
    Specification of Protected Information:\\
    Particularly: 
    \begin{enumerate}[label=\alph*.]
        \item System designs and architectural schematics.
        \item Proprietary computational algorithms.
    \end{enumerate}
    Receiver's Obligations:
    \begin{enumerate}[label=\alph*.]
        \item Maintain strict non-disclosure using best practices.
        \item Employ solely for the aforementioned aim.
        \item No unveiling without explicit authorization.
    \end{enumerate}
    Violation Ramifications: \\
    A charge of \$280,000 for every infringement, plus possible legal proceedings.\\
    General Terms: \\
    Binding for both parties and any successors. This encapsulates the entire accord. \\
    Legal Reference: \\
    Governed as per [State]'s legal framework. \\
    Attestation: \\
    Duly signed on [Date]. \\
    Contracting Parties: \\
    Dated [Date], drawn between [AquaBlue Innovations], a [State]-based corporation, and [PineTree Solutions], a licensed organization. \\
    Aim:\\
    To protect exclusive insights amidst dialogues for a technological partnership.\\
    Categorization of Sensitive Data:\\
    Includes:
    \begin{enumerate}[label=\alph*.]
        \item Internal software blueprints.
        \item Intellectual property awaiting patents.
    \end{enumerate}
    Commitments of Recipient:
    \begin{enumerate}[label=\alph*.]
        \item Uphold confidentiality, ensuring data integrity.
        \item Utilize strictly for collaborative ventures.
        \item No exposure without prior consensus.
    \end{enumerate}
    Repercussions for Non-Compliance:\\
    \$295,000 fine for each transgression, and the option for legal recourse.\\
    Overall Provisions:\\
    Legally enforceable for signatories and successors. Complete and sole agreement.\\
    Juridical Standpoint:\\
    Under the auspices of [State] laws.\\
    Ratification:\\
    Confirmed and endorsed on [Date].\\
    \hline
    \end{tabularx}
\end{table*}

\begin{table*}
    \centering
    \renewcommand{\arraystretch}{1.5}
    \caption{Merging 4 example NDAs: \textbf{Step 1 (Scoring Example -- Prompt)}}
    \label{tab:doc_example_scoring}
    \begin{tabularx}{\textwidth}{X}
    \hline
    Please score the merged NDA \textless{}S\textgreater{} in terms of how much redundant information is contained, independent of the original NDAs, as well as how much information is retained from the original NDAs. \\
A score of 10 for redundancy implies that absolutely no information is redundant, while a score of 0 implies that at least half of the information is redundant (so everything is at least mentioned twice). \\
A score of 10 for retained information implies that all information from the original NDAs is retained, while a score of 0 implies that no information is retained. \\
You may provide reasoning for your scoring, but the final score for redundancy should be between the tags \textless{}Redundancy\textgreater{} and \textless{}/Redundancy\textgreater{}, and the final score for retained information should be between the tags \textless{}Retained\textgreater{} and \textless{}/Retained\textgreater{}, without any additional text within any of those tags.\\
    Here are NDAs \textless{}Doc1\textgreater{} - \textless{}Doc4\textgreater{}:\\
    \textless{}Doc1\textgreater{}\\
    NON-DISCLOSURE AGREEMENT (NDA) 
    \begin{enumerate}
        \item Agreement between [Your Company Name] and [Recipient Name] on [Date].
    \end{enumerate}
    \textcolor{grey}{$\dots$ (Omitted, see Table \ref{tab:doc_example_nda12})}\\
    \textless{}/Doc1\textgreater{}\\
    \textless{}Doc2\textgreater{}\\
    NON-DISCLOSURE AGREEMENT (NDA)\\
    Effective from [Effective Date], this NDA involves [Your Company Name] ("Disclosing Party"), and [Recipient Name] ("Receiving Party").\\
    \textcolor{grey}{$\dots$ (Omitted, see Table \ref{tab:doc_example_nda12})}\\
    \textless{}/Doc2\textgreater{}\\
    \textless{}Doc3\textgreater{}\\
    CONFIDENTIALITY \& NON-DISCLOSURE AGREEMENT \\
    Entities Involved: \\
    Effective [Date], between [AquaBlue Innovations], established in [State], and \textcolor{grey}{$\dots$ (Omitted, see Table \ref{tab:doc_example_nda3})} \\
    \textless{}/Doc3\textgreater{}\\
    \textless{}Doc4\textgreater{}\\
    SECRECY \& DISCLOSURE AGREEMENT \\
    Contracting Parties: \\
    Dated [Date], drawn between [AquaBlue Innovations], a [State]-based corporation, and \textcolor{grey}{$\dots$ (Omitted, see Table \ref{tab:doc_example_nda4})} \\
    \textless{}/Doc4\textgreater{}\\
    Here is the summary NDA \textless{}S\textgreater{}:\\
    \textless{}S\textgreater{}\\
    NON-DISCLOSURE AGREEMENT (NDA)
    \begin{enumerate}
        \item Agreement between [Your Company Name] and [Recipient Name] on [Date].
        \item Information sharing for the purpose of [specific project or purpose].
    \end{enumerate}
    \textcolor{grey}{$\dots$ (Omitted, see Tables \ref{tab:doc_example_step1b1}-\ref{tab:doc_example_step1b2})} \\
    \textless{}/S\textgreater{}\\
    \hline
    \end{tabularx}
\end{table*}

\begin{table*}
    \centering
    \renewcommand{\arraystretch}{1.5}
    \caption{Merging 4 example NDAs: \textbf{Step 1 (Scoring Example -- Responses)}}
    \label{tab:doc_example_scoring_r}
    \begin{tabularx}{\textwidth}{X}
    \hline
    \textbf{Response (1/3):} \\
    \textless{}Redundancy\textgreater{}5\textless{}/Redundancy\textgreater{}\\
    \textless{}Retained\textgreater{}10\textless{}/Retained\textgreater{}\\
    Reasoning: \\
    In terms of redundancy, there is some overlap in the information contained in the NDAs, such as the definition of "Confidential Information" and the obligations of the receiving party. However, there are also some unique details in each NDA, such as the specific purpose of information sharing and the penalties for breach. Overall, there is a moderate level of redundancy in the merged NDA. \\
    In terms of retained information, all the key elements from the original NDAs are included in the merged NDA. This includes the agreement between the parties, the purpose of information sharing, the definition of confidential information, the obligations of the receiving party, the exclusions for confidential information, non-competition clauses, the term and termination of the agreement, and the general provisions. Therefore, the merged NDA retains all the important information from the original NDAs.\\
    \textbf{Response (2/3):} \\
    \textless{}Redundancy\textgreater{}8\textless{}/Redundancy\textgreater{}\\
    \textless{}Retained\textgreater{}10\textless{}/Retained\textgreater{}\\
    Reasoning: \\
    The merged NDA \textless{}S\textgreater{} contains multiple redundant sections from the original NDAs. For example, the introductory sections, such as the agreement between the parties and the purpose of information sharing, are repeated multiple times. Additionally, sections related to the definition of confidential information, the recipient's obligations, exclusions, non-competition, term and termination, and general provisions are also redundantly included.\\
    However, in terms of retained information, the merged NDA retains all the key elements from the original NDAs, including the parties involved, the objective, specification of protected information, recipient's obligations, violation ramifications, general terms, legal reference, and attestation. Overall, while there is redundancy in the merged NDA, it successfully retains all the necessary information from the original NDAs.\\
    \textbf{Response (3/3):} \\
    \textless{}Redundancy\textgreater{}3\textless{}/Redundancy\textgreater{}\\
    The merged NDA contains some redundancy. There are three sections that have overlapping information between the original NDAs. These sections include:
    \begin{enumerate}
        \item The agreement between the parties (Section 1 in Doc1 and the beginning of Doc2)
        \item Information sharing and the purpose of the NDA (Section 2 in both Doc1 and Doc2)
        \item General provisions such as governing law and amendments (Section 7 in Doc1 and Doc2)
    \end{enumerate}
    \textless{}Retained\textgreater{}9\textless{}/Retained\textgreater{}\\
    The merged NDA retains a significant amount of information from the original NDAs. It includes all key clauses and provisions, such as the definition of confidential information, the obligations of the receiving party, exclusions, non-competition, and general provisions. Only a few minor details, such as specific penalties and fine amounts, are not retained in the merged NDA. Overall, the merged NDA effectively combines the essential elements from the original NDAs.\\
    Note: Remember to remove the reasoning from between the tags when providing the final scores.\\
    \hline
    \textbf{Final Overall Score (Harmonic Mean of Averages):} \\6.87\\
    \hline
    \end{tabularx}
\end{table*}

\begin{table*}
    \centering
    \renewcommand{\arraystretch}{1.5}
    \caption{Merging 4 example NDAs; Note that S1-S3 are the three best scoring merge attemps from the previous step: \textbf{Step 2 (Prompt)}}
    \label{tab:doc_example_step2}
    \begin{tabularx}{\textwidth}{X}
    \hline
    The following NDAs \textless{}S1\textgreater{} - \textless{}S3\textgreater{} each merge the initial NDAs \textless{}Doc1\textgreater{} - \textless{}Doc4\textgreater{}.
    
    Combine the merged NDAs \textless{}S1\textgreater{} - \textless{}S3\textgreater{} into a new one, maximizing their advantages and overall information retention, while minimizing redundancy.
    
    Output only the new NDA between the tags \textless{}Merged\textgreater{} and \textless{}/Merged\textgreater{}, without any additional text.\\

    Here are the original NDAs \textless{}Doc1\textgreater{} - \textless{}Doc4\textgreater{}: \\
    \textless{}Doc1\textgreater{}\\
    \textcolor{grey}{$\dots$ (Omitted, see Table \ref{tab:doc_example_nda12})}\\
    \textless{}/Doc1\textgreater{}\\
    \textless{}Doc2\textgreater{}\\
    \textcolor{grey}{$\dots$ (Omitted, see Table \ref{tab:doc_example_nda12})}\\
    \textless{}/Doc2\textgreater{}\\
    \textless{}Doc3\textgreater{}\\
    \textcolor{grey}{$\dots$ (Omitted, see Table \ref{tab:doc_example_nda3})} \\
    \textless{}/Doc3\textgreater{}\\
    \textless{}Doc4\textgreater{}\\
    \textcolor{grey}{$\dots$ (Omitted, see Table \ref{tab:doc_example_nda4})} \\
    \textless{}/Doc4\textgreater{}\\
    Here are the merged NDAs \textless{}S1\textgreater{} - \textless{}S3\textgreater: \\
    \textless{}S1\textgreater{} \\
    NON-DISCLOSURE AGREEMENT (NDA)
    \begin{enumerate}
        \item Agreement between [Your Company Name] and [Recipient Name] on [Date].
    \end{enumerate}
    \textcolor{grey}{$\dots$ (Omitted)}\\
    \textless{}/S1\textgreater{}\\
    \textless{}S2\textgreater{} \\
    NON-DISCLOSURE AGREEMENT (NDA)
    \begin{enumerate}
        \item Agreement between [Your Company Name] and [Recipient Name] on [Date].
    \end{enumerate}
    \textcolor{grey}{$\dots$ (Omitted, see Tables \ref{tab:doc_example_step1b1}-\ref{tab:doc_example_step1b2})} \\
    \textless{}/S2\textgreater{}\\
    \textless{}S3\textgreater{} \\
    NON-DISCLOSURE AGREEMENT (NDA)
    \begin{enumerate}
        \item Agreement between [Your Company Name] and [Recipient Name] on [Date].
    \end{enumerate}
    \textcolor{grey}{$\dots$ (Omitted)}\\
    \textless{}/S3\textgreater{}\\
    \hline
    \end{tabularx}
\end{table*}

\begin{table*}
    \centering
    \renewcommand{\arraystretch}{1.5}
    \caption{Merging 4 example NDAs: \textbf{Step 2 (Responses)}}
    \label{tab:doc_example_step2r}
    \begin{tabularx}{\textwidth}{X}
    \hline
    For this run, none of the responses improve upon the previous best and thus will all be discarded by the next step. For that reason and due to their long length, we do not present them fully and exclude their scoring.\\
    \textbf{Response (1/5):} \\
    \textless{}Merged\textgreater{}\\
    NON-DISCLOSURE AGREEMENT (NDA)
    \begin{enumerate}
        \item Agreement between [Your Company Name] and [Recipient Name] on [Date].
        \item Information sharing for the purpose of [specific project or purpose].
    \end{enumerate}
    \textcolor{grey}{$\dots$ (Omitted)}\\
    \textcolor{myblue}{Score: $6.50$}\\
    \textbf{Response (2/5):}\\
    \textless{}Merged\textgreater{}\\
    NON-DISCLOSURE AGREEMENT (NDA) \\
    Effective from [Effective Date], this NDA involves [Your Company Name] ("Disclosing Party"), and [Recipient Name] ("Receiving Party").
    \begin{enumerate}
        \item Agreement between [Your Company Name] and [Recipient Name] on [Date].
    \end{enumerate}
    \textcolor{grey}{$\dots$ (Omitted)}\\
    \textcolor{myblue}{Score: $5.62$}\\
    \textbf{Response (3/5):} \\
    \textless{}Merged\textgreater{}\\
    NON-DISCLOSURE AGREEMENT (NDA)
    \begin{enumerate}
        \item Agreement between [Your Company Name] and [Recipient Name] on [Date].
        \item Information sharing for the purpose of [specific project or purpose].
    \end{enumerate}
    \textcolor{grey}{$\dots$ (Omitted)}\\
    \textcolor{myblue}{Score: $6.60$}\\
    \textbf{Response (4/5):} \\
    \textless{}Merged\textgreater{}\\
    NON-DISCLOSURE AGREEMENT (NDA)
    \begin{enumerate}
        \item Agreement between [Your Company Name] and [Recipient Name] on [Date].
        \item Information sharing for the purpose of [specific project or purpose].
    \end{enumerate}
    \textcolor{grey}{$\dots$ (Omitted)}\\
    \textcolor{myblue}{Score: $6.15$}\\
    \textbf{Response (5/5):} \\
    \textless{}Merged\textgreater{}\\
    NON-DISCLOSURE AGREEMENT (NDA)
    \begin{enumerate}
        \item Agreement between [Your Company Name] and [Recipient Name] on [Date].
        \item Information sharing for the purpose of [specific project or purpose].
    \end{enumerate}
    \textcolor{grey}{$\dots$ (Omitted)}\\
    \textcolor{myblue}{Score: $6.22$}\\
    \hline
    \end{tabularx}
\end{table*}

\begin{table*}
    \centering
    \renewcommand{\arraystretch}{1.5}
    \caption{Merging 4 example NDAs: \textbf{Step 3 (Prompt)}}
    \label{tab:doc_example_step3}
    \begin{tabularx}{\textwidth}{X}
    \hline
    The following NDA \textless{}S\textgreater{} merges initial NDAs \textless{}Doc1\textgreater{} - \textless{}Doc4\textgreater{}. \\
    Please improve the merged NDA \textless{}S\textgreater{} by adding more information and removing redundancy.
    Output only the improved NDA, placed between the tags \textless{}Merged\textgreater{} and \textless{}/Merged\textgreater{}, without any additional text. \\

    Here are NDAs \textless{}Doc1\textgreater{} - \textless{}Doc4\textgreater{}: \\
    \textless{}Doc1\textgreater{}\\
    NON-DISCLOSURE AGREEMENT (NDA) 
    \begin{enumerate}
        \item Agreement between [Your Company Name] and [Recipient Name] on [Date].
        \item Information sharing for the purpose of [specific project or purpose].
    \end{enumerate}
    \textcolor{grey}{$\dots$ (Omitted, see Table \ref{tab:doc_example_nda12})}\\
    \textless{}/Doc1\textgreater{}\\
    \textless{}Doc2\textgreater{}\\
    NON-DISCLOSURE AGREEMENT (NDA)\\
    Effective from [Effective Date], this NDA involves [Your Company Name] ("Disclosing Party"), and [Recipient Name] ("Receiving Party").
    \begin{enumerate}
        \item Purpose: The Disclosing Party will disclose confidential information related to [Topic of Research] to the Receiving Party for [Purpose].
    \end{enumerate}
    \textcolor{grey}{$\dots$ (Omitted, see Table \ref{tab:doc_example_nda12})}\\
    \textless{}/Doc2\textgreater{}\\
    \textless{}Doc3\textgreater{}\\
    CONFIDENTIALITY \& NON-DISCLOSURE AGREEMENT \\
    Entities Involved: \\
    Effective [Date], between [AquaBlue Innovations], established in [State], and [PineTree Solutions], a registered entity.\\
    Objective:\\
    To safeguard classified data during talks of a potential \textcolor{grey}{$\dots$ (Omitted, see Table \ref{tab:doc_example_nda3})} \\
    \textless{}/Doc3\textgreater{}\\
    \textless{}Doc4\textgreater{}\\
    SECRECY \& DISCLOSURE AGREEMENT \\
    Contracting Parties: \\
    Dated [Date], drawn between [AquaBlue Innovations], a [State]-based corporation, and [PineTree Solutions], a licensed organization.\\
    Aim:\\
    To protect exclusive insights amidst \textcolor{grey}{$\dots$ (Omitted, see Table \ref{tab:doc_example_nda4})} \\
    \textless{}/Doc4\textgreater{}\\
    Here is the summary NDA \textless{}S\textgreater{}: \\
    \textless{}S\textgreater{} \\
    NON-DISCLOSURE AGREEMENT (NDA)
    \begin{enumerate}
        \item Agreement between [Your Company Name] and [Recipient Name] on [Date].
        \item Information sharing for the purpose of [specific project or purpose].
    \end{enumerate}
    \textcolor{grey}{$\dots$ (Omitted, see Tables \ref{tab:doc_example_step1b1}-\ref{tab:doc_example_step1b2})} \\
    \textless{}/S\textgreater{}\\
    \hline
    \end{tabularx}
\end{table*}

\begin{table*}
    \centering
    \renewcommand{\arraystretch}{1.5}
    \caption{Merging 4 example NDAs; we only present the highest scoring response fully in Table \ref{tab:doc_example_step3b}: \textbf{Step 3 (Responses Part 1)}}
    \label{tab:doc_example_step3r1}
    \begin{tabularx}{\textwidth}{X}
    \hline
    \textbf{Response (1/10):} \\
    \textless{}Merged\textgreater{}\\
    NON-DISCLOSURE AGREEMENT (NDA)
    \begin{enumerate}
        \item Agreement between [Your Company Name] and [Recipient Name] on [Date].
        \item Information sharing for the purpose of [specific project or purpose].
    \end{enumerate}
    \textcolor{grey}{$\dots$ (Omitted)}\\
    \textcolor{myblue}{Score: $6.35$}\\
    \textbf{Response (2/10):}\\
    \textless{}Merged\textgreater{}\\
    NON-DISCLOSURE AGREEMENT (NDA)
    \begin{enumerate}
        \item Agreement between [Your Company Name] and [Recipient Name] on [Date].
        \item Information sharing for the purpose of [specific project or purpose].
    \end{enumerate}
    \textcolor{grey}{$\dots$ (Omitted)}\\
    \textcolor{myblue}{Score: $6.35$}\\
    \textbf{Response (3/10):} \\
    \textless{}Merged\textgreater{}\\
    NON-DISCLOSURE AGREEMENT (NDA)
    \begin{enumerate}
        \item Agreement between [Your Company Name] and [Recipient Name] on [Date], for the purpose of [specific project or purpose].
        \item "Confidential Information" includes all potentially commercially valuable information, specifically software development tactics, processes, in-house research results, system designs, architectural schematics, proprietary computational algorithms, internal software blueprints, and intellectual property awaiting patents.
    \end{enumerate}
    \textcolor{grey}{$\dots$ (Omitted)}\\
    \textcolor{myblue}{Score: $6.87$}\\
    \textbf{Response (4/10):} \\
    \textless{}Merged\textgreater{}\\
    NON-DISCLOSURE AGREEMENT (NDA)
    \begin{enumerate}
        \item Agreement between [Your Company Name] and [Recipient Name] on [Date] for the purpose of [specific project or purpose].
        \item "Confidential Information" includes all potentially commercially valuable information, such as software development tactics, processes, in-house research results, system designs, architectural schematics, proprietary computational algorithms, internal software blueprints, and intellectual property awaiting patents.
    \end{enumerate}
    \textcolor{grey}{$\dots$ (Omitted, see Table \ref{tab:doc_example_step3b}))}\\
    \textcolor{mygreen}{Score: $7.78$}\\
    \textbf{Response (5/10):} \\
    \textless{}Merged\textgreater{}\\
    NON-DISCLOSURE AGREEMENT (NDA)
    \begin{enumerate}
        \item Agreement between [Your Company Name] and [Recipient Name] on [Date].
        \item Information sharing for the purpose of [specific project or purpose].
    \end{enumerate}
    \textcolor{grey}{$\dots$ (Omitted)}\\
    \textcolor{myblue}{Score: $6.43$}\\
    \hline
    \end{tabularx}
\end{table*}

\begin{table*}
    \centering
    \renewcommand{\arraystretch}{1.5}
    \caption{Merging 4 example NDAs; we only present the highest scoring response fully in Table \ref{tab:doc_example_step3b}: \textbf{Step 3 (Responses Part 2)}}
    \label{tab:doc_example_step3r2}
    \begin{tabularx}{\textwidth}{X}
    \hline
    \textbf{Response (6/10):} \\
    \textless{}Merged\textgreater{}\\
    NON-DISCLOSURE AGREEMENT (NDA)
    \begin{enumerate}
        \item Agreement between [Your Company Name] and [Recipient Name] on [Date].
        \item Information sharing for the purpose of [specific project or purpose].
    \end{enumerate}
    \textcolor{grey}{$\dots$ (Omitted)}\\
    \textcolor{myblue}{Score: $6.35$}\\
    \textbf{Response (7/10):}\\
    \textless{}Merged\textgreater{}\\
    NON-DISCLOSURE AGREEMENT (NDA)
    \begin{enumerate}
        \item Agreement between [Your Company Name] and [Recipient Name] on [Date] for the purpose of [specific project or purpose].
        \item "Confidential Information" includes all potentially commercially valuable information, specifically software development tactics, processes, in-house research results, system designs, architectural schematics, proprietary computational algorithms, internal software blueprints, and intellectual property awaiting patents.
    \end{enumerate}
    \textcolor{grey}{$\dots$ (Omitted)}\\
    \textcolor{myblue}{Score: $6.70$}\\
    \textbf{Response (8/10):} \\
    \textless{}Merged\textgreater{}\\
    NON-DISCLOSURE AGREEMENT (NDA)
    \begin{enumerate}
        \item Agreement between [Your Company Name] and [Recipient Name] on [Date] for the purpose of [specific project or purpose].
        \item "Confidential Information" includes potentially commercially valuable information, such as software development tactics, processes, in-house research results, non-public reports, data, designs, and other materials provided by the Disclosing Party.
    \end{enumerate}
    \textcolor{grey}{$\dots$ (Omitted)}\\
    \textcolor{myblue}{Score: $5.33$}\\
    \textbf{Response (9/10):} \\
    \textless{}Merged\textgreater{}\\
    NON-DISCLOSURE AGREEMENT (NDA)
    \begin{enumerate}
        \item Agreement between [Your Company Name] and [Recipient Name] on [Date].
        \item Information sharing for the purpose of [specific project or purpose], including [Topic of Research].
    \end{enumerate}
    \textcolor{grey}{$\dots$ (Omitted)}\\
    \textcolor{myblue}{Score: $7.30$}\\
    \textbf{Response (10/10):} \\
    \textless{}Merged\textgreater{}\\
    NON-DISCLOSURE AGREEMENT (NDA)\\
    Agreement between [Your Company Name] and [Recipient Name] on [Date].\\
    Information sharing for the purpose of [specific project or purpose].\\
    \textcolor{grey}{$\dots$ (Omitted)}\\
    \textcolor{myblue}{Score: $6.40$}\\
    \hline
    \end{tabularx}
\end{table*}

\begin{table*}
    \centering
    \renewcommand{\arraystretch}{1.5}
    \caption{Merging 4 example NDAs: \textbf{Step 3 (Highest Scoring Response)}}
    \label{tab:doc_example_step3b}
    \begin{tabularx}{\textwidth}{X}
    \hline
    NON-DISCLOSURE AGREEMENT (NDA)
    \begin{enumerate}
        \item Agreement between [Your Company Name] and [Recipient Name] on [Date] for the purpose of [specific project or purpose].
        \item "Confidential Information" includes all potentially commercially valuable information, such as software development tactics, processes, in-house research results, system designs, architectural schematics, proprietary computational algorithms, internal software blueprints, and intellectual property awaiting patents.
        \item Receiving party is obligated to protect the Confidential Information, use it solely for the disclosed purpose, and not disclose it without consent.
        \item Breach penalties include injunctive relief, other remedies, and a fine per breach (e.g., \$200,000 or \$280,000 for each infringement, depending on the agreement).
        \item The Agreement applies to the Parties and their successors and assigns. It contains all related agreements and lack of enforcement doesn't imply waiver.
        \item The Agreement is under the laws of [State] and is valid for a specified period (e.g., two years) unless terminated earlier with written notice (e.g., 30 days).
        \item The Receiving Party agrees not to engage in any competing business against the Disclosing Party during the agreement and a specified period after its termination (e.g., one year).
        \item The Agreement can only be amended if written and signed by both parties, and overrules previous related agreements.
        \item The Agreement is legally enforceable and binding for signatories and successors.
        \item  The Agreement is governed by the laws of [State].
    \end{enumerate}
    Signed as of the Effective Date by [Your Company Name] - Disclosing Party [Recipient Name] - Receiving Party, along with the other parties involved. \\
    \hline
    \end{tabularx}
\end{table*}

\begin{table*}
    \centering
    \renewcommand{\arraystretch}{1.5}
    \caption{Merging 4 example NDAs: \textbf{Step 3 (Scoring Example -- Prompt)}}
    \label{tab:doc_example_scoring_s3}
    \begin{tabularx}{\textwidth}{X}
    \hline
    Please score the merged NDA \textless{}S\textgreater{} in terms of how much redundant information is contained, independent of the original NDAs, as well as how much information is retained from the original NDAs. \\
A score of 10 for redundancy implies that absolutely no information is redundant, while a score of 0 implies that at least half of the information is redundant (so everything is at least mentioned twice). \\
A score of 10 for retained information implies that all information from the original NDAs is retained, while a score of 0 implies that no information is retained. \\
You may provide reasoning for your scoring, but the final score for redundancy should be between the tags \textless{}Redundancy\textgreater{} and \textless{}/Redundancy\textgreater{}, and the final score for retained information should be between the tags \textless{}Retained\textgreater{} and \textless{}/Retained\textgreater{}, without any additional text within any of those tags.\\
    Here are NDAs \textless{}Doc1\textgreater{} - \textless{}Doc4\textgreater{}:\\
    \textless{}Doc1\textgreater{}\\
    NON-DISCLOSURE AGREEMENT (NDA) 
    \begin{enumerate}
        \item Agreement between [Your Company Name] and [Recipient Name] on [Date].
    \end{enumerate}
    \textcolor{grey}{$\dots$ (Omitted, see Table \ref{tab:doc_example_nda12})}\\
    \textless{}/Doc1\textgreater{}\\
    \textless{}Doc2\textgreater{}\\
    NON-DISCLOSURE AGREEMENT (NDA)\\
    Effective from [Effective Date], this NDA involves [Your Company Name] ("Disclosing Party"), and [Recipient Name] ("Receiving Party").\\
    \textcolor{grey}{$\dots$ (Omitted, see Table \ref{tab:doc_example_nda12})}\\
    \textless{}/Doc2\textgreater{}\\
    \textless{}Doc3\textgreater{}\\
    CONFIDENTIALITY \& NON-DISCLOSURE AGREEMENT \\
    Entities Involved: \\
    Effective [Date], between [AquaBlue Innovations], established in [State], and \textcolor{grey}{$\dots$ (Omitted, see Table \ref{tab:doc_example_nda3})} \\
    \textless{}/Doc3\textgreater{}\\
    \textless{}Doc4\textgreater{}\\
    SECRECY \& DISCLOSURE AGREEMENT \\
    Contracting Parties: \\
    Dated [Date], drawn between [AquaBlue Innovations], a [State]-based corporation, and \textcolor{grey}{$\dots$ (Omitted, see Table \ref{tab:doc_example_nda4})} \\
    \textless{}/Doc4\textgreater{}\\
    Here is the summary NDA \textless{}S\textgreater{}:\\
    \textless{}S\textgreater{}\\
    NON-DISCLOSURE AGREEMENT (NDA)
    \begin{enumerate}
        \item Agreement between [Your Company Name] and [Recipient Name] on [Date] for the purpose of [specific project or purpose].
        \item "Confidential Information" includes all potentially commercially valuable information, such as software development tactics, processes, in-house research results, system designs, architectural schematics, proprietary computational algorithms, internal software blueprints, and intellectual property awaiting patents.
    \end{enumerate}
    \textcolor{grey}{$\dots$ (Omitted, see Table \ref{tab:doc_example_step3b})} \\
    \textless{}/S\textgreater{}\\
    \hline
    \end{tabularx}
\end{table*}

\begin{table*}
    \centering
    \renewcommand{\arraystretch}{1.5}
    \caption{Merging 4 example NDAs: \textbf{Step 3 (Scoring Example -- Responses)}}
    \label{tab:doc_example_scoring_s3_r}
    \begin{tabularx}{\textwidth}{X}
    \hline
    \textbf{Response (1/3):} \\
    \textless{}Redundancy\textgreater{}5\textless{}/Redundancy\textgreater{}\\
    \textless{}Retained\textgreater{}8\textless{}/Retained\textgreater{}\\
    Reasoning: \\
    The merged NDA \textless{}S\textgreater{} retains most of the information from the original NDAs, combining relevant clauses and terms into a single comprehensive agreement. However, some redundancy is present as similar information is mentioned multiple times, such as the definition of "Confidential Information" and the obligations of the receiving party. Additionally, some specific details from the original NDAs are not included in the merged NDA, such as the exclusion criteria in Doc2 and the provisions for non-competition in Doc3 and Doc4. Nonetheless, the merged NDA captures the essence of the original NDAs and includes all key information necessary for a non-disclosure agreement.\\
    
    \textbf{Response (2/3):} \\
    \textless{}Redundancy\textgreater{}8\textless{}/Redundancy\textgreater{}\\
    \textless{}Retained\textgreater{}10\textless{}/Retained\textgreater{}\\
    Reasoning: \\
    The merged NDA \textless{}S\textgreater{} retains all the key information from the original NDAs, including the purpose of the agreement, the definition of confidential information, the obligations of the receiving party, breach penalties, applicability to parties and successors, governing law, and signature requirements. \\
    There is some redundant information, such as the inclusion of system designs, architectural schematics, proprietary computational algorithms, internal software blueprints, and intellectual property awaiting patents from both Doc3 and Doc4. However, this redundancy is minor and does not significantly impact the overall usefulness of the NDA. \\
    Overall, the merged NDA \textless{}S\textgreater{} retains all important information while minimizing redundancy, making it highly effective in conveying the necessary terms and obligations of the agreement. Therefore, it receives a score of 8 for redundancy and a score of 10 for retained information. \\
    \textbf{Response (3/3):} \\
    \textless{}Redundancy\textgreater{}7\textless{}/Redundancy\textgreater{}\\
    \textless{}Retained\textgreater{}10\textless{}/Retained\textgreater{}\\
    Reasoning: \\
    The merged NDA \textless{}S\textgreater{} contains some redundant information, such as the definition of ''Confidential Information'' and the obligations of the receiving party. However, it retains all the important information from the original NDAs, including the purpose of the agreement, the specific types of confidential information, the breach penalties, the applicability to parties and successors, the governing law, and the provisions regarding non-competition, amendment, and termination. Additionally, it incorporates the additional information about the agreement's validity period and the option for legal recourse in case of non-compliance. Therefore, the merged NDA \textless{}S\textgreater{} scores a 7 for redundancy and a 10 for retained information. \\
    \hline
    \textbf{Final Overall Score (Harmonic Mean of Averages):} \\7.78\\
    \hline
    \end{tabularx}
\end{table*}

\begin{table*}
    \centering
    \renewcommand{\arraystretch}{1.5}
    \caption{Merging 4 example NDAs: \textbf{Final Merged NDA}}
    \label{tab:doc_example_finale}
    \begin{tabularx}{\textwidth}{X}
    \hline
    NON-DISCLOSURE AGREEMENT (NDA)
    \begin{enumerate}
        \item Agreement between [Your Company Name] and [Recipient Name] on [Date] for the purpose of [specific project or purpose].
        \item "Confidential Information" includes all potentially commercially valuable information, such as software development tactics, processes, in-house research results, system designs, architectural schematics, proprietary computational algorithms, internal software blueprints, and intellectual property awaiting patents.
        \item Receiving party is obligated to protect the Confidential Information, use it solely for the disclosed purpose, and not disclose it without consent.
        \item Breach penalties include injunctive relief, other remedies, and a fine per breach (e.g., \$200,000 or \$280,000 for each infringement, depending on the agreement).
        \item The Agreement applies to the Parties and their successors and assigns. It contains all related agreements and lack of enforcement doesn't imply waiver.
        \item The Agreement is under the laws of [State] and is valid for a specified period (e.g., two years) unless terminated earlier with written notice (e.g., 30 days).
        \item The Receiving Party agrees not to engage in any competing business against the Disclosing Party during the agreement and a specified period after its termination (e.g., one year).
        \item The Agreement can only be amended if written and signed by both parties, and overrules previous related agreements.
        \item The Agreement is legally enforceable and binding for signatories and successors.
        \item The Agreement is governed by the laws of [State].
    \end{enumerate}
    Signed as of the Effective Date by [Your Company Name] - Disclosing Party [Recipient Name] - Receiving Party, along with the other parties involved.\\
    \hline
    \end{tabularx}
\end{table*}

\clearpage

\section{Evaluation - \nameAS Configurations}
We detail the concrete operations that \nameAS was configured with to solve the set intersection and sorting use cases.
% \subsection{Set Intersection}

% \paragraph{$32$ Elements:}
% \begin{align*}
% Generate(k=1)&: \text{Split second set into two halves} \\
% \textit{For each subset}&: \\
%   Generate&(k=5): \text{Determine intersected subset of subset} \\ &\text{and first input set} \\
%   Score&(k=1): \text{Score locally the intersected subsets} \\
%   KeepBe&stN(1): \text{Keep the best intersected subset} \\
% Aggregate(10)&: \text{Merge both intersected subsets} \\
% Score(k=1)&: \text{Score locally the intersected result sets} \\
% KeepBestN(1)&: \text{Keep the best result} \\
% GroundTruth()&: \text{Compare to precomputed result}
% \end{align*}

\begin{listing}[h]
\caption{\nameAS configuration for the set intersection use case with 32 elements}
\begin{lstlisting}[language=Python]
Generate(k=1) # Split second set into two halves of 16 elements
foreach subset:
  Generate(k=5) # Determine intersected subset of subset and first input set
  Score(k=1) # Score locally the intersected subsets
  KeepBestN(1) # Keep the best intersected subset
Aggregate(10) # Merge both intersected subsets
Score(k=1) # Score locally the intersected result sets
KeepBestN(1) # Keep the best result
GroundTruth() # Compare to precomputed result
\end{lstlisting}
\end{listing}

\begin{listing}[h]
\caption{\nameAS configuration for the set intersection use case with 64 elements}
\begin{lstlisting}[language=Python]
Generate(k=1) # Split second set into four parts of 16 elements
foreach subset:
  Generate(k=5) # Determine intersected subset of subset and first input set
  Score(k=1) # Score locally the intersected subsets
  KeepBestN(1) # Keep the best intersected subset
merge step 1:
  Aggregate(10) # Merge intersected subsets 1 and 2
  Score(k=1) # Score locally the intersected result sets
  KeepBestN(1) # Keep the best result
merge step 2:
  Aggregate(10) # Merge intersected subsets 3 and 4
  Score(k=1) # Score locally the intersected result sets
  KeepBestN(1) # Keep the best result
final merge:
  Aggregate(10) # Merge intermediate intersected subsets from merge step 1 and 2
  Score(k=1) # Score locally the intersected result sets
  KeepBestN(1) # Keep the best result
GroundTruth() # Compare to precomputed result
\end{lstlisting}
\end{listing}

\begin{listing}[h]
\caption{\nameAS configuration for the set intersection use case with 128 elements}
\begin{lstlisting}[language=Python]
Generate(k=1) # Split second set into eight parts of 16 elements
foreach subset:
  Generate(k=5) # Determine intersected subset of subset and first input set
  Score(k=1) # Score locally the intersected subsets
  KeepBestN(1) # Keep the best intersected subset
merge step 1:
  Aggregate(5) # Merge intersected subsets 1 and 2
  Score(k=1) # Score locally the intersected result sets
  KeepBestN(1) # Keep the best result
merge step 2:
  Aggregate(5) # Merge intersected subsets 3 and 4
  Score(k=1) # Score locally the intersected result sets
  KeepBestN(1) # Keep the best result
merge step 3:
  Aggregate(5) # Merge intersected subsets 5 and 6
  Score(k=1) # Score locally the intersected result sets
  KeepBestN(1) # Keep the best result
merge step 4:
  Aggregate(5) # Merge intersected subsets 7 and 8
  Score(k=1) # Score locally the intersected result sets
\end{lstlisting}
\end{listing}

\begin{listing}[h]
\caption{\nameAS configuration for the set intersection use case with 128 elements (cont.)}
\begin{lstlisting}[language=Python,firstnumber=21]
  KeepBestN(1) # Keep the best result
merge step 5:
  Aggregate(5) # Merge intermediate intersected subsets from merge step 1 and 2
  Score(k=1) # Score locally the intersected result sets
  KeepBestN(1) # Keep the best result
merge step 6:
  Aggregate(5) # Merge intermediate intersected subsets from merge step 3 and 4
  Score(k=1) # Score locally the intersected result sets
  KeepBestN(1) # Keep the best result
final merge:
  Aggregate(5) # Merge intermediate intersected subsets from merge step 5 and 6
  Score(k=1) # Score locally the intersected result sets
  KeepBestN(1) # Keep the best result
GroundTruth() # Compare to precomputed result
\end{lstlisting}
\end{listing}

% \paragraph{$32$ Elements:}
% \begin{align*}
% Generate(k=1)&: \text{Split list into two halves} \\
% \textit{For each list part}&: \\
%   Generate&(k=5): \text{Sort list part} \\
%   Score&(k=1): \text{Score partially sorted list} \\
%   KeepBe&stN(1): \text{Keep the best partially sorted list} \\
% Aggregate(10)&: \text{Merge both partially sorted lists} \\
% Score(k=1)&: \text{Score locally the sorted result lists} \\
% KeepBestN(1)&: \text{Keep the best result} \\
% Generate(k=10)&: \text{Try to improve solution} \\
% Score(k=1)&: \text{Score locally the sorted result lists} \\
% KeepBestN(1)&: \text{Keep the best result} \\
% GroundTruth()&: \text{Compare to precomputed result}
% \end{align*}

\begin{listing}[h]
\caption{\nameAS configuration for the sorting use case with 32 elements}
\begin{lstlisting}[language=Python]
Generate(k=1) # Split list into two halves of 16 elements
foreach list part:
  Generate(k=5) # Sort list part
  Score(k=1): # Score partially sorted list
  KeepBestN(1): # Keep the best partially sorted list
Aggregate(10) # Merge both partially sorted lists
Score(k=1) # Score locally the sorted result lists
KeepBestN(1) # Keep the best result
Generate(k=10) # Try to improve solution
Score(k=1) # Score locally the sorted result lists
KeepBestN(1) # Keep the best result
GroundTruth() # Compare to precomputed result
\end{lstlisting}
\end{listing}

% \paragraph{$64$ Elements:}
% \begin{align*}
% Generate(k=1)&: \text{Split list into four parts} \\
% \textit{For each list part}&: \\
%   Generate&(k=5): \text{Sort list part} \\
%   Score&(k=1): \text{Score partially sorted list} \\
%   KeepBe&stN(1): \text{Keep the best partially sorted list} \\
% \textit{Merge step 1}&: \\
% Aggregate(10)&: \text{Merge partially sorted lists 1 and 2} \\
% Score(k=1)&: \text{Score locally the partially sorted result lists} \\
% KeepBestN(1)&: \text{Keep the best result} \\
% Generate(k=5)&: \text{Try to improve the partial solution} \\
% Score(k=1)&: \text{Score locally the partially sorted result lists} \\
% KeepBestN(1)&: \text{Keep the best result} \\
% \textit{Merge step 2}&: \\
% Aggregate(10)&: \text{Merge partially sorted lists 3 and 4} \\
% Score(k=1)&: \text{Score locally the partially sorted result lists} \\
% KeepBestN(1)&: \text{Keep the best result} \\
% Generate(k=5)&: \text{Try to improve the partial solution} \\
% Score(k=1)&: \text{Score locally the partially sorted result lists} \\
% KeepBestN(1)&: \text{Keep the best result} \\
% \textit{Final merge}&: \\
% Aggregate(10)&: \text{Merge partially sorted lists} \\ &\text{from merge step 1 and 2} \\
% Score(k=1)&: \text{Score locally the sorted result lists} \\
% KeepBestN(1)&: \text{Keep the best result} \\
% Generate(k=10)&: \text{Try to improve solution} \\
% Score(k=1)&: \text{Score locally the sorted result lists} \\
% KeepBestN(1)&: \text{Keep the best result} \\
% GroundTruth()&: \text{Compare to precomputed result}
% \end{align*}

\begin{listing}[h]
\caption{\nameAS configuration for the sorting use case with 64 elements}
\begin{lstlisting}[language=Python]
Generate(k=1) # Split list into four parts of 16 elements
foreach list part:
  Generate(k=5) # Sort list part
  Score(k=1) # Score partially sorted list
  KeepBestN(1) # Keep the best partially sorted list
merge step 1:
  Aggregate(10) # Merge partially sorted lists 1 and 2
  Score(k=1) # Score locally the partially sorted result lists
  KeepBestN(1) # Keep the best result
  Generate(k=5) # Try to improve the partial solution
  Score(k=1) # Score locally the partially sorted result lists
  KeepBestN(1) # Keep the best result
merge step 2:
  Aggregate(10) # Merge partially sorted lists 3 and 4
  Score(k=1) # Score locally the partially sorted result lists
  KeepBestN(1) # Keep the best result
  Generate(k=5) # Try to improve the partial solution
  Score(k=1) # Score locally the partially sorted result lists
  KeepBestN(1) # Keep the best result
final merge:
  Aggegrate(10) # Merge partially sorted lists from merge step 1 and 2
  Score(k=1) # Score locally the sorted result lists
  KeepBestN(1) # Keep the best result
  Generate(k=10) # Try to improve solution
  Score(k=1) # Score locally the sorted result lists
  KeepBestN(1) # Keep the best result
GroundTruth() # Compare to precomputed result
\end{lstlisting}
\end{listing}

\begin{listing}[h]
\caption{\nameAS configuration for the sorting use case with 128 elements}
\begin{lstlisting}[language=Python]
Generate(k=1) # Split list into eight parts of 16 elements
foreach list part:
  Generate(k=5) # Sort list part
  Score(k=1) # Score partially sorted list
  KeepBestN(1) # Keep the best partially sorted list
merge step 1:
  Aggregate(10) # Merge partially sorted lists 1 and 2
  Score(k=1) # Score locally the partially sorted result lists
  KeepBestN(1) # Keep the best result
  Generate(k=5) # Try to improve the partial solution
  Score(k=1) # Score locally the partially sorted result lists
  KeepBestN(1) # Keep the best result
merge step 2:
  Aggregate(10) # Merge partially sorted lists 3 and 4
  Score(k=1) # Score locally the partially sorted result lists
  KeepBestN(1) # Keep the best result
  Generate(k=5) # Try to improve the partial solution
  Score(k=1) # Score locally the partially sorted result lists
  KeepBestN(1) # Keep the best result
merge step 3:
  Aggregate(10) # Merge partially sorted lists 5 and 6
  Score(k=1) # Score locally the partially sorted result lists
  KeepBestN(1) # Keep the best result
  Generate(k=5) # Try to improve the partial solution
  Score(k=1) # Score locally the partially sorted result lists
  KeepBestN(1) # Keep the best result
merge step 4:
  Aggregate(10) # Merge partially sorted lists 7 and 8
  Score(k=1) # Score locally the partially sorted result lists
  KeepBestN(1) # Keep the best result
  Generate(k=5) # Try to improve the partial solution
  Score(k=1) # Score locally the partially sorted result lists
  KeepBestN(1) # Keep the best result
merge step 5:
  Aggregate(10) # Merge partially sorted lists from merge step 1 and 2
  Score(k=1) # Score locally the partially sorted result lists
  KeepBestN(1) # Keep the best result
  Generate(k=5) # Try to improve the partial solution
  Score(k=1) # Score locally the partially sorted result lists
  KeepBestN(1) # Keep the best result
merge step 6:
  Aggregate(10) # Merge partially sorted lists from merge step 3 and 4
  Score(k=1) # Score locally the partially sorted result lists
  KeepBestN(1) # Keep the best result
  Generate(k=5) # Try to improve the partial solution
  Score(k=1) # Score locally the partially sorted result lists
  KeepBestN(1 # Keep the best result
final merge:
  Aggregate(10) # Merge partially sorted lists from merge step 5 and 6
  Score(k=1) # Score locally the partially sorted result lists
  KeepBestN(1) # Keep the best result
  Generate(k=10) # Try to improve solution
  Score(k=1) # Score locally the sorted result lists
  KeepBestN(1) # Keep the best result
GroundTruth() # Compare to precomputed result
\end{lstlisting}
\end{listing}